%% file: neurips_2026.tex
\documentclass{article}

\PassOptionsToPackage{numbers, sort&compress}{natbib}

\usepackage[eandd, final]{neurips_2026}

\usepackage[utf8]{inputenc}
\usepackage[T1]{fontenc}
\usepackage{hyperref}
\usepackage{url}
\usepackage{booktabs}
\usepackage{multirow}
\usepackage{graphicx}
\usepackage[table]{xcolor}
\usepackage{amsmath}
\usepackage[most]{tcolorbox}
\usepackage{fontawesome5}
\usepackage{amsfonts}
\usepackage{dsfont}
\usepackage{nicefrac}
\usepackage{microtype}
\usepackage{pifont}
\usepackage{enumitem}
\usepackage{wrapfig}
\usepackage{float}
\usepackage{placeins}
\usepackage{listings}
\usepackage{etoc}

\definecolor{rankgold}{HTML}{FFF3CD}
\definecolor{ranksilver}{HTML}{D6EAF8}
\definecolor{rankbronze}{HTML}{E8E8E8}
\newcommand{\first}[1]{\cellcolor{rankgold}\textbf{#1}}
\newcommand{\second}[1]{\cellcolor{ranksilver}\underline{#1}}
\newcommand{\third}[1]{\cellcolor{rankbronze}#1}
\newcommand{\rankkey}[2]{{\setlength{\fboxsep}{0.5pt}\colorbox{#1}{\vphantom{Gy}\,#2\,}}}

\title{PCFBench: How Far Can Large Vision-Language Models Go in Physics-aware Photonic Inverse Design?}

\author{%
  Shengchao Chen$^{1}$,
  Ting Shu$^{2}$,
  Sufen Ren$^{3}$ \\[0.4ex]
  $^{1}$Australian AI Institute, University of Technology Sydney \\
  $^{2}$School of Artificial Intelligence, Shenzhen University \\
  $^{3}$School of Information and Communication Engineering, Hainan University \\[0.4ex]
  \texttt{shengchao.chen.uts@gmail.com}
}

\newcommand{\pcfseticon}[2]{\expandafter\def\csname pcficon@#1\endcsname{#2}}
\pcfseticon{ai2}{\includegraphics[height=1.85ex,keepaspectratio]{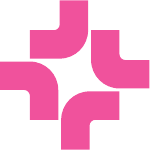}}
\pcfseticon{claude}{\includegraphics[height=1.85ex,keepaspectratio]{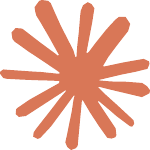}}
\pcfseticon{gemini}{\includegraphics[height=1.85ex,keepaspectratio]{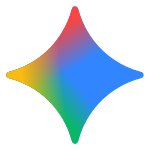}}
\pcfseticon{gemma}{\includegraphics[height=1.85ex,keepaspectratio]{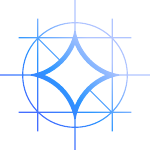}}
\pcfseticon{grok}{\includegraphics[height=1.85ex,keepaspectratio]{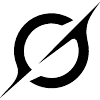}}
\pcfseticon{meta}{\includegraphics[height=1.85ex,keepaspectratio]{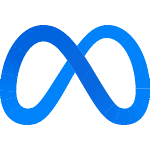}}
\pcfseticon{mistral}{\includegraphics[height=1.85ex,keepaspectratio]{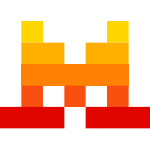}}
\pcfseticon{openai}{\includegraphics[height=1.85ex,keepaspectratio]{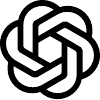}}
\pcfseticon{qwen}{\includegraphics[height=1.85ex,keepaspectratio]{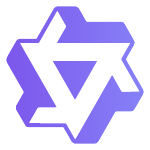}}
\pcfseticon{zhipu}{\includegraphics[height=1.85ex,keepaspectratio]{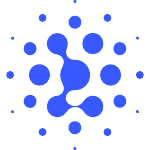}}
\newcommand{\modelicon}[1]{%
  \raisebox{-0.35ex}{%
    \csname pcficon@#1\endcsname%
  }%
}
\newcommand{\model}[3]{\modelicon{#1}\,#2~\cite{#3}}

\begin{document}

\maketitle

\begin{abstract}
Photonic crystal fibers (PCFs) underpin critical applications in telecommunications, sensing, and high-power laser delivery, yet their design remains a manual, expert-intensive process that couples geometric intuition with deep waveguide physics. Vision-language models (VLMs) offer a promising path toward automating PCF design, but \textbf{\emph{can they truly reason about photonic physics from microscopy images, or do they merely recognize visual patterns?}} We introduce \textbf{PCFBench}, the first comprehensive benchmark for evaluating VLMs on the perception, reasoning, and decision steps of the photonic inverse-design loop. PCFBench spans 26 tasks organized into six categories of increasing cognitive demand (Geometry Perception, Physics Understanding, Text Generation, Multi-modal Reasoning, Inverse-Design Decision, and Code Generation), built from over 220K FDTD simulation samples across 18 structurally diverse fiber families. We evaluate both individual VLMs on isolated tasks and multi-agent systems (MAS) on the full sequential pipeline from perception to design. Our results reveal a striking \emph{perception--design gap}: current models can identify fiber geometry with moderate success, yet consistently fail to translate that visual understanding into physics-aware design actions. Moreover, cascading agents through the pipeline exposes significant error propagation, where upstream perception mistakes compound into downstream design failures. These findings suggest that neither model scaling nor naive agent chaining suffices, and that closing the loop demands explicit physical grounding. PCFBench provides the community with a rigorous, multi-dimensional testbed to measure progress toward AI-assisted photonic engineering. The project is available at \href{https://pcfbench.github.io/}{\textcolor{blue}{PCFBench}}.
\end{abstract}

\addtocontents{toc}{\protect\setcounter{tocdepth}{-1}}
\input{sections/introduction}
\input{sections/related_work}
\input{sections/pcfbench}
\input{sections/experiments}

\input{sections/conclusion}

\input{sections/acknowledgements}

\bibliographystyle{unsrtnat}
\bibliography{references}

\clearpage
\appendix
\addtocontents{toc}{\protect\setcounter{tocdepth}{2}}

\bgroup
  \etocsetnexttocdepth{subsection}
  \etocsettocstyle{%
    \section*{Contents of the Appendix}%
    \noindent\hrulefill\par\medskip%
  }{%
    \par\medskip\noindent\hrulefill%
  }
  \tableofcontents
\egroup
\bigskip

\input{sections/appendix}

\end{document}

%% file: sections/introduction.tex
\section{Introduction}
\label{sec:intro}

Photonic crystal fibers (PCFs) are a cornerstone of modern photonics, enabling applications from ultra-low-loss telecom transmission and high-power laser delivery to biochemical sensing and quantum-state preservation~\cite{gan2023dispersion,nanophotonic_survey2025,ren2025unbalanced,wang2025towards}. Unlike conventional step-index fibers, PCFs derive their guiding properties from a microstructured cladding composed of periodic arrangements of air holes, tubular elements, or high-index rods, whose geometry governs confinement loss, chromatic dispersion, and modal characteristics through Maxwell's equations~\cite{zhang2025high,chen2022cost}. Designing a PCF that meets target specifications therefore demands a tightly coupled \emph{inverse-design} loop, in which the engineer inspects a cross-section micrograph, infers optical properties from the mode-field pattern, diagnoses performance bottlenecks, proposes geometric modifications, and validates them by electromagnetic simulation. This loop is currently manual and expert-intensive, with each design iteration taking hours of finite-difference time-domain (FDTD) simulation combined with trial-and-error parameter tuning.

Vision-language models (VLMs) offer a tantalizing alternative. Trained on internet-scale multimodal data, models such as GPT-4o, Gemini, and Qwen-VL have achieved remarkable visual reasoning in natural-image domains~\cite{mmstar2024,mathvista2023}. More recently, researchers have begun deploying VLMs as autonomous agents for scientific workflows~\cite{liu2025mimicking,yamada2025ai,pang2026sci}, from data-driven discovery~\cite{chen2025scienceagentbench} and interactive scientific experiments~\cite{sun2026scienceboard} to physical-law inference~\cite{zheng2026newtonbench} and data-science automation~\cite{jing2025dsbench}. A natural yet unexplored question arises, namely \emph{can VLMs handle the precision and physics-awareness required for real-world photonic inverse design?}

Answering this question requires a benchmark that links visual perception to physical design. Existing VLM benchmarks focus on natural images and general scientific knowledge~\cite{mmstar2024,mathvista2023,geobench2025}, while machine-learning approaches to PCF design bypass visual reasoning entirely by operating on tabular parameter vectors~\cite{gan2023dispersion,yao2023inverse,li2023deep}. No benchmark evaluates the step from \emph{seeing} a fiber cross-section to \emph{proposing a physically sound design modification}. Moreover, photonic inverse design poses several unique challenges for VLMs: (i) \textbf{fine-grained visual parsing}, distinguishing 18 structurally diverse fiber families, some of which share superficial visual similarities, (ii) \textbf{quantitative physical reasoning}, mapping subtle mode-field patterns (often near-black images with $\mu$W-level intensity) to confinement loss spanning 16 orders of magnitude, and (iii) \textbf{actionable design decisions}, translating visual understanding into parameter modifications grounded in anti-resonance theory, photonic bandgap physics, or index-guiding principles, not generic pattern matching.

\begin{figure}[!htbp]
    \centering
    \includegraphics[width=1\textwidth]{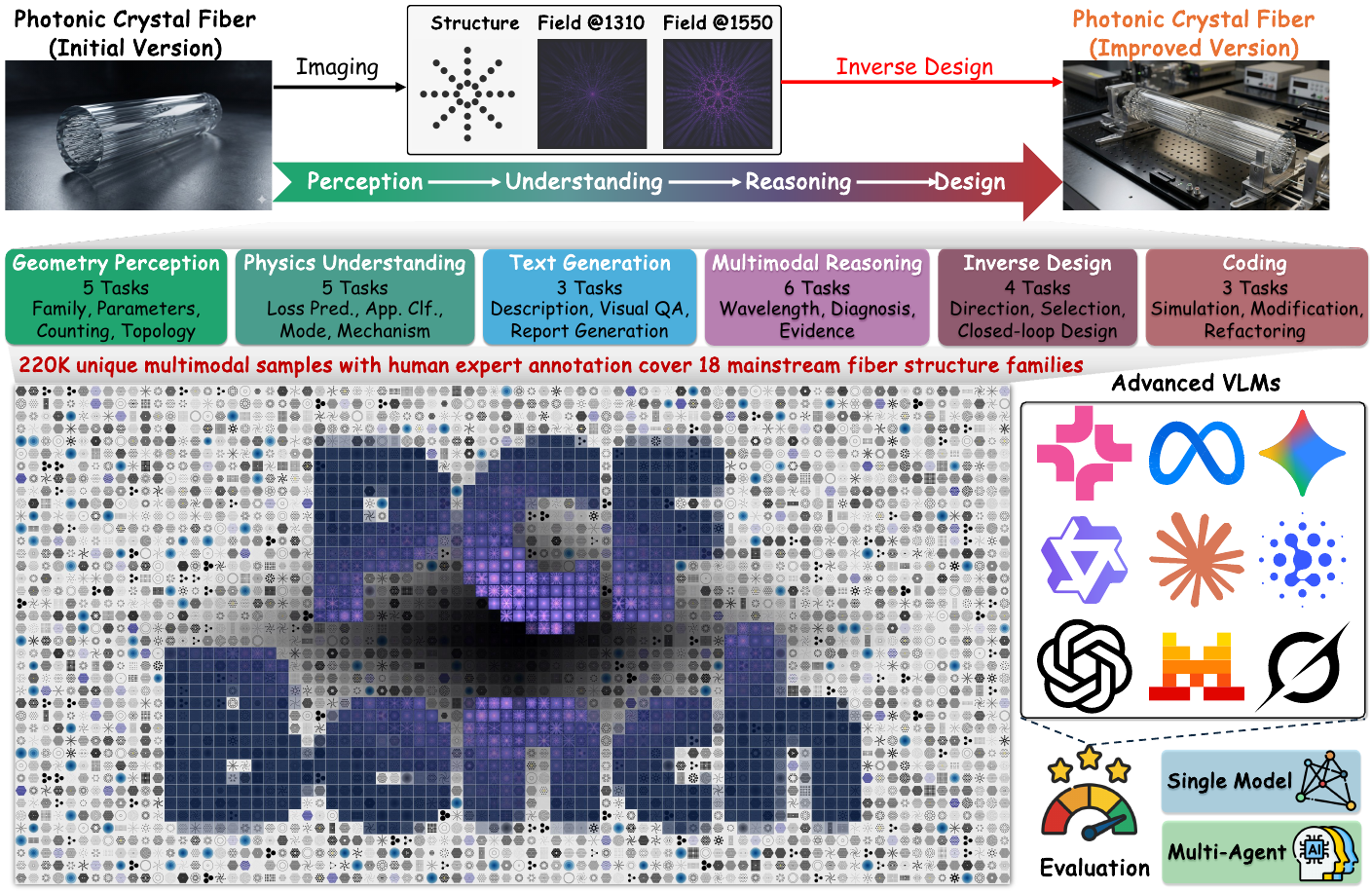}
    \vspace{-6pt}
    \caption{\small \textbf{PCFBench overview.} PCFBench formulates PCF inverse design as a multi-stage, multimodal reasoning problem, where complementary visual inputs, geometric cross-sections and mode fields, are processed through a sequential pipeline of \emph{perception}, \emph{understanding}, \emph{reasoning}, and \emph{design}. The benchmark comprises 26 tasks organized into a six-dimensional hierarchy spanning geometry perception, physics understanding, text generation, multi-modal reasoning, inverse-design decisions, and code generation, arranged along an increasing difficulty gradient that mirrors real-world workflows. Built on over 220K multimodal samples with human expert annotations across 18 fiber families, PCFBench supports both single-model and multi-agent evaluation.}
    \label{fig:overview}
    \vspace{-3pt}
\end{figure}

To address these challenges, we introduce \textbf{PCFBench}, the first comprehensive benchmark for evaluating VLMs on the perception, reasoning, and decision steps of this inverse-design loop. As shown in Fig.~\ref{fig:overview}, PCFBench comprises 26 tasks organized along a six-dimensional evaluation hierarchy that reflects the real-world design workflow, progressing from geometry perception (e.g., family classification and parameter estimation), through \emph{physics understanding} and \emph{text generation}, to \emph{multi-modal reasoning} and ultimately \emph{inverse-design decisions} and \emph{code generation}. The benchmark is constructed from over 220K FDTD simulation samples spanning 18 structurally diverse fiber families, each paired with structure and mode-field images at multiple operating wavelengths and enriched with multi-level expert annotations. Our main contributions are:
\begin{itemize}
    \item \textbf{A Unified Benchmark for Photonic Design.} We introduce PCFBench, a comprehensive benchmark that unifies a six-dimensional evaluation hierarchy, physics-aware protocols, and a dual evaluation paradigm, spanning 26 tasks from geometry perception to inverse-design decisions and code generation, and enabling both single-task and multi-agent evaluation for analyzing inter-stage dependencies and error propagation.

    \item \textbf{A Large-Scale, Expert-Annotated, Physics-Grounded Dataset.} PCFBench is constructed from over 220K FDTD simulation samples across 18 structurally diverse fiber families, each paired with structure and mode-field images and enriched with multi-level human expert annotations, providing high-fidelity supervision for both perception and design tasks.

    \item \textbf{New Empirical Insights into VLM Limitations for Inverse Design.} Extensive evaluation of 41 frontier VLMs and multi-agent system paradigms reveals a persistent \emph{perception--design gap} across model scales and coordination strategies, highlighting the need for explicit physical grounding beyond visual pattern matching.
\end{itemize}

%% file: sections/related_work.tex
\section{Related Work}
\label{sec:related}

\paragraph{Photonic Inverse Design.} Photonic device design is traditionally posed as optimization under Maxwell's equations~\cite{chen2025pcf,yuan2022efficient}, with candidates evaluated by FDTD or eigenmode solvers. Such approaches are accurate but costly and hard to scale, since they rely on large parameter sweeps or iterative optimization. Learning-based alternatives include deep networks for property prediction~\cite{gan2023dispersion,peurifoy2018nanophotonic,chen2023collaborative,ren2025unbalanced,ren2024distributed,wang2025towards}, LSTM-based inverse design~\cite{yao2023inverse}, reinforcement learning~\cite{li2023deep}, and physics-informed neural networks that embed Maxwell's equations in training~\cite{raissi2019physics}. Recent surveys cover surrogate modeling, adjoint-based optimization, and generative design~\cite{nanophotonic_survey2025}. The foundation model OptoGPT generates thin-film designs from target spectra~\cite{ma2024optogpt}, LLMs have been applied to photonic-integrated-circuit design~\cite{wu2025picbench}, and agentic frameworks autonomously model and optimize metamaterials~\cite{lu2025agentic} or run propose-simulate-verify loops for photonic devices~\cite{kharel2026autonomous}. Most of these approaches operate on structured parameters and treat inverse design as numerical mapping, whereas practical design relies on visual inspection of structures and mode fields. This gap raises a complementary question: can general-purpose VLMs reason about photonic design from images, without task-specific training?

\paragraph{Vision-Language Models.}
VLMs, trained on internet-scale image-text pairs, have shown strong visual reasoning capabilities across diverse domains~\cite{mmstar2024,mathvista2023,feng2025taming,feng2026visual,chen2025restyled}, and benchmarks such as MathVista~\cite{mathvista2023} and MMStar~\cite{mmstar2024}, along with domain-specific extensions to geospatial analysis~\cite{geobench2025}, medical imaging~\cite{chen2024gmai}, chart understanding~\cite{masry2022chartqa}, and visualization literacy~\cite{pandey2025benchmarking}, demonstrate non-trivial spatial and quantitative reasoning. However, these benchmarks focus on natural-image domains where human intuition transfers readily, and whether such capabilities extend to the fine-grained, physics-governed visual patterns of photonic microstructures remains unclear. PCFBench is designed to address this gap by evaluating the perception of photonic geometry together with reasoning about the underlying physics and acting on it for inverse design.

\paragraph{Scientific Problem Solving Benchmarks.}
Recent work has begun evaluating LLMs and VLMs as autonomous agents in scientific workflows. Benchmarks such as ScienceAgentBench~\cite{chen2025scienceagentbench}, ScienceBoard~\cite{sun2026scienceboard}, DSBench~\cite{jing2025dsbench}, and NewtonBench~\cite{zheng2026newtonbench} place agents in data-driven discovery, analysis, and simulation settings, revealing limited performance on complex multi-step tasks, while InnoGym~\cite{zhang2026innogym} extends evaluation beyond correctness to assess innovation. These studies expose a key limitation: current models struggle with long-horizon reasoning requiring domain-specific physical intuition, and most benchmarks focus on isolated tasks, leaving error propagation in sequential workflows underexplored. PCFBench addresses this gap by introducing a domain-specific benchmark for photonic inverse design that evaluates both VLMs and multi-agent systems along a perception-to-design pipeline, enabling systematic analysis of inter-stage dependencies and error cascades.

%% file: sections/pcfbench.tex
\section{PCFBench}
\label{sec:pcfbench}

\begin{figure}[!htbp]
    \centering
    \includegraphics[width=1\textwidth]{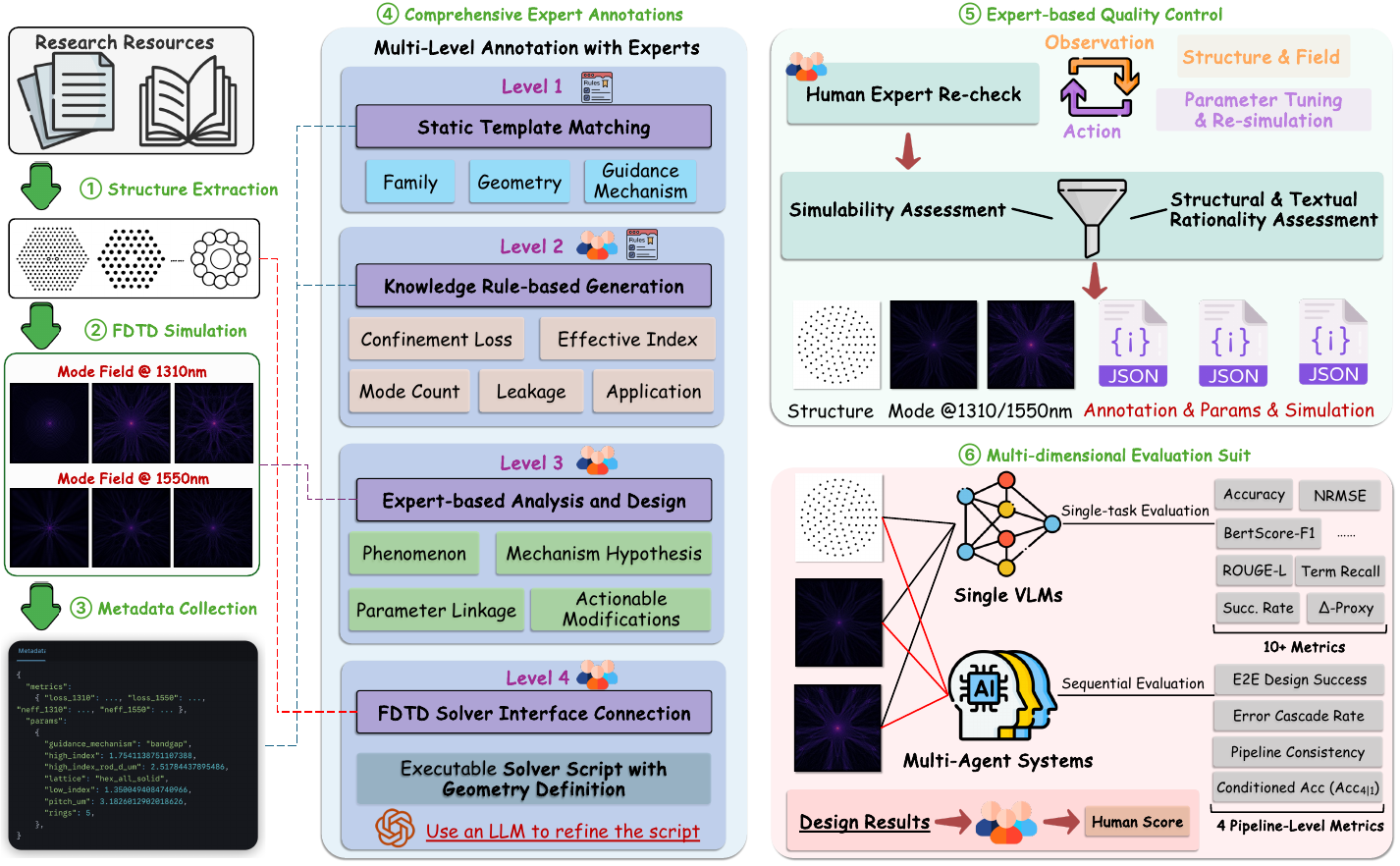}
    \vspace{-6pt}
    \caption{PCFBench is constructed through a unified pipeline: \textcolor{green!60!black}{\textbf{\ding{172}}} structure extraction from diverse fiber geometries, \textcolor{green!60!black}{\textbf{\ding{173}}} FDTD simulation of multi-wavelength mode fields, \textcolor{green!60!black}{\textbf{\ding{174}}} aggregation of geometric and physical metadata, \textcolor{green!60!black}{\textbf{\ding{175}}} multi-level expert annotation, \textcolor{green!60!black}{\textbf{\ding{176}}} expert quality control via re-simulation, and \textcolor{green!60!black}{\textbf{\ding{177}}} a multi-dimensional evaluation suite for single-model and multi-agent assessment.}
    \label{fig:pipeline}
\end{figure}

\paragraph{Overview.}
PCFBench is a physics-grounded benchmark for evaluating VLMs across the \emph{perception-to-design chain} of photonic inverse design, rather than isolated perception tasks. It is constructed through a six-stage pipeline (\textbf{Fig.~\ref{fig:pipeline}}) that pairs large-scale FDTD simulation with sustained \emph{human-expert involvement}. Stages~\textcolor{green!60!black}{\textbf{\ding{172}}}--\textcolor{green!60!black}{\textbf{\ding{174}}} curate 18 structurally diverse fiber families, run a unified Meep FDTD solver at two telecom wavelengths (1310 and 1550\,nm), and aggregate geometric and physical metadata into a 220K-sample corpus. Stage~\textcolor{green!60!black}{\textbf{\ding{175}}} then layers \emph{four levels of expert annotation} on top of every sample: L1 templated geometry descriptors, L2 rule-grounded physics QA (loss, $n_\text{eff}$, mode count, leakage, application), L3 expert-authored mechanism hypotheses with parameter--performance linkages and actionable design suggestions, and L4 executable solver scripts that close the simulation--annotation loop. Stage~\textcolor{green!60!black}{\textbf{\ding{176}}} subjects each annotation to expert re-checking, structural and textual rationality assessments, and re-simulation verification that removes inconsistent samples. Built on this curated corpus, Stage~\textcolor{green!60!black}{\textbf{\ding{177}}} defines 26 tasks (Sec.~\ref{sec:tasks}) organized along a \emph{perception--understanding--reasoning--design} hierarchy (\textbf{Fig.~\ref{fig:overview}}) and a multi-dimensional evaluation suite (Sec.~\ref{sec:eval_protocol}) that covers both single-VLM and multi-agent evaluation.

\subsection{Dataset Construction}
\label{sec:dataset}

\begin{wrapfigure}{r}{0.575\linewidth}
  \vspace{-3.4em}
  \centering
  \includegraphics[width=\linewidth]{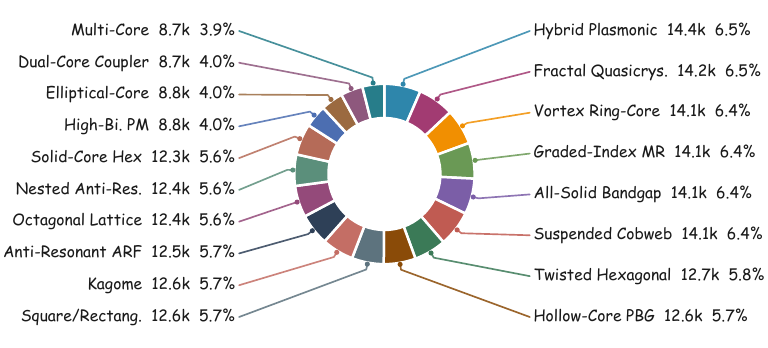}
  \vspace{-1.7em}
  \caption{Dataset composition of PCFBench. The 220K PCF samples span 18 canonical families. Each label gives the family's sample count and proportion.}
  \label{fig:dataset_composition}
\end{wrapfigure}

\begin{figure}[!htbp]
  \centering
  \includegraphics[width=1\textwidth]{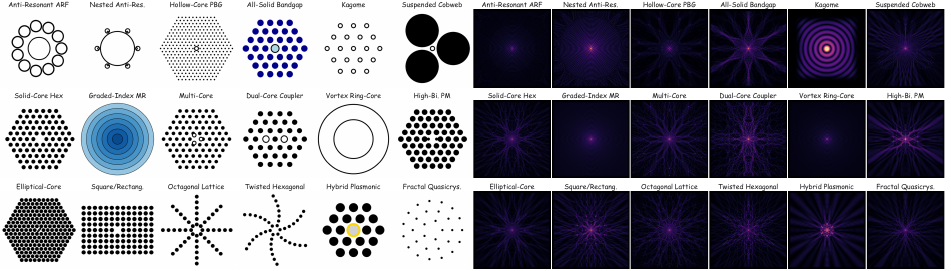}
  \vspace{-14pt}
\caption{Representative structures (left) and corresponding mode fields at $\lambda = 1550$\,nm (right) for all families. The paired views reveal distinct guidance mechanisms: hollow-core families exhibit dark cores with bright anti-resonant rings, solid-core families show compact guided modes, and specialty geometries (plasmonic, vortex, multi-core) produce characteristic field patterns. See Appendix~\ref{app:cases} for additional exemplars per family.}
\vspace{-6pt}
  \label{fig:family_overview}
  \vspace{-6pt}
\end{figure}

PCFBench is constructed from a large-scale FDTD simulation corpus spanning 18 structurally diverse fiber families, including conventional telecom designs (e.g., solid-core hexagonal lattice, high-birefringence polarization-maintaining, multi-core), hollow-core anti-resonant fibers (e.g., single-ring ARF, nested ARF, kagome), and specialty geometries (e.g., plasmonic, vortex ring-core, fractal quasicrystal). The per-family sample distribution is summarized in \textbf{Fig.~\ref{fig:dataset_composition}}, with paired structure and mode-field exemplars for every family shown in \textbf{Fig.~\ref{fig:family_overview}}. Further details on structural motifs, guidance mechanisms, and applications are given in Appendix~\ref{app:cases}. Each of the 220K samples contains four components: (i) a geometric parameter vector $\boldsymbol{\theta} = \{\Lambda, d, t, \ldots\}$ describing pitch, hole diameter, strut thickness, core size, cladding ring count, and family-specific attributes, (ii) a structural image at $336\times 336$ resolution representing the refractive-index distribution, (iii) mode-field images simulated by FDTD at two telecom wavelengths, and (iv) multi-level expert annotations supervising geometry description (L1), mechanism and physics QA (L2), parameter--performance reasoning with actionable design suggestions (L3), and executable Meep code (L4). All simulations share a unified Meep~FDTD configuration (boundary conditions, resolution, source profile, and run time, listed in \textbf{Tab.~\ref{tab:fdtd_config}} of Appendix~\ref{app:fdtd_setup}), ensuring comparability across families.

\subsection{Task Taxonomy}
\label{sec:tasks}

\input{Tables/tab_task_overview}

PCFBench organizes \textbf{26 tasks} into six categories of increasing cognitive demand, forming a \emph{perception--understanding--reasoning--design} hierarchy (\textbf{Tab.~\ref{tab:task_overview}}). The hierarchy begins with \textbf{Geometry Perception} (5 tasks), which requires a VLM to parse the fiber microstructure from a single cross-section image. \textbf{Physics Understanding} (5 tasks) then requires the model to infer physical quantities, mainly from mode-field images, including confinement-loss prediction at two telecom wavelengths, application and mode classification, and mechanism reasoning. \textbf{Text Generation} (3 tasks) evaluates open-ended scientific communication via geometry description, conversational QA, and structured report generation. \textbf{Multi-modal Reasoning} (6 tasks) requires integrating evidence from both structure and field images for wavelength comparison, mechanism diagnosis, and evidence-based explanation, each evaluated under \emph{Paired} and \emph{Field-only} settings to isolate the contribution of structural context. \textbf{Inverse-Design Decision} (4 tasks) is the most challenging category. Each iteration of an objective-driven inverse-design loop involves three decisions, namely which parameter to change (parameter selection), in which direction (design-direction prediction), and by how much (closed-loop improvement), and these tasks score each decision against family-specific physics rules (Appendix~\ref{app:physics_gt}). Sec.~\ref{sec:case_study} re-simulates the proposals to measure their effect on loss. These decision tasks are the central evaluation target of the benchmark, and the preceding categories diagnose where a design failure originates. Finally, \textbf{Code Generation} (3 tasks) requires the model to produce executable Meep FDTD scripts, modify parameters, and refactor code. Task definitions, prompts, rules, and metrics are in Appendix~\ref{app:tasks}.

\subsection{Multi-dimensional Evaluation Suite}
\label{sec:eval_protocol}

PCFBench evaluates VLMs along two complementary axes. First, a \textbf{single-VLM track} assesses each model independently on every task using category-aligned metrics, with the full evaluation protocol in Appendix~\ref{app:tasks_shared}. Second, a \textbf{multi-agent track} decomposes the end-to-end design workflow into a four-stage Sequential Pipeline and benchmarks 12 additional multi-agent coordination strategies under identical inputs (Appendix~\ref{app:mas_metrics}). To ground the inverse-design decision tasks in physics rather than surface-level heuristics, direction and closed-loop ground truth are derived from family-specific expert rules that account for non-monotonic regimes such as anti-resonant strut bands (Appendix~\ref{app:physics_gt}). A composite Human Score evaluated by two photonics experts across six dimensions (Appendix~\ref{app:human_eval}) complements automated metrics by assessing both correctness and physical plausibility. All samples are newly simulated and previously unpublished, so none can appear in model pretraining data. A side-by-side comparison with prior benchmarks is in Appendix~\ref{app:bench_comparison}.

%% file: Tables/tab_task_overview.tex
\begin{wraptable}{r}{0.52\textwidth}
  \vspace{-2.2em}
  \centering
\caption{\small Overview of the 26 PCFBench tasks (hierarchy in \textbf{Fig.~\ref{fig:overview}}). ``S'' = structure image, ``F'' = mode-field image.}
  \label{tab:task_overview}
  \resizebox{\linewidth}{!}{%
  \begin{tabular}{ c c l l l l }
  \toprule
  \textbf{Hierarchy} & \textbf{\#} & \textbf{Task} & \textbf{Category} & \textbf{Img} & \textbf{Metric} \\
  \midrule
  \multirow{5}{*}{\parbox{1.6cm}{\centering\textcolor[HTML]{10B981}{\faEye}\\[1pt]\scriptsize\textbf{Percep.}}}
    & 1  & Family Classif.          & Geo.~Percep. & S         & Acc.\,$\uparrow$ \\
    & 2  & Param.~Pred.~(Pitch)     & Geo.~Percep. & S         & NRMSE\,$\downarrow$ \\
    & 3  & Param.~Pred.~(Hole D)    & Geo.~Percep. & S         & NRMSE\,$\downarrow$ \\
    & 4  & Structure Counting       & Geo.~Percep. & S         & Exact Acc.\,$\uparrow$ \\
    & 5  & Topology Recog.          & Geo.~Percep. & S         & Acc.\,$\uparrow$ \\
  \midrule
  \multirow{5}{*}{\parbox{1.6cm}{\centering\textcolor[HTML]{0D9488}{\faBrain}\\[1pt]\scriptsize\textbf{Underst.}}}
    & 6  & Loss Pred.~(1310\,nm)    & Phys.~Und.   & F$_{1310}$ & Log-MAE\,$\downarrow$ \\
    & 7  & Loss Pred.~(1550\,nm)    & Phys.~Und.   & F$_{1550}$ & Log-MAE\,$\downarrow$ \\
    & 8  & Application Classif.     & Phys.~Und.   & F          & Acc.\,$\uparrow$ \\
    & 9  & Mode Classif.            & Phys.~Und.   & F          & Acc.\,$\uparrow$ \\
    & 10 & Mechanism Reasoning      & Phys.~Und.   & S          & Kwd.~Rec.\,$\uparrow$ \\
  \midrule
  \multirow{9}{*}{\parbox{1.6cm}{\centering\textcolor[HTML]{4F46E5}{\faLightbulb}\\[1pt]\scriptsize\textbf{Reason.}}}
    & 11 & Geometry Descr.          & Text Gen.    & S            & BS-F1\,$\uparrow$ \\
    & 12 & Conversational QA        & Text Gen.    & F            & BS-F1\,$\uparrow$ \\
    & 13 & Report Generation        & Text Gen.    & S+F          & Compl.\,$\uparrow$ \\
    \cmidrule(l){2-6}
    & 14 & Wavelength Comp.~(P)     & MM Reas.     & S+F$_{2\lambda}$ & Acc.\,$\uparrow$ \\
    & 15 & Wavelength Comp.~(F)     & MM Reas.     & F$_{2\lambda}$   & Acc.\,$\uparrow$ \\
    & 16 & Mech.~Diag.~(Paired)     & MM Reas.     & S+F          & M-F1\,$\uparrow$ \\
    & 17 & Mech.~Diag.~(Field)      & MM Reas.     & F            & M-F1\,$\uparrow$ \\
    & 18 & Evid.~Expl.~(Paired)     & MM Reas.     & S+F          & BS-F1\,$\uparrow$ \\
    & 19 & Evid.~Expl.~(Field)      & MM Reas.     & F            & BS-F1\,$\uparrow$ \\
  \midrule
  \multirow{7}{*}{\parbox{1.6cm}{\centering\textcolor[HTML]{DC2626}{\faPencilRuler}\\[1pt]\scriptsize\textbf{Design}}}
    & 20 & Parameter Selection      & Inv.~Decision & S+F          & Acc.\,$\uparrow$ \\
    & 21 & Design Dir.~(Pitch)      & Inv.~Decision & S+F          & Acc.\,$\uparrow$ \\
    & 22 & Design Dir.~(Hole D)     & Inv.~Decision & S+F          & Acc.\,$\uparrow$ \\
    & 23 & Closed-Loop Improv.      & Inv.~Decision & S+F          & Succ.\,$\uparrow$ \\
    \cmidrule(l){2-6}
    & 24 & Sim.~Code Generation     & Code Gen.    & S            & Qual.\,$\uparrow$ \\
    & 25 & Parameter Modification   & Code Gen.    & S            & Syn.\,$\uparrow$ \\
    & 26 & Code Refactoring         & Code Gen.    & S            & Qual.\,$\uparrow$ \\
  \bottomrule
  \end{tabular}%
  }%
\end{wraptable}

%% file: sections/experiments.tex
\section{Experiments}
\label{sec:experiments}

\subsection{Can Individual VLMs Bridge Perception to Design?}
\label{sec:exp_single}

\paragraph{Implementation Details.}
We evaluate 41 VLMs from ten vendors, spanning 4B to 400B+ parameters (full catalog in Appendix~\ref{app:models}). 
The suite covers a diverse set of open-weight families, including Qwen-VL~\cite{wang2024qwen2,bai2025qwen25vltechnicalreport,bai2025qwen3vltechnicalreport,qwenteam2026qwen35omnitechnicalreport}, Gemma-3~\cite{team2025gemma}, Mistral~\cite{ministral-2024,agrawal2024pixtral12b,mistrallarge-2024}, Llama~\cite{grattafiori2024llama,llama4-2025}, and Molmo~\cite{deitke2025molmo}, as well as proprietary systems from OpenAI~\cite{hurst2024gpt,openai2025gpt41,singh2025openai,openai2025o3}, Anthropic~\cite{anthropic2024claude35,anthropic2025claude45,anthropic2025opus45}, Google~\cite{comanici2025gemini,gemini3-2025}, xAI~\cite{grok4-2025}, and Zhipu~\cite{hong2025glm}. 
All models are queried via their respective APIs with a fixed temperature of $0.1$ and task-specific token limits. 
Further details of the evaluation protocol are provided in Appendix~\ref{app:tasks_shared}.

\input{Tables/tab_main_overview}

\textbf{Tab.~\ref{tab:overview}} presents the leaderboard across all six evaluation dimensions. Per-metric breakdowns appear in Appendix~\ref{app:detailed_results} (Tabs.~\ref{tab:geo_phys}--\ref{tab:inv_code}). Several main findings emerge.

\noindent\textcolor[HTML]{F59E0B}{\faLightbulb}~\textit{\textbf{Perception is the easy part, but design is not.}} Gemini-3-Flash performs strongly on geometry perception, yet no model reaches 50\% on topology recognition, indicating a gap in fine-grained lattice understanding. More importantly, perception correlates weakly with inverse-design success: Qwen3.5-122B-A10B leads in topology accuracy but achieves only 35.5\% closed-loop success, while the smaller Qwen2.5-VL-7B reaches 63.8\%. This \emph{perception--design gap} holds across models, revealing a fundamental bottleneck in translating visual understanding into physics-aware action.

\noindent\textcolor[HTML]{F59E0B}{\faLightbulb}~\textit{\textbf{Bigger is not always better.}} Scaling provides inconsistent returns. Within the Qwen-VL family, scaling from 7B to 235B improves topology recognition by 25\,pp but \emph{degrades} closed-loop success by 47\,pp. GPT-4o-mini outperforms both GPT-4o and GPT-5 on closed-loop design, suggesting that instruction-tuning distribution matters more than raw scale for physically grounded tasks.

\noindent\textcolor[HTML]{F59E0B}{\faLightbulb}~\textit{\textbf{Text generation and code syntax saturate.}} BERTScore-F1 for geometry description clusters in a narrow 0.27--0.36 band across models. For simulation code, syntax and parameter-modification completeness are near-saturated. End-to-end executability is in Appendix~\ref{app:res_inv_code}. VLMs write syntactically valid Meep scripts even when they cannot reason about the encoded physics.

\noindent\textcolor[HTML]{F59E0B}{\faLightbulb}~\textit{\textbf{Paired prompts expose majority-class shortcuts.}} Several models score perfect macro-F1 on \emph{Field-only} mechanism diagnosis by defaulting to the majority class, while scoring near zero on the \emph{Paired} variant with all seven options. PCFBench's dual-prompt design (Paired vs.\ Field-only) turns this shortcut into a diagnostic signal, since any model whose Field-only score vastly exceeds its Paired score is exploiting label imbalance, not reasoning about physics.

\subsection{Does Multi-Agent Coordination Close the Gap?}
\label{sec:exp_mas}

\paragraph{Implementation Details.}
We benchmark 13 MAS on the inverse-design track. These comprise 12 MAS methods (Vanilla, CoT~\cite{wei2022chain}, Self-Consistency~\cite{wang2022self}, LLM Debate~\cite{du2023improving}, MAD~\cite{liang2024encouraging}, CAMEL~\cite{li2023camel}, DyLAN~\cite{liu2023dynamic}, MacNet~\cite{qian2025scaling}, AgentVerse~\cite{chen2024agentverse}, MAV~\cite{lifshitz2025multi}, AutoGen~\cite{wu2023autogen}, EvoMAC~\cite{hu2025self}) together with the PCFBench Sequential Pipeline, a four-stage domain-decomposed baseline (Perception $\to$ Understanding $\to$ Reasoning $\to$ Design) evaluated under Oracle/Cascaded modes. All methods are implemented in the unified MASLab~\cite{ye2025maslab} framework and receive identical queries under three backbones (GPT-4o-mini, Qwen-VL-Max, and Gemini-3.1-Lite). More details are in Appendix~\ref{app:mas_metrics}. \textbf{Tab.~\ref{tab:mas_methods}} shows three findings on backbone dependence, agent count, and domain decomposition.

\input{Tables/tab_mas_methods}

\noindent\textcolor[HTML]{F59E0B}{\faLightbulb}~\textit{\textbf{The best coordination strategy is backbone-dependent.}} MAD (judge-mediated debate) leads on GPT-4o-mini (CL\,=\,.731) but drops to .411 on Qwen-VL-Max, where even a single Vanilla call achieves .580. Conversely, EvoMAC, the worst method on GPT-4o-mini (CL\,=\,.232), becomes the \emph{best} on Qwen-VL-Max (.732). No coordination pattern universally dominates, and the interaction between backbone capability and agent topology is the primary determinant of design quality.

\noindent\textcolor[HTML]{F59E0B}{\faLightbulb}~\textit{\textbf{More agents $\neq$ better design.}} On GPT-4o-mini, the two-agent AutoGen (CL\,=\,.680) outperforms DyLAN, MacNet, MAV, and EvoMAC, which all use more agents. On Gemini-3.1-Lite, the single-call Vanilla baseline also ties or beats five multi-agent methods in Human Score. Extra agents amplify shared biases rather than correcting them when the backbone lacks domain knowledge.

\noindent\textcolor[HTML]{F59E0B}{\faLightbulb}~\textit{\textbf{Domain decomposition outperforms generic MAS, but error cascade is the price.}} The Sequential Pipeline (Cascaded) consistently surpasses generic MAS baselines on PSel because its explicit perception stage forces family identification before design decisions are made. However, \textbf{Tab.~\ref{tab:mas_stages}} in Appendix~\ref{app:mas_diagnostic} shows that upstream perception errors propagate to downstream design at a cascade rate exceeding 60\%, suggesting that closing the perception--design gap needs physics-grounded fine-tuning or simulation-in-the-loop verification beyond better coordination.

\subsection{Analysis and Discussion}
\label{sec:analysis}

\begin{figure}[tbh]
  \centering
  \includegraphics[width=1\textwidth]{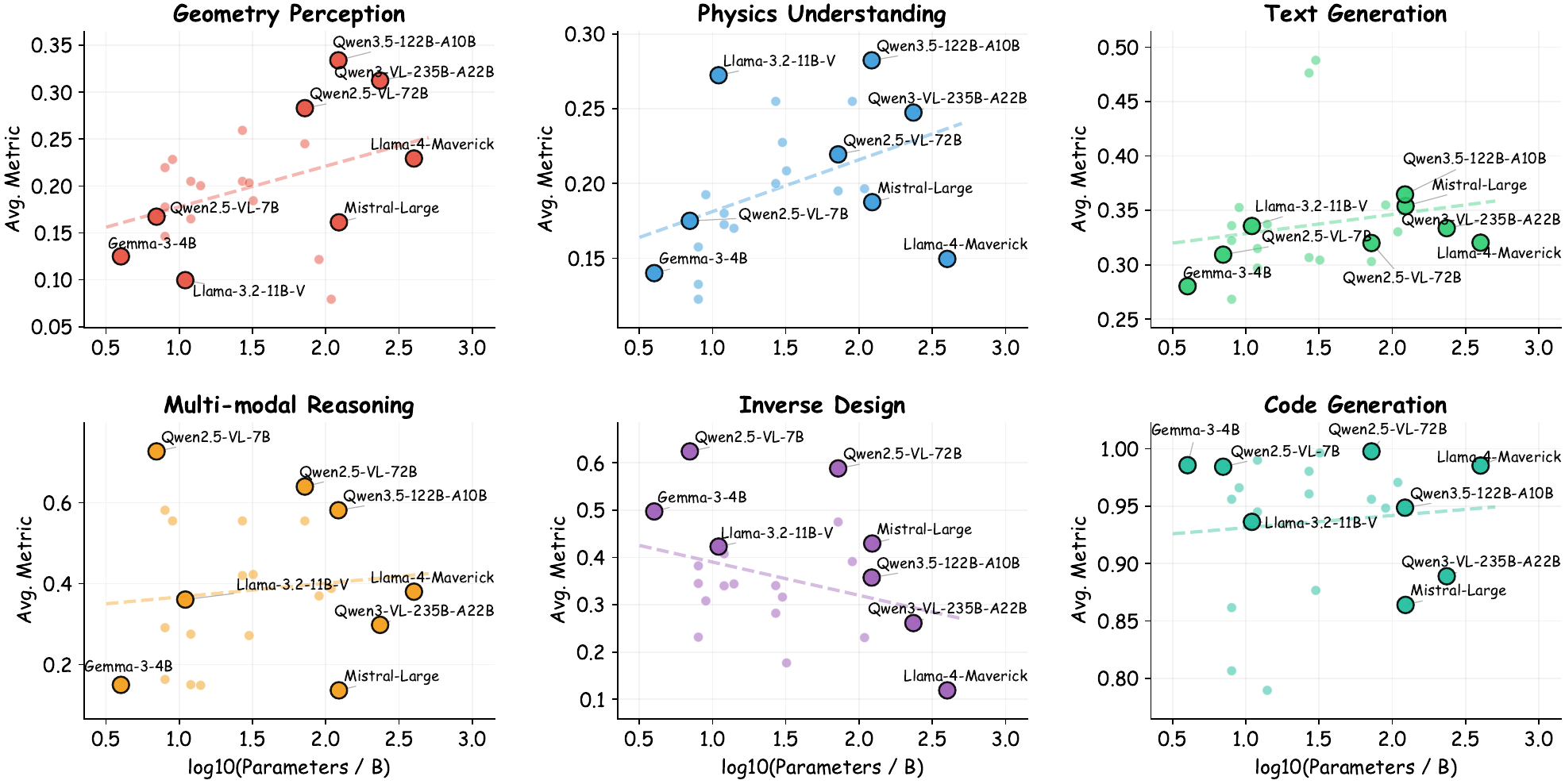}
  \vspace{-6pt}
\caption{Scaling curves: parameter count vs.\ category-averaged performance.}
  \label{fig:scaling}
  \vspace{-6pt}
\end{figure}

\paragraph{Does Scaling Help?}
\textbf{Fig.~\ref{fig:scaling}} plots log parameter count against category-averaged performance for all open-weight VLMs. Geometry perception improves with scale ($R^2 > 0.4$), whereas inverse-design decisions remain flat or even decline. For example, Qwen3.5-122B-A10B attains the highest topology accuracy yet sits well below Qwen2.5-VL-7B on closed-loop success, despite a $\sim$15$\times$ parameter advantage. This decoupling suggests that the perception--design gap is not primarily a scaling issue but a deeper deficiency in physical reasoning that more parameters alone do not resolve.

\input{Tables/tab_fewshot_main}
\paragraph{Is the Gap Visual or Physical?}
\textbf{Tab.~\ref{tab:fewshot_main}} shows that in-context exemplars lift perception but not design: eight exemplars raise family accuracy by up to .161, while closed-loop success moves by at most .009. GPT-4o-mini makes the decoupling explicit, keeping the highest closed-loop success of the benchmark (about .80) while recognizing fewer than a third of the families. Families farther from natural images in SigLIP~\cite{zhai2023sigmoid} space are also recognized less accurately (Spearman $\rho \approx -.6$), yet this distance is unrelated to closed-loop success. Better perception is thus neither necessary nor sufficient for better design, and the gap stems from physical reasoning, not visual unfamiliarity. SigLIP details are in Appendix~\ref{app:additional_analysis}.

\begin{figure}[!htbp]
\centering
\vspace{-6pt}
\begin{minipage}[b]{0.385\textwidth}
\input{Tables/tab_sem_main}\end{minipage}\hfill
\begin{minipage}[b]{0.595\textwidth}
\centering
\includegraphics[width=\linewidth]{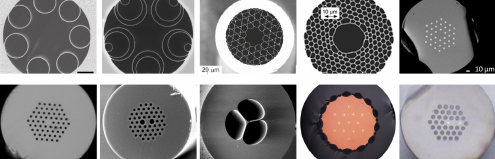}
\caption{Micrographs of fabricated PCFs.}
\label{fig:sem_examples}
\end{minipage}
\vspace{-4pt}
\end{figure}

\paragraph{Do Rankings Transfer to Real Fibers?}
\textbf{Tab.~\ref{tab:sem_main}} shows that model rankings carry over from simulated to fabricated cross-sections. We collect 120 SEM micrographs of fabricated fibers from the literature, covering 10 families, and \textbf{Fig.~\ref{fig:sem_examples}} shows one per family, from left to right and top to bottom: ARF~\cite{zhao2020mode}, Nested ARF~\cite{petrovich2025broadband}, Kagome~\cite{wheeler2017low}, HC-PBG~\cite{slavik2015ultralow}, All-Solid BG~\cite{caillaud2014photonic}, Solid-Core Hex~\cite{cardenas2013high}, Hi-Bi PM~\cite{zhang2018thermally}, Suspended Cobweb~\cite{catalano2023tuning}, Multi-Core~\cite{zhang2024novel}, and Dual-Core~\cite{zhao2018simultaneous} (optical images for the last two). On these micrographs, the six models keep exactly their Family Classification order on a matched simulated subset (Kendall $\tau = 1$), while every model loses some accuracy. Real micrographs make the task harder without changing which models are better.

\begin{figure}[!htbp]
  \centering
  \includegraphics[width=\textwidth]{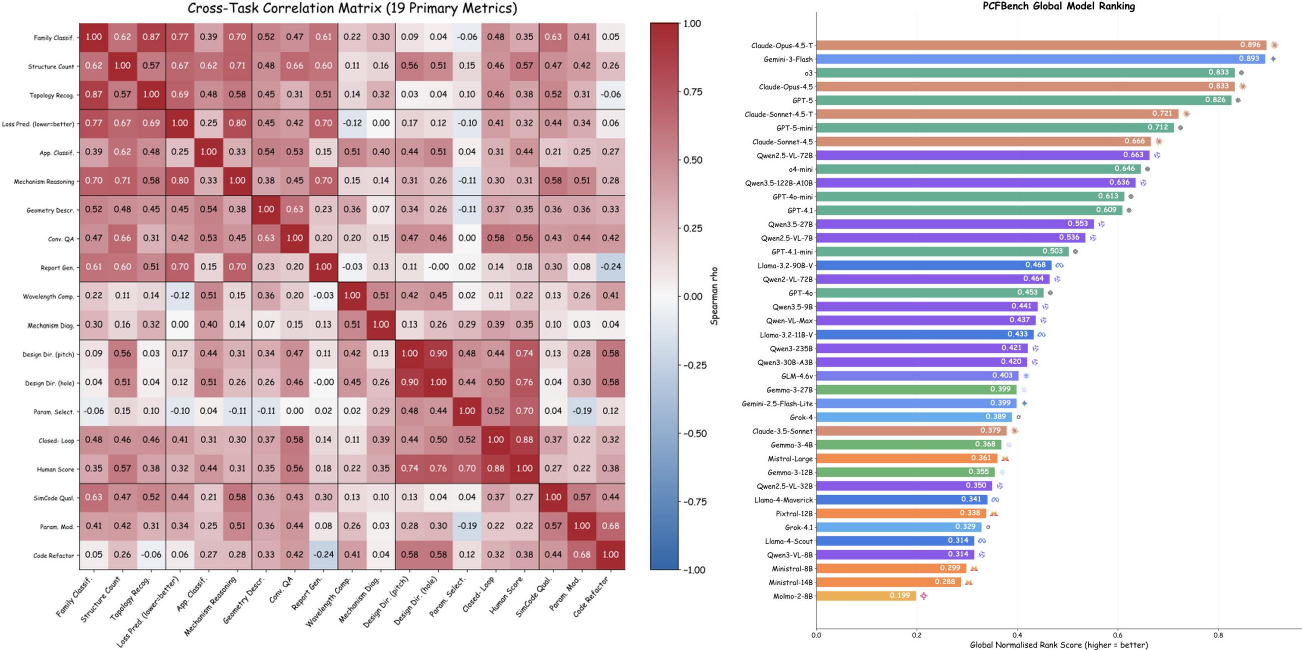}
  \vspace{-12pt}
\caption{\textbf{Left} Cross-task Spearman $\rho$. \textbf{Right} Global ranking (mean normalized rank, 19 metrics).}
  \label{fig:analysis}
  \vspace{-6pt}
\end{figure}

\paragraph{Cross-Task Correlations and Global Ranking.}
The Spearman correlation matrix (\textbf{Fig.~\ref{fig:analysis}}, left) shows that geometry tasks cluster tightly but exhibit weak correlation with inverse-design decisions, indicating that PCFBench captures complementary capability axes. Code generation appears largely decoupled from other categories, suggesting a distinct skill set beyond photonic reasoning. When all 19 metrics are aggregated into a normalized rank score (\textbf{Fig.~\ref{fig:analysis}}, right), frontier models with extended reasoning dominate the top tier (e.g., Claude-Opus-4.5-T, Gemini-3-Flash, o3), while Qwen2.5-VL-72B leads among open-weight models. Notably, Llama-4-Maverick (400B) ranks below several smaller models, indicating that scale alone is insufficient for photonic inverse design.

\paragraph{Beyond the Cross-Section.}
\textbf{Tab.~\ref{tab:propagation_main}} shows that propagation-level reasoning is harder. On two binary tasks over 600 index-guiding fibers, classifying the dispersion regime at 1550\,nm and the broadening regime of a 100\,fs pulse, no model exceeds .69 and .59, respectively, and errors cluster near the regime boundary. Models sort clear cases from coarse geometric cues but miss the balance of dispersion and nonlinearity, mirroring the transverse perception--design gap. Details are in Appendix~\ref{app:propagation}.

\input{Tables/tab_propagation_feedback}

\input{sections/case_study}

%% file: Tables/tab_main_overview.tex
\begin{table}[tbh]
  \vspace{-4pt}
\caption{PCFBench leaderboard: one representative metric per task across 6 dimensions (full results in Tabs.~\ref{tab:geo_phys}--\ref{tab:inv_code}). Top-3 per column are \rankkey{rankgold}{\textbf{Gold}} / \rankkey{ranksilver}{\underline{Blue}} / \rankkey{rankbronze}{Gray}.}
\vspace{-4pt}
\label{tab:overview}
\centering
\resizebox{\textwidth}{!}{%
\footnotesize
\setlength{\tabcolsep}{3pt}
\renewcommand{\arraystretch}{0.92}
\begin{tabular}{l r ccc ccc ccc cc ccccc ccc}
\toprule
& &
  \multicolumn{3}{c}{\textbf{Geometry}} &
  \multicolumn{3}{c}{\textbf{Physics}} &
  \multicolumn{3}{c}{\textbf{Text Gen.}} &
  \multicolumn{2}{c}{\textbf{Multi-modal}} &
  \multicolumn{5}{c}{\textbf{Inverse-Design Decision}} &
  \multicolumn{3}{c}{\textbf{Code Gen.}} \\
\cmidrule(lr){3-5}\cmidrule(lr){6-8}\cmidrule(lr){9-11}\cmidrule(lr){12-13}
\cmidrule(lr){14-18}\cmidrule(lr){19-21}
\textbf{Model} & \textbf{Size} &
  \rotatebox{70}{Fam. Acc} &
  \rotatebox{70}{Count Acc} &
  \rotatebox{70}{Topo. Acc} &
  \rotatebox{70}{Loss-1550 LMAE$\downarrow$} &
  \rotatebox{70}{App. Acc} &
  \rotatebox{70}{Mech. KR} &
  \rotatebox{70}{Geo.D BS-F1} &
  \rotatebox{70}{QA BS-F1} &
  \rotatebox{70}{Report Comp.} &
  \rotatebox{70}{WC Paired} &
  \rotatebox{70}{MD Paired} &
  \rotatebox{70}{Dir. Pitch} &
  \rotatebox{70}{Dir. Hole\,D} &
  \rotatebox{70}{P.Sel. Acc} &
  \rotatebox{70}{CL Succ.} &
  \rotatebox{70}{Human Score} &
  \rotatebox{70}{SimCode Qual.} &
  \rotatebox{70}{ParamMod Syn.} &
  \rotatebox{70}{Refactor Qual.} \\
\midrule
\model{gemma}{Gemma-3-4B}{team2025gemma} & 4B & .045 & .155 & .175 & 2.850 & .035 & .245 & .270 & .106 & .465 & .247 & .052 & .473 & .388 & \second{.651} & .477 & .42 & .982 & \first{1.00} & .975 \\
\model{qwen}{Qwen2.5-VL-7B}{bai2025qwen25vltechnicalreport} & 7B & .096 & .176 & .230 & 2.655 & .055 & .295 & .334 & .112 & .482 & \second{.758} & \first{.696} & .521 & \second{.730} & .608 & .638 & .53 & .953 & \first{1.00} & \first{1.00} \\
\model{qwen}{Qwen3-VL-8B}{bai2025qwen3vltechnicalreport} & 8B & .192 & .174 & .293 & 2.435 & .050 & .265 & .317 & .096 & .554 & .663 & \second{.500} & .226 & .151 & .222 & .327 & .27 & .949 & \first{1.00} & .919 \\
\model{mistral}{Ministral-8B}{ministral-2024} & 8B & .054 & .200 & .185 & 2.745 & .040 & .225 & .285 & .098 & .625 & .457 & .125 & .423 & .215 & .641 & .250 & .33 & .952 & .784 & .684 \\
\model{ai2}{Molmo-2-8B}{deitke2025molmo} & 8B & .093 & .194 & .246 & 2.895 & .030 & .215 & .285 & .095 & .425 & .305 & .021 & .355 & .255 & .475 & .295 & .31 & .885 & .875 & .825 \\
\model{meta}{Llama-3.2-11B-V}{grattafiori2024llama} & 11B & .031 & .248 & .020 & 2.491 & \first{.245} & .300 & .339 & .165 & .503 & .690 & .031 & \first{.750} & \first{.771} & .110 & .061 & .32 & .820 & \first{1.00} & .989 \\
\cmidrule(lr){1-21}
\model{gemma}{Gemma-3-12B}{team2025gemma} & 12B & .085 & .175 & .235 & 2.350 & .050 & .295 & .295 & .097 & .499 & .243 & .057 & .308 & .247 & .559 & .517 & .40 & .996 & \second{.994} & .980 \\
\model{mistral}{Pixtral-12B}{agrawal2024pixtral12b} & 12B & .133 & .143 & .339 & 2.485 & .055 & .305 & .305 & .105 & .535 & .425 & .125 & .325 & .245 & .485 & .305 & .33 & .945 & .975 & .915 \\
\model{mistral}{Ministral-14B}{ministral-2024} & 14B & .115 & .199 & .287 & 2.315 & .055 & .285 & .294 & .096 & .622 & .243 & .054 & .238 & .164 & \third{.650} & .324 & .33 & .950 & .645 & .774 \\
\model{gemma}{Gemma-3-27B}{team2025gemma} & 27B & .125 & .195 & .295 & 1.885 & .065 & .335 & .294 & .096 & .530 & .610 & \second{.500} & .286 & .279 & .305 & .258 & .28 & \third{.998} & \second{.994} & .949 \\
\model{qwen}{Qwen3-VL-30B-A3B}{bai2025qwen3vltechnicalreport} & 30B & .115 & .204 & .291 & 1.745 & .080 & .375 & .334 & .548 & .582 & .308 & .235 & .227 & .218 & .200 & .620 & .38 & .862 & .977 & .791 \\
\model{qwen}{Qwen2.5-VL-32B}{bai2025qwen25vltechnicalreport} & 32B & .180 & .113 & .260 & 2.671 & .065 & .352 & .285 & .103 & .525 & \first{.780} & .065 & .239 & .156 & .130 & .183 & .22 & \first{1.00} & \first{1.00} & .989 \\
\cmidrule(lr){1-21}
\model{qwen}{Qwen2-VL-72B}{wang2024qwen2} & 72B & .155 & .195 & .385 & 2.590 & .105 & .285 & .325 & .099 & .485 & .735 & .375 & .395 & .485 & .545 & .475 & .43 & .945 & .988 & .935 \\
\model{qwen}{Qwen2.5-VL-72B}{bai2025qwen25vltechnicalreport} & 72B & .185 & .234 & .430 & 2.448 & .135 & .304 & .338 & .108 & .514 & \first{.780} & \second{.500} & .433 & .661 & \first{.670} & .589 & .51 & .993 & \first{1.00} & \first{1.00} \\
\model{meta}{Llama-3.2-90B-V}{grattafiori2024llama} & 90B & .025 & .275 & .065 & 2.095 & \second{.185} & .325 & .345 & .175 & .545 & .715 & .024 & .625 & \third{.680} & .155 & .105 & .31 & .855 & \first{1.00} & \third{.990} \\
\model{meta}{Llama-4-Scout}{llama4-2025} & 109B & .105 & .033 & .100 & 2.350 & .045 & .348 & \third{.350} & .101 & .540 & \third{.755} & .021 & .389 & .193 & .140 & .200 & .25 & .973 & .972 & .967 \\
\model{mistral}{Mistral-Large}{mistrallarge-2024} & 123B & .060 & .226 & .198 & 1.895 & .050 & .325 & .300 & .095 & .667 & .241 & .031 & .479 & .330 & .625 & .284 & .37 & .977 & .808 & .807 \\
\model{qwen}{Qwen3-VL-235B-A22B}{bai2025qwen3vltechnicalreport} & 235B & .225 & .230 & \second{.482} & 1.225 & .090 & .405 & .301 & .091 & .609 & .471 & .125 & .231 & .151 & .495 & .168 & .28 & .986 & .983 & .698 \\
\model{qwen}{Qwen3.5-9B}{qwenteam2026qwen35omnitechnicalreport} & 9B & .175 & .195 & .315 & 2.180 & .075 & .310 & .328 & .145 & .585 & .685 & .425 & .315 & .285 & .278 & .355 & .30 & .965 & .988 & .945 \\
\model{qwen}{Qwen3.5-27B}{qwenteam2026qwen35omnitechnicalreport} & 27B & .205 & .218 & .355 & 1.580 & .115 & .395 & .346 & .478 & .605 & .525 & .315 & .350 & .305 & .305 & .402 & .38 & .978 & \third{.992} & .912 \\
\model{qwen}{Qwen3.5-122B-A10B}{qwenteam2026qwen35omnitechnicalreport} & 122B & .242 & .265 & \first{.495} & 1.085 & \third{.140} & .425 & .342 & .128 & .624 & .685 & \third{.478} & .385 & .265 & .425 & .355 & .42 & .988 & \first{1.00} & .858 \\
\model{meta}{Llama-4-Maverick}{llama4-2025} & 400B & .190 & .128 & .370 & 2.577 & .065 & .234 & \first{.358} & .104 & .499 & .715 & .045 & .172 & .138 & .105 & .061 & .16 & \first{1.00} & \first{1.00} & .956 \\
\model{qwen}{Qwen-VL-Max}{bai2025qwen25vltechnicalreport} & --- & .210 & .220 & .372 & 1.420 & .085 & .385 & .335 & .092 & .587 & .293 & .235 & .247 & .161 & .171 & .352 & .28 & .996 & \second{.994} & .960 \\
\cmidrule(lr){1-21}
\model{openai}{GPT-4o-mini}{hurst2024gpt} & --- & .130 & .228 & .348 & .895 & .060 & .395 & .337 & .139 & .529 & .241 & .021 & .548 & .578 & \second{.651} & \first{.800} & \first{.59} & \second{.999} & \first{1.00} & \second{.994} \\
\model{openai}{GPT-4o}{hurst2024gpt} & --- & .185 & .205 & .461 & 1.339 & .083 & .310 & .346 & .148 & .581 & .241 & .030 & .226 & .151 & .227 & .396 & .30 & \second{.999} & \first{1.00} & .924 \\
\model{openai}{GPT-4.1-mini}{openai2025gpt41} & --- & .305 & .397 & .370 & .676 & .055 & .376 & .342 & \second{.594} & .591 & .220 & .052 & .228 & .156 & .060 & .450 & .30 & \first{1.00} & \first{1.00} & .956 \\
\model{openai}{GPT-4.1}{openai2025gpt41} & --- & .360 & .440 & .420 & .520 & .075 & .410 & .343 & \first{.610} & .610 & .270 & .047 & .300 & .210 & .250 & .500 & .37 & \first{1.00} & \first{1.00} & .965 \\
\model{openai}{GPT-5-mini}{singh2025openai} & --- & .385 & .405 & .425 & .355 & .090 & .440 & .344 & .570 & .605 & .480 & .060 & .490 & .545 & .320 & .540 & .46 & \first{1.00} & \first{1.00} & .970 \\
\model{openai}{GPT-5}{singh2025openai} & --- & \second{.455} & \second{.480} & .470 & .225 & .110 & \third{.500} & .348 & .585 & .625 & .740 & .125 & .570 & .620 & .420 & .620 & .53 & \first{1.00} & \first{1.00} & .975 \\
\model{openai}{o4-mini}{openai2025o3} & --- & .335 & .390 & .415 & .385 & .085 & .420 & .328 & .395 & .580 & .555 & .085 & .510 & .485 & .280 & .555 & .45 & .995 & \first{1.00} & .940 \\
\model{openai}{o3}{openai2025o3} & --- & .420 & .450 & \third{.475} & \first{.170} & .115 & \second{.505} & .332 & .445 & .645 & .690 & .185 & .615 & .650 & .475 & \second{.695} & \third{.56} & \first{1.00} & \first{1.00} & .985 \\
\cmidrule(lr){1-21}
\model{claude}{Claude-3.5-Sonnet}{anthropic2024claude35} & --- & .065 & .199 & .210 & .832 & .045 & .377 & .325 & .199 & .485 & .220 & .031 & .383 & .239 & .295 & .206 & .28 & .982 & \first{1.000} & \first{1.000} \\
\model{claude}{Claude-Sonnet-4.5}{anthropic2025claude45} & --- & .265 & .295 & .380 & .485 & .065 & .445 & .345 & .315 & .620 & .545 & .125 & .530 & .395 & .490 & .455 & .45 & .993 & \first{1.00} & .975 \\
\model{claude}{Claude-Opus-4.5}{anthropic2025opus45} & --- & .395 & .415 & .455 & .275 & .095 & .485 & \third{.350} & .525 & .660 & .710 & .265 & .620 & .545 & .580 & .625 & .55 & \first{1.00} & \first{1.00} & .985 \\
\model{claude}{Claude-Sonnet-4.5-T}{anthropic2025claude45} & --- & .305 & .335 & .410 & .395 & .070 & .470 & .346 & .340 & .635 & .575 & .195 & .595 & .465 & .535 & .545 & .50 & .995 & \first{1.00} & .980 \\
\model{claude}{Claude-Opus-4.5-T}{anthropic2025opus45} & --- & \third{.445} & \third{.455} & \third{.475} & .205 & .105 & \third{.500} & \second{.352} & .565 & .665 & .745 & .345 & \second{.660} & .605 & .625 & \third{.690} & \first{.59} & \first{1.00} & \first{1.00} & \third{.990} \\
\cmidrule(lr){1-21}
\model{gemini}{Gemini-3-Flash}{gemini3-2025} & --- & \first{.668} & \first{.757} & .425 & \second{.175} & .110 & \second{.505} & \second{.352} & \third{.590} & \third{.680} & \third{.755} & .285 & \third{.645} & .585 & .605 & .680 & \second{.57} & \first{1.00} & \first{1.00} & .985 \\
\model{gemini}{Gemini-2.5-Flash-Lite}{comanici2025gemini} & --- & .264 & .176 & .434 & \third{.202} & .065 & .194 & .346 & .101 & .553 & .264 & .049 & .258 & .170 & .422 & .609 & .38 & .799 & .746 & .598 \\
\model{grok}{Grok-4}{grok4-2025} & --- & .215 & .220 & .405 & .203 & .025 & \first{.510} & .299 & .089 & \first{.710} & .295 & .060 & .233 & .193 & .020 & .128 & .22 & .978 & \first{1.00} & .800 \\
\model{grok}{Grok-4.1}{grok4-2025} & --- & .240 & .184 & .385 & .203 & .020 & .452 & .295 & .087 & \second{.694} & .265 & .020 & .222 & .165 & .080 & .139 & .22 & .961 & \first{1.00} & .933 \\
\model{zhipu}{GLM-4.6V}{hong2025glm} & --- & .145 & .210 & .325 & 1.925 & .085 & .295 & .315 & .205 & .555 & .415 & .125 & .335 & .285 & .385 & .355 & .33 & .925 & .965 & .895 \\
\bottomrule
\end{tabular}}%
\end{table}

%% file: Tables/tab_mas_methods.tex
\begin{table}[tbh]
\caption{Multi-agent inverse design results. Metrics: \textbf{PSel} (Param. Selection Acc.), \textbf{CL} (Closed-Loop Success), and \textbf{HS} (Human Score). \#Ag.\ is the number of agents per run, and $N,M$ are method counts (Tab.~\ref{tab:mas_baselines}). Sequential Pipeline reports Cascaded and Oracle (perfect upstream) modes. }
\label{tab:mas_methods}
\centering
\resizebox{\textwidth}{!}{%
\begin{tabular}{l l c ccc ccc ccc}
\toprule
& & &
  \multicolumn{3}{c}{\textbf{GPT-4o-mini}} &
  \multicolumn{3}{c}{\textbf{Qwen-VL-Max}} &
  \multicolumn{3}{c}{\textbf{Gemini-3.1-Lite}} \\
\cmidrule(lr){4-6}\cmidrule(lr){7-9}\cmidrule(lr){10-12}
\textbf{Method} & \textbf{Coord.} & \textbf{\#Ag.} &
  \textbf{PSel} & \textbf{CL} & \textbf{HS} &
  \textbf{PSel} & \textbf{CL} & \textbf{HS} &
  \textbf{PSel} & \textbf{CL} & \textbf{HS} \\
\midrule
Vanilla            & Single     & 1      & .110 & .540 & .34 & .360 & .580 & \third{.45} & .240 & .270 & .28 \\
CoT                & Single     & 1      & .110 & .490 & .32 & \third{.440} & .520 & \third{.45} & .240 & .270 & .28 \\
Self-Consistency   & Voting     & $N$    & .110 & .590 & .36 & .410 & .520 & .44 & .230 & .270 & .28 \\
\cmidrule(lr){1-12}
LLM Debate         & Debate     & 3      & .120 & .500 & .33 & .330 & .530 & .42 & .170 & .290 & .26 \\
MAD                & Debate     & 3+     & \third{.290} & \second{.731} & \second{.48} & .126 & .411 & .30 & \third{.280} & \second{.462} & \second{.37} \\
CAMEL              & Role-play  & 2      & .150 & .280 & .25 & .285 & .315 & .31 & .195 & .225 & .25 \\
\cmidrule(lr){1-12}
DyLAN              & DynGraph   & 4      & .135 & .585 & .37 & .345 & .505 & .41 & .215 & .305 & .28 \\
MacNet             & Graph      & $N$    & .125 & .525 & .34 & .310 & .475 & .39 & .205 & .285 & .27 \\
AgentVerse         & Expert     & 4+     & .120 & \third{.720} & .42 & .244 & .549 & .39 & .235 & .337 & .30 \\
\cmidrule(lr){1-12}
MAV                & Verify     & $N$+$M$ & .110 & .600 & .37 & .320 & .450 & .38 & .250 & .280 & .29 \\
AutoGen            & Code-exec  & 2      & .160 & .680 & .42 & .190 & .350 & .29 & .210 & .290 & .28 \\
EvoMAC             & Evolution  & $N$    & .040 & .232 & .19 & .155 & \second{.732} & .43 & .141 & \third{.444} & .31 \\
\cmidrule(lr){1-12}
Seq.\ Pipeline (Cascaded) & Dom.-decomp.\ & 4 & \second{.325} & .655 & \third{.46} & \second{.475} & \third{.620} & \second{.50} & \second{.305} & .425 & \third{.36} \\
Seq.\ Pipeline (Oracle)   & Dom.-decomp.\ & 4 & \first{.485} & \first{.785} & \first{.57} & \first{.580} & \first{.755} & \first{.59} & \first{.420} & \first{.595} & \first{.47} \\
\bottomrule
\end{tabular}}%

\end{table}

%% file: Tables/tab_fewshot_main.tex
\begin{wraptable}{r}{0.535\textwidth}
\vspace{-14pt}
\centering
\caption{\small Family accuracy and closed-loop success with $k$ in-context exemplars (mean over five support-set draws).}
\label{tab:fewshot_main}
\footnotesize
\setlength{\tabcolsep}{1.6pt}
\renewcommand{\arraystretch}{1.15}
\begin{tabular}{l cccc cccc}
\toprule
& \multicolumn{4}{c}{\textbf{Family Acc}} & \multicolumn{4}{c}{\textbf{CL Succ.}} \\
\cmidrule(lr){2-5}\cmidrule(lr){6-9}
\textbf{Model} & $k{=}0$ & $k{=}2$ & $k{=}4$ & $k{=}8$ & $k{=}0$ & $k{=}2$ & $k{=}4$ & $k{=}8$ \\
\midrule
Gemini-3-Flash  & .668 & .706 & .741 & .763 & .680 & .682 & .676 & .681 \\
GPT-5           & .455 & .521 & .566 & .592 & .620 & .626 & .617 & .624 \\
GPT-4o-mini     & .130 & .217 & .259 & .291 & .800 & .794 & .791 & .796 \\
Qwen2.5-VL-72B  & .185 & .224 & .251 & .268 & .589 & .593 & .584 & .590 \\
Qwen2.5-VL-7B   & .096 & .134 & .159 & .172 & .638 & .643 & .634 & .641 \\
\bottomrule
\end{tabular}
\vspace{-10pt}
\end{wraptable}

%% file: Tables/tab_sem_main.tex
\centering
\makeatletter\def\@captype{table}\makeatother
\caption{Family accuracy on corresponding simulated and real SEM images (10 families).}
\label{tab:sem_main}
\footnotesize
\setlength{\tabcolsep}{3pt}
\renewcommand{\arraystretch}{1.12}
\begin{tabular*}{\linewidth}[b]{@{\extracolsep{\fill}} l cc}
\toprule
\textbf{Model} & \textbf{Simulated} & \textbf{Real SEM} \\
\midrule
Gemini-3-Flash  & .642 & .525 \\
GPT-5           & .467 & .383 \\
Claude-Opus-4.5 & .408 & .317 \\
Qwen2.5-VL-72B  & .175 & .133 \\
GPT-4o-mini     & .142 & .092 \\
Qwen2.5-VL-7B   & .083 & .058 \\
\bottomrule
\end{tabular*}

%% file: Tables/tab_propagation_feedback.tex
\begin{table}[!htbp]
\centering
\begin{minipage}[t]{0.40\textwidth}
\centering
\caption{Accuracy over 600 fibers on the balanced dispersion (P1) and pulse broadening (P2) tasks.}
\label{tab:propagation_main}
\footnotesize
\setlength{\tabcolsep}{4pt}
\begin{tabular}{l cc}
\toprule
& \textbf{Dispersion} & \textbf{Broadening} \\
\cmidrule(lr){2-2}\cmidrule(lr){3-3}
\textbf{Model} & P1 & P2 \\
\midrule
Gemini-3-Flash  & \textbf{.683} & .562 \\
GPT-5           & .657 & \textbf{.583} \\
Claude-Opus-4.5 & .635 & .572 \\
GPT-4o-mini     & .568 & .528 \\
Qwen2.5-VL-72B  & .552 & .515 \\
Qwen2.5-VL-7B   & .520 & .492 \\
Chance          & .500 & .500 \\
\bottomrule
\end{tabular}
\end{minipage}\hfill
\begin{minipage}[t]{0.57\textwidth}
\centering
\definecolor{pipbase}{HTML}{2E6DA4}\definecolor{pipgain}{HTML}{9EC5E8}\definecolor{piptrack}{HTML}{E4E4E4}%
\newcount\pipn
\newcommand{\pips}[2]{\pipn=#2\relax\loop\ifnum\pipn>0 \textcolor{#1}{\rule{1.4pt}{1.3ex}}\hspace{0.4pt}\advance\pipn-1\repeat}%
\newcommand{\pipkey}[1]{\textcolor{#1}{\rule{3pt}{1.3ex}}}%
\newcommand{\gainbar}[2]{\makebox[0.9em][r]{#1}\hspace{3pt}\pips{pipbase}{#1}\pips{pipgain}{\numexpr#2-#1\relax}\pips{piptrack}{\numexpr20-#2\relax}\hspace{2.7pt}\makebox[0.9em][l]{#2}}%
\caption{Cases improved out of 20 with single-axis or joint edits, after one proposal (\pipkey{pipbase}) and after three feedback rounds (gain in \pipkey{pipgain}).}
\label{tab:feedback_main}
\footnotesize
\setlength{\tabcolsep}{4pt}
\begin{tabular}{l cc}
\toprule
& \textbf{Single-axis edits} & \textbf{Joint edits} \\
\cmidrule(lr){2-2}\cmidrule(lr){3-3}
\textbf{System} & 1 $\to$ 3 rounds & 1 $\to$ 3 rounds \\
\midrule
GPT-4o-mini              & \gainbar{6}{10} & \gainbar{6}{11} \\
GPT-5                    & \gainbar{7}{11} & \gainbar{8}{11} \\
Gemini-3-Flash           & \gainbar{8}{11} & \gainbar{7}{11} \\
Claude-Opus-4.5          & \gainbar{4}{9}  & \gainbar{5}{9}  \\
Qwen2.5-VL-72B           & \gainbar{5}{9}  & \gainbar{6}{10} \\
Qwen2.5-VL-7B            & \gainbar{6}{8}  & \gainbar{5}{8}  \\
Seq.\ Pipeline (Cascaded) & \gainbar{3}{8}  & \gainbar{4}{9}  \\
\bottomrule
\end{tabular}
\end{minipage}
\vspace{-6pt}
\end{table}

%% file: sections/case_study.tex
\subsection{What Do Models Actually Design?}
\label{sec:case_study}

\begin{figure}[!htbp]
  \centering
  \includegraphics[width=0.95\textwidth]{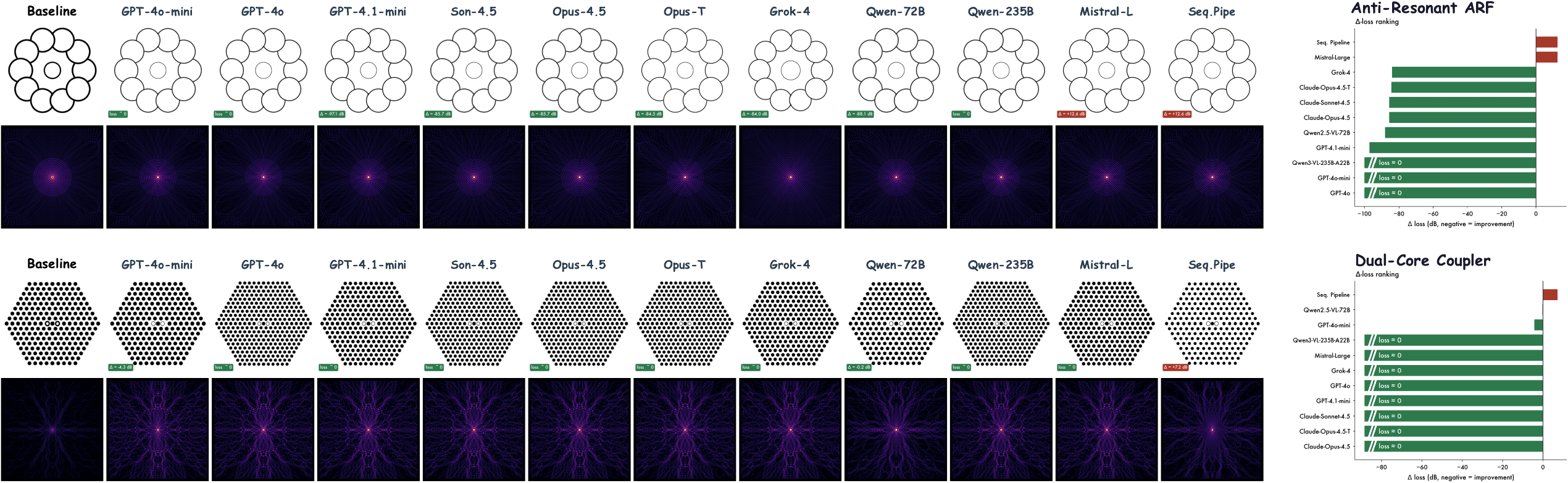}
  \vspace{-4pt}
  \caption{\small Closed-loop inverse design on two PCFs (left: ARF, right: dual-core coupler). Top: structures with $\Delta$-loss, bottom: mode fields and ranking (negative = improvement). More cases in Appendix~\ref{app:case_catalog}.}
  \vspace{-10pt}
  \label{fig:case_study}
\end{figure}

To move beyond aggregate metrics, we close the simulation loop on two representative fibers (Fig.~\ref{fig:case_study}): an ARF (moderate loss) and a dual-core coupler (near-floor loss).  For each case, 11 systems propose a parameter change, which is applied to the original geometry and re-simulated with the same Meep FDTD solver.  The resulting loss change, $\Delta = 10\log_{10}(L_\text{new}/L_\text{base})$\,dB, quantifies improvement (negative) or degradation (positive). On the ARF, 9/11 systems improve, six by 84 to 97\,dB and three to a numerically zero loss, mostly by retuning the tube wall thickness.  The dual-core coupler shows a complementary pattern: 10/11 systems improve despite a near-floor baseline, most of them by adding cladding rings.  In both cases, success occurs when the task reduces to a \emph{single-axis, monotonic} parameter change aligned with common training patterns. Results on 20 cases (Appendix~\ref{app:case_catalog}) show that this tractability is family-dependent, as nested and quasi-crystalline geometries that need coordinated multi-parameter changes stay unsolved. \textbf{Tab.~\ref{tab:feedback_main}} shows that joint two-parameter edits change outcomes by at most one case, whereas three feedback rounds add two to five improved cases, so models gain more from feedback than from a larger action space.

\paragraph{Looking Forward.}
Closing the \emph{perception--design gap} hinges on three directions. First, supervision should match the task: simulation-grounded structure--field--loss data provide signals absent from internet-scale pretraining, and domain-specific adaptation can outweigh scale. Second, closed-loop success depends on magnitude, constraints, and multi-parameter coordination, not only on direction, so training signals should reward feasibility, consistency, and bounded updates. Third, solver feedback should be integrated into the loop, via tool use at inference or as a reward for closed-loop tuning, offering greater gains than further scaling of models.

%% file: sections/conclusion.tex
\section{Conclusion}
\label{sec:conclusion}

This paper proposes PCFBench, the first comprehensive benchmark for PCF inverse design with vision-language models, built on 220K FDTD-simulated samples across 18 fiber families. It defines 26 tasks from perception to design and evaluates 41 VLMs alongside 13 MAS. Our results reveal a persistent \emph{perception--design gap}: accurate geometry recognition fails to translate into physically grounded design decisions, even under scaling or multi-agent coordination. Closed-loop re-simulation shows that success largely reduces to single-axis edits, while multi-parameter geometries remain challenging. PCFBench serves as a testbed for simulation-in-the-loop training, VLM-based agentic design, and solver-coupled decision making.

%% file: sections/acknowledgements.tex
\begin{ack}
We thank the School of Electronic Science and Technology and the School of Information and Communication Engineering at Hainan University for their valuable expert guidance and domain knowledge throughout the construction and validation of PCFBench. We are also grateful to the School of Artificial Intelligence at Shenzhen University for providing the computational infrastructure and API access that enabled the large-scale evaluation of VLMs reported in this work.
\end{ack}

%% file: sections/appendix.tex
\definecolor{promptbg}{HTML}{F6F8FB}
\definecolor{promptborder}{HTML}{D5DCE5}
\definecolor{promptrole}{HTML}{475569}
\definecolor{promptimg}{HTML}{2563EB}
\definecolor{promptfield}{HTML}{7C3AED}

\definecolor{promptboxfill}{HTML}{F0F7F6}
\definecolor{promptboxframe}{HTML}{2E6E66}
\definecolor{extractgreen}{HTML}{1F7A4C}

\newenvironment{taskprompt}[1][Prompt Template]
{%
  \par\nopagebreak
  \begin{tcolorbox}[%
    enhanced, breakable,
    colback=promptboxfill,
    colframe=promptboxframe,
    title=\textbf{#1},
    coltitle=white,
    fonttitle=\bfseries\small,
    fontupper=\small,
    boxrule=0.6pt,
    arc=2pt,
    left=8pt, right=8pt, top=5pt, bottom=6pt,
    width=\textwidth,
    before skip=6pt, after skip=12pt,
    before upper={\setlength{\parskip}{2.5pt}\setlength{\parindent}{0pt}\raggedright}
  ]%
}
{%
  \end{tcolorbox}%
}

\newcommand{\imgtoken}[1]{\textcolor{promptimg}{\texttt{\footnotesize[img:#1]}}}
\newcommand{\promptvar}[1]{\textcolor{promptfield}{\texttt{\{#1\}}}}
\newcommand{\sysrole}{\textbf{System Role:}}
\newcommand{\prompthead}[1]{\textbf{#1:}}
\lstdefinestyle{pcfcode}{basicstyle=\ttfamily\footnotesize, upquote=true, columns=fullflexible, keepspaces=true, breaklines=true, breakatwhitespace=true, showstringspaces=false, frame=none, xleftmargin=0pt, aboveskip=2pt, belowskip=2pt, keywordstyle={}, commentstyle=\color{promptrole}\itshape}
\newcommand{\promptextraction}[1]{\par\vspace{1pt}\textcolor{extractgreen}{\textbf{Extraction:} #1}}
\newcommand{\promptmetric}[1]{\par\textcolor{extractgreen}{\textbf{Metric:} #1}}
\newcommand{\promptimages}[1]{\par\textcolor{extractgreen}{\textbf{Images:} #1}}

\clearpage
\section{Dataset}
\label{app:dataset_top}

PCFBench is organized to expose both the geometric diversity of photonic crystal fibers and the controlled simulation assumptions underlying the benchmark. This section covers dataset composition (Appendix~\ref{app:dataset}), the 18 families (Appendix~\ref{app:cases}), the annotation levels (Appendix~\ref{app:annotation_showcase}), quality control (Appendix~\ref{app:qc_failure_modes}), the comparison with prior benchmarks (Appendix~\ref{app:bench_comparison}), and the simulator configuration that makes per-family outputs comparable (Appendix~\ref{app:fdtd_setup}).

\subsection{Statistics and Composition}
\label{app:dataset}

\begin{figure}[H]
\centering
\begin{minipage}[t]{0.5\textwidth}
\vspace{0pt}
The term \emph{family} in PCFBench refers to a \textbf{canonical physics category} distinguished by its dominant guidance mechanism and topological motif (e.g., anti-resonant guidance with a single tube ring, or solid-core total-internal-reflection over a hexagonal lattice). Each family may admit multiple structural variants (single-ring vs.\ nested-ring anti-resonance, hexagonal vs.\ octagonal lattice spacing, $1$-, $3$-, or $7$-tube cobweb struts, etc.), but for the purposes of benchmarking these are grouped under the parent family to keep per-family statistics meaningful. PCFBench comprises \textbf{220{,}000 samples across 18 families}, partitioned into four groups by guidance mechanism (as shown in \textbf{Fig.~\ref{fig:dataset_groups}}). \textbf{Fig.~\ref{fig:distributions}} shows the corpus distributions of pitch, hole diameter, ring count, and confinement loss.
\end{minipage}\hfill
\begin{minipage}[t]{0.47\textwidth}
\vspace{0pt}
\centering
\includegraphics[width=\linewidth]{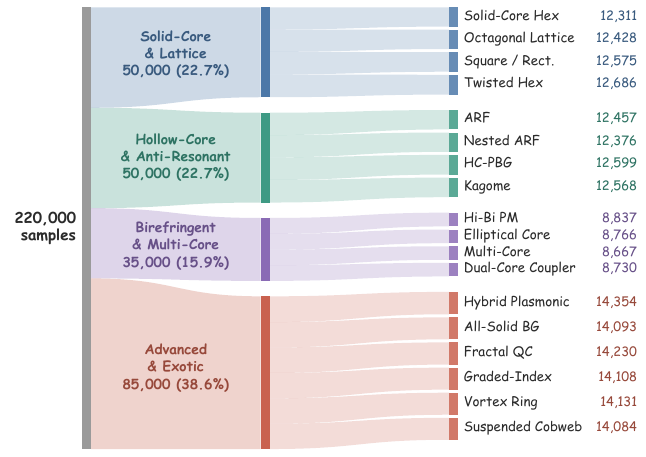}
\caption{The 220{,}000-sample corpus, split into four groups by guidance mechanism and then into 18 families. Node heights and ribbon widths are proportional to sample counts.}
\label{fig:dataset_groups}
\end{minipage}
\end{figure}

\begin{figure}[tbh]
\centering
\includegraphics[width=\textwidth]{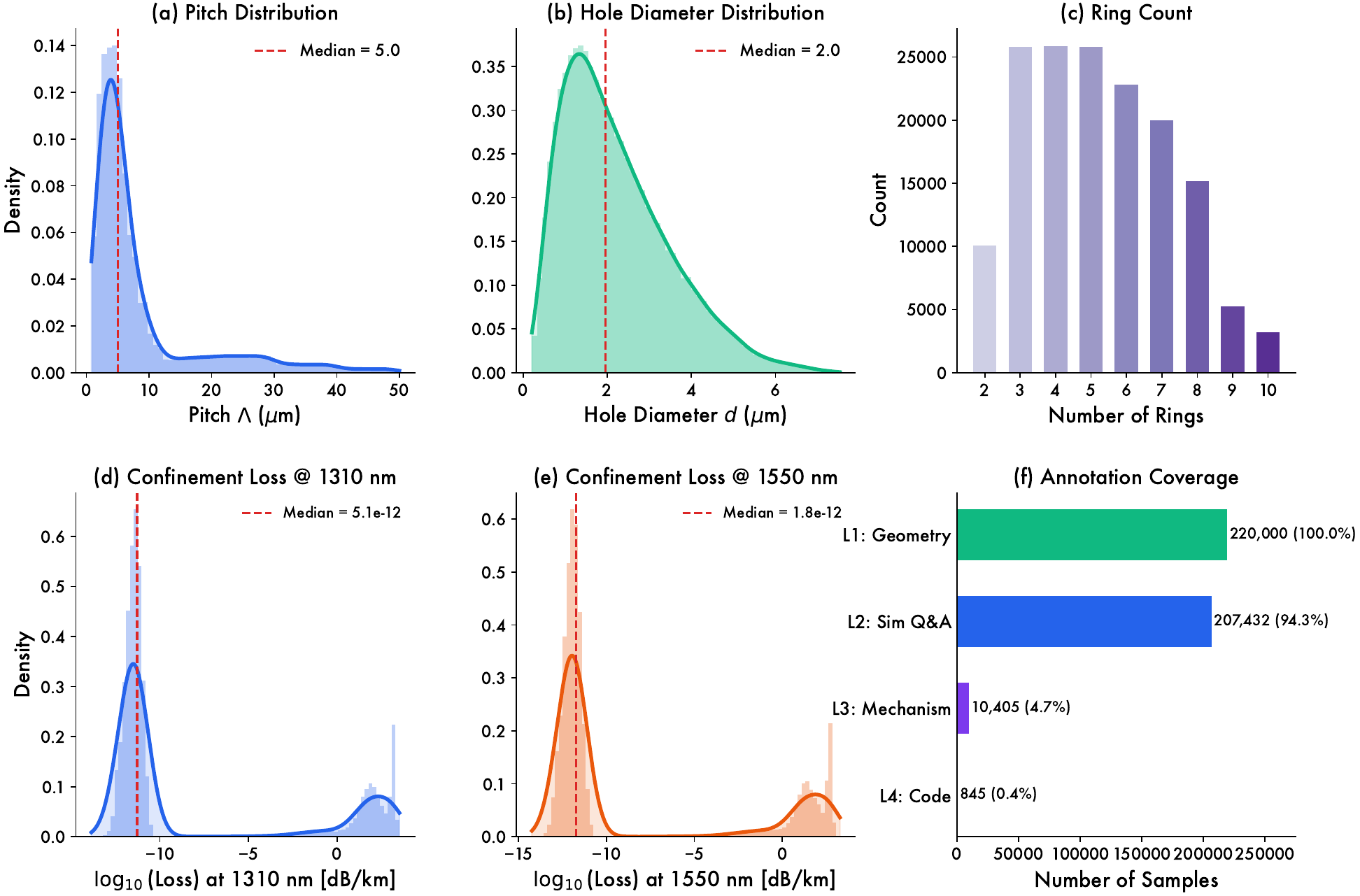}
\caption{Dataset distributions. \textbf{(a--b)} Pitch and hole diameter with KDE (dashed: median). \textbf{(c)} Ring count. \textbf{(d--e)} Log-scale loss at 1310\,nm and 1550\,nm. \textbf{(f)} Annotation-level coverage.}
\label{fig:distributions}
\end{figure}

\subsection{Family Catalog}
\label{app:cases}

\begin{figure}[!htbp]
\centering
\includegraphics[width=\textwidth]{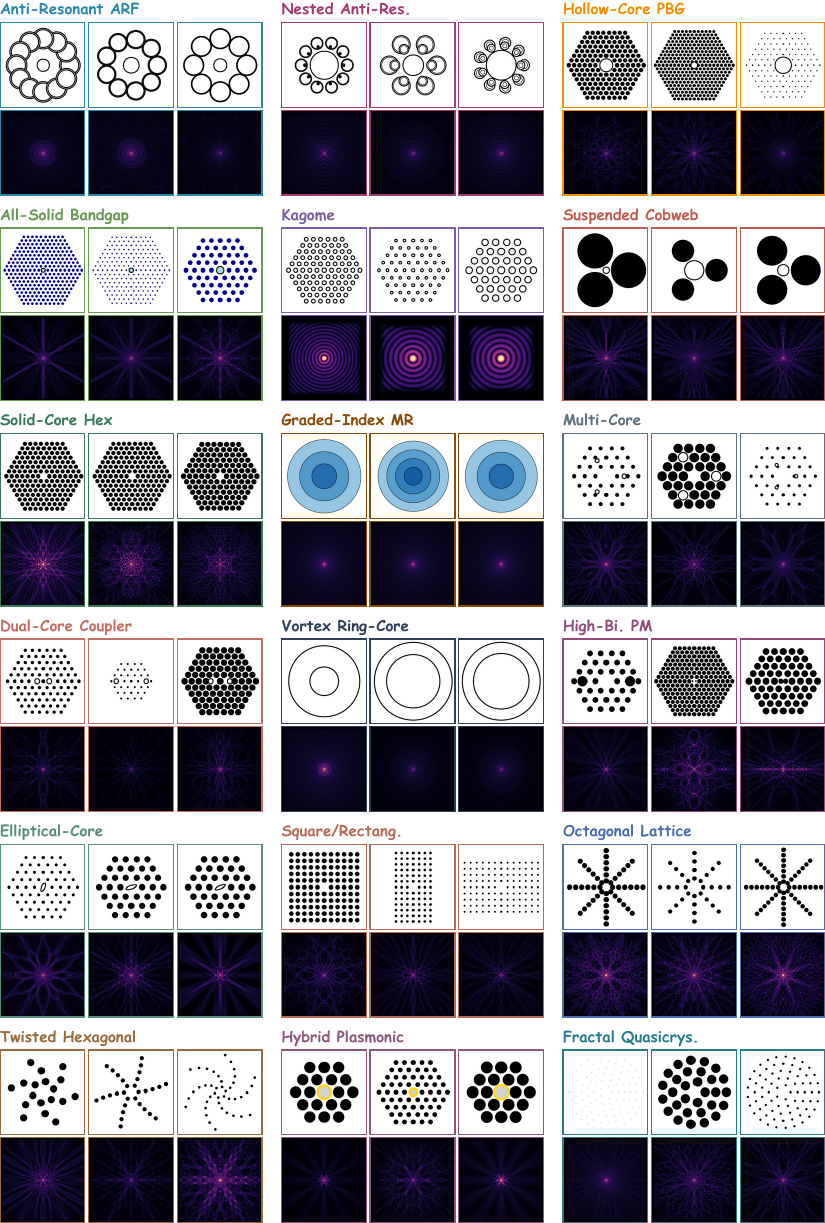}
\caption{Three exemplars per family from the 220{,}000-sample corpus. Each structure (upper row) sits above its FDTD-simulated fundamental mode field at $\lambda=1550$\,nm (lower row).}
\label{fig:family_gallery}
\end{figure}

\textbf{Fig.~\ref{fig:family_gallery}} shows three exemplars of each family, with every structure paired with its simulated mode field. The list below gives a one-sentence structural description and the principal application domain of each family, in the same order.

\begin{itemize}
  \item \textbf{Anti-Resonant ARF}: a hollow core surrounded by a single ring of non-touching cylindrical tubes whose wall thickness lies at an anti-resonance of the guided wavelength. It is used for high-power laser delivery and gas-cell sensing where low nonlinearity matters.
  \item \textbf{Nested Anti-Resonant}: an ARF augmented with a second (inner) concentric tube inside each outer tube, which suppresses coupling between core and cladding modes. It is the current state of the art for ultra-low-loss short-pulse delivery in the UV--mid-IR.
  \item \textbf{Hollow-Core PBG}: a hollow defect in a triangular air-hole lattice that forms a photonic bandgap. Its high-Q narrow-band guidance suits low-latency telecom and spectroscopy.
  \item \textbf{All-Solid Bandgap}: a high-index-rod lattice with a lower-index solid core, producing a bandgap without air holes. It is used in high-power amplifiers and radiation-hard systems.
  \item \textbf{Kagome}: a star-of-David lattice of large air voids around a hollow core. It is the canonical inhibited-coupling geometry, used in UV-visible gas lasers and short-pulse delivery.
  \item \textbf{Suspended Cobweb}: a small glass core suspended by thin struts inside a large air void. Its high evanescent-field fraction serves gas/liquid sensing and supercontinuum.
  \item \textbf{Solid-Core Hex}: the classical microstructured silica fiber with a solid defect core in a triangular lattice of air holes. It is the workhorse geometry for endlessly single-mode transmission and dispersion-engineered supercontinuum.
  \item \textbf{Graded-Index Multi-Ring}: concentric air-hole rings with radially varying diameter, giving a smooth effective-index profile for bandwidth-flat wavelength-division-multiplexing links.
  \item \textbf{Multi-Core}: several independent solid cores on a lattice in one common cladding. It is the standard platform for space-division multiplexing in next-generation telecom.
  \item \textbf{Dual-Core Coupler}: two cores set at a chosen spacing so modes couple periodically along $z$. It serves as a compact all-fiber coupler, wavelength-sensitive filter, and sensor.
  \item \textbf{Vortex Ring-Core}: an annular high-index core supporting orbital-angular-momentum (OAM) states for mode-division multiplexing and structured-light telecom.
  \item \textbf{High-Birefringence PM}: an asymmetric cladding (e.g., two enlarged stress elements) breaking the fiber's rotational symmetry so the two polarization eigenmodes differ strongly. It is the workhorse for polarization-maintaining interferometric sensors.
  \item \textbf{Elliptical-Core}: a solid core with an elliptical cross-section producing form-induced birefringence, an alternative polarization-maintaining platform for integrated optics.
  \item \textbf{Square/Rectangular Lattice}: a holey cladding arranged on a square (or rectangular) lattice and engineered for anisotropic dispersion and supercontinuum applications.
  \item \textbf{Octagonal Lattice}: an eight-fold symmetric air-hole array whose broader first Brillouin zone gives a wider bandgap than a triangular lattice, suiting low-loss hollow-core telecom.
  \item \textbf{Twisted Hexagonal}: a hexagonal lattice helically twisted along the fiber axis, inducing circular birefringence. It serves polarization-insensitive transport and OAM conversion.
  \item \textbf{Hybrid Plasmonic}: a PCF incorporating metallic (typically gold) inclusions that excite surface plasmons resonant with the fundamental core mode. It is the standard geometry for ultra-sensitive surface-plasmon-resonance biosensing.
  \item \textbf{Fractal Quasicrystal}: an aperiodic cladding built from a fractal (Penrose- or Vogel-like) lattice of holes, studied for engineered dispersion, slow light, and on-chip photonic computing.
\end{itemize}

\subsection{Annotation Showcase}
\label{app:annotation_showcase}

The boxes below trace one anti-resonant sample through all four annotation levels. The structural configuration is fixed throughout: a single ring of seven silica tubes around a hollow core, with pitch $\Lambda=34.34\,\mu$m, core radius $20.60\,\mu$m, tube radius $2.49\,\mu$m, and wall thickness $1.47\,\mu$m, simulated at 1310\,nm and 1550\,nm. Each level inherits the facts of the level below and adds one semantic tier (caption $\to$ quantitative $\to$ diagnostic $\to$ executable).

\definecolor{lvl1}{HTML}{1E3A5F}
\definecolor{lvl2}{HTML}{2E6E66}
\definecolor{lvl3}{HTML}{B4671B}
\definecolor{lvl4}{HTML}{475569}

\newtcolorbox{annotbox}[2]{%
    enhanced, breakable,
    colback=promptbg,
    colframe=promptborder,
    coltitle=white,
    fonttitle=\bfseries\footnotesize,
    fontupper=\small,
    boxrule=0.4pt,
    arc=2pt,
    left=10pt, right=10pt, top=5pt, bottom=5pt,
    width=\textwidth,
    before skip=5pt, after skip=5pt,
    title={#2},
    colbacktitle=#1,
    attach boxed title to top left={xshift=8pt, yshift=-7pt},
    boxed title style={%
        colback=#1, colframe=#1, arc=2pt,
        boxrule=0pt, left=6pt, right=6pt, top=2pt, bottom=2pt%
    },
    before upper={\setlength{\parskip}{3pt}\setlength{\parindent}{0pt}}}

\begin{annotbox}{lvl1}{L1 \textbullet\ Geometry Caption}
{\color{promptrole}\small\textsc{prompt}}\quad Describe the PCF cross-section geometry, identify its family, and explain the guidance mechanism.

{\color{promptrole}\small\textsc{annotation}}\quad Hollow-core anti-resonant fiber featuring a single ring of seven identical circular silica tubes around a central air core. The dominant guidance mechanism is anti-resonant reflection: confinement arises from destructive interference of light coupling into the tube walls. Pitch $34.34\,\mu$m, tube radius $2.49\,\mu$m, wall thickness $1.47\,\mu$m, core radius $20.60\,\mu$m.
\end{annotbox}

\begin{annotbox}{lvl2}{L2 \textbullet\ Quantitative Q\&A (5 Q\&A pairs)}
\small
\renewcommand{\arraystretch}{1.15}
\begin{tabular}{@{}p{0.42\textwidth}p{0.52\textwidth}@{}}
\textbf{Q1.} Confinement loss at 1550\,nm?
& \textbf{A1.} $147.48$ (2D-Meep flux-integral proxy).\\
\textbf{Q2.} Effective index $n_{\text{eff}}$ at 1550\,nm?
& \textbf{A2.} $1.01$ (hollow-core ARF benchmark proxy).\\
\textbf{Q3.} Single-mode or multi-mode at 1550\,nm?
& \textbf{A3.} Multi-mode, because the large core radius supports several transverse modes within the anti-resonance window.\\
\textbf{Q4.} Confined or leaky in this configuration?
& \textbf{A4.} Leaky, because the elevated loss proxy ($\sim 10^{2}$) places it outside the anti-resonance plateau.\\
\textbf{Q5.} Most likely application class?
& \textbf{A5.} High-power laser delivery and gas-cell sensing.\\
\end{tabular}
\end{annotbox}

\begin{annotbox}{lvl3}{L3 \textbullet\ Structured Diagnosis (Observation $\to$ Thought $\to$ Reasoning $\to$ Action $\to$ Verification)}
\small
\textbf{\color{lvl3}Observation.}\enspace The field image shows bright intensity rings along the tube walls, indicating significant interaction between the guided mode and the tube walls.\par
\textbf{\color{lvl3}Thought.}\enspace A single-ring 7-tube ARF whose bright wall rings suggest resonant coupling to surface modes on the silica tubes, consistent with anti-resonant guidance: confinement relies on destructive interference of light reflected from the inner tube walls.\par
\textbf{\color{lvl3}Reasoning.}\enspace Wall thickness $t = 1.47\,\mu$m places the first-order resonance at $\lambda_1 = 2t\sqrt{n_g^{2}-1} \approx 3.06\,\mu$m and the second-order resonance at $\lambda_1/2 \approx 1.53\,\mu$m, so the 1550\,nm band sits on a resonance, which explains the elevated loss proxy.\par
\textbf{\color{lvl3}Action.}\enspace Decrease wall thickness toward $\sim 0.42\,\mu$m, which moves $\lambda_1$ below $0.9\,\mu$m so that both telecom bands fall in the fundamental anti-resonance window, or adopt a nested-tube cladding to suppress core--cladding coupling without retuning $t$.\par
\textbf{\color{lvl3}Verification.}\enspace Both proposals are consistent with the closed-form ARROW resonance condition and with published anti-resonance maps for single-ring ARF designs.
\end{annotbox}

\begin{annotbox}{lvl4}{L4 \textbullet\ Executable Meep Script (first lines shown)}
\begin{lstlisting}[style=pcfcode, language=Python]
import meep as mp
import numpy as np

ORIGINAL_PARAMS = {
    'family': 'anti_resonant_arf',
    'core_radius_um': 20.60,
    'pitch_um': 34.34,
    'tube_count': 7,
    'tube_radius_um': 2.49,
    'wall_thickness_um': 1.47,
}
# geometry, source, monitors, run() ...
\end{lstlisting}
\end{annotbox}

\noindent The four levels are stored per sample and used independently by tasks: Tasks~1, 11, and 12 use L1, Tasks~6--10 use L2, Tasks~10 and 16--17 use L3, and Tasks~24--26 use L4. Cross-level consistency is enforced during quality control (Appendix~\ref{app:qc_failure_modes}) so that no L3 diagnosis contradicts the L2 numbers or the L4 simulation outcome.

\subsection{Quality Control}
\label{app:qc_failure_modes}

Every candidate sample passes five automated checks (\textbf{Tab.~\ref{tab:failure_modes}}) before expert review, and the released corpus contains only samples that clear all five. A rejected sample is regenerated with resampled parameters, a longer simulation time, a finer grid, or a new L3 annotation, so the released annotations carry no post-hoc corrections.

\input{Tables/tab_failure_modes}

\subsection{Comparison with Prior Benchmarks}
\label{app:bench_comparison}

\textbf{Tab.~\ref{tab:bench_comparison}} positions PCFBench against representative multimodal, scientific, and agent-oriented benchmarks released in 2023--2026. Three distinctions stand out. PCFBench is the only entry whose ground truth comes fully from an in-the-loop physical simulator (FDTD) rather than from human-curated text or image labels alone. It is also, with PICBench, one of two entries that require \emph{actionable inverse design} (parameter selection, direction, and magnitude). It also provides native multi-agent evaluation through a four-stage Sequential Pipeline plus 12 MASLab baselines, unlike most prior benchmarks.

\definecolor{cmpyes}{HTML}{15803D}
\definecolor{cmpno}{HTML}{7B8494}
\newcommand{\cmpY}{\textcolor{cmpyes}{\ding{51}}}
\newcommand{\cmpN}{\textcolor{cmpno}{\textendash}}

\input{Tables/tab_bench_comparison}

\subsection{Unified Simulation Configuration}
\label{app:fdtd_setup}

PCFBench uses a single 2D FDTD configuration across all 18 fiber families, summarized in \textbf{Tab.~\ref{tab:fdtd_config}}. The configuration is chosen for cross-family consistency at corpus scale: a continuous-wave point source on a uniform Yee grid, a fixed run time, and a flux-based loss proxy read from four near-PML monitors. Holding the grid, source, and monitors fixed across families keeps per-family loss and mode-field outputs comparable, so the closed-loop $\Delta$-loss in Sec.~\ref{sec:case_study} reflects the proposed design change. The plasmonic family models the metal as a lossy effective medium.

\input{Tables/tab_fdtd_config}

\paragraph{Numerical Sensitivity of the Simulator.}
\textbf{Tab.~\ref{tab:add_sensitivity}} reports a per-cluster resolution sweep ($40\!\to\!60$ pixels per $\mu$m) and a Sellmeier-vs.\ fixed-silica comparison on a stratified $200$-sample subset. Switching from fixed-silica $n_{\mathrm{SiO}_2}\!=\!1.444$ to a Malitson Sellmeier index leaves the model leaderboard ranking essentially unchanged (Spearman $\rho = 0.97$), so the rankings reported in Appendix~\ref{app:detailed_results} are robust to the silica dispersion model.

\input{Tables/tab_add_sensitivity}

\section{Evaluation Suite}
\label{app:eval_suite}

This section specifies the evaluation setup for the perception, reasoning, and inverse-design tasks. We define the shared protocol for every model call (Appendix~\ref{app:tasks_shared}), the per-category task prompts that fix the input--output format (Appendix~\ref{app:tasks}), the evaluated models (Appendix~\ref{app:models}), and the expert rubric that turns inverse-design proposals into the Human Score (Appendix~\ref{app:human_eval}).

\input{sections/app_task_prompts}

\subsection{Evaluated Models}
\label{app:models}

PCFBench evaluates a heterogeneous cohort of VLMs spanning ten vendors and roughly two orders of magnitude in parameter count. \textbf{Tab.~\ref{tab:model_catalogue}} catalogs every model with its parameter count, access type, and characterization. Thinking-variant rows (suffixed ``-T'') share a backbone with their non-thinking counterpart but are evaluated with extended test-time deliberation enabled.

\input{Tables/tab_model_catalogue}

\subsection{Human Evaluation Rubric}
\label{app:human_eval}

Human scoring targets the inverse-design tasks, where expert judgments of physical validity, coherence, and fabrication feasibility add to the sign and scalar scored by direction accuracy and the closed-loop $\Delta$-proxy. Following recent photonic inverse-design practice~\cite{li2023deep,nanophotonic_survey2025}, we complement these automated scores with a \textbf{Human Score} in $[0,1]$ rating each proposal along six dimensions: physics validity, quantitative correctness, fabrication feasibility, performance improvement, design coherence, and justification quality. \textbf{Tab.~\ref{tab:human_eval_rubric}} specifies concrete anchors for each dimension. A stratified 120-sample evaluation balances fiber-family group and loss regime, while dual expert ratings and Cohen's $\kappa$ quantify the reliability of the resulting Human Score, reported alongside the composite mean in the main leaderboard. Experts scored independently before adjudication. Disagreements larger than one rubric point were resolved by joint review, and smaller ones were averaged.

\input{Tables/tab_human_eval_rubric}

\paragraph{Inter-Rater Reliability of the Human Score.}
On the stratified $120$-sample human-evaluation subset, the two photonics experts achieve substantial agreement (\textbf{Tab.~\ref{tab:add_kappa}}). Adjudication was triggered for $12.5\%$ of items (those differing by more than one rubric point). Kendall $\tau$ between the leaderboards before and after adjudication is $0.97$, so the conclusions are insensitive to the adjudication step.

\input{Tables/tab_add_kappa}

\section{Detailed Per-Category Results}
\label{app:detailed_results}

\textbf{Tab.~\ref{tab:overview}} in the main text reports one representative metric per task. Tabs.~\ref{tab:geo_phys}--\ref{tab:inv_code} give every metric, with the six categories grouped in pairs, and supply its values.

\subsection{Geometry Perception and Physics Understanding}
\label{app:res_geo_phys}
\textbf{Tab.~\ref{tab:geo_phys}} shows the per-task results for geometry perception (left block) and physics understanding (right block). Models diverge far more on physics than on perception. The perception tasks ask for a family label, a count, a topology, or two metric parameters from the structure image, and the physics tasks ask for loss, application, mode, and mechanism. Scale and reasoning depth help most where the answer is discrete, yet no model recovers pitch or hole diameter accurately, so VLMs still read cross-sections as categorical patterns rather than as metric geometry. Loss prediction opens the widest proprietary/open-weight gap of the benchmark, which ties physics understanding to reasoning depth rather than to visual capacity. Near-perfect Mode Classification scores come from majority-class collapse, a shortcut that the Loss columns expose.
\input{Tables/tab_geo_phys}

\subsection{Text Generation and Multi-modal Reasoning}
\label{app:res_text_mm}
\textbf{Tab.~\ref{tab:text_mm}} shows the per-task results for text generation (left block) and multi-modal reasoning (right block). Fluent description no longer separates models, whereas combining two images does. The generation tasks score free-form descriptions, answers, and reports, and the reasoning tasks use paired structure and field images or the field image alone. Similarity scores for geometry description saturate across the whole roster, so the completeness of long-form technical reports carries the discriminative signal of this category. The Paired setting is the stricter reasoning test because it requires evidence from both images, and models that score perfectly on Field-only Mechanism Diagnosis do so by predicting the majority class, so a large Field-only/Paired gap marks a label shortcut.
\input{Tables/tab_text_mm}

\subsection{Inverse-Design Decision and Code Generation}
\label{app:res_inv_code}
\textbf{Tab.~\ref{tab:inv_code}} shows the per-task results for inverse-design decisions (left block) and code generation (right block), and \textbf{Tab.~\ref{tab:add_executability}} adds end-to-end execution of the generated simulation code. Answering design questions correctly does not mean producing designs or code that work. The decision tasks score the chosen parameter and direction, closed-loop success further requires a physically reasonable step within family constraints, and the code tasks score syntax, completeness, and refactoring quality. The model with the best direction accuracy is among the weakest in closed loop, and the strongest open-weight model in closed loop is one of the smallest, so direction accuracy alone overstates design competence and the Closed-Loop and Human Score columns measure the perception--design gap most directly. Code shows the same split. Syntax and completeness are near ceiling for most models, but far fewer scripts run and reproduce the reference physics, so physical fidelity, not syntax, is the open challenge in code generation.
\input{Tables/tab_inv_code}

\input{Tables/tab_add_executability}

\subsection{Robustness Analyses}
\label{app:additional_analysis}

The checks below vary the loss metric, the decoding randomness, and the prompt wording, and measure the shift from natural to photonic imagery.

\paragraph{Robustness of the Loss Metric.}
\textbf{Tab.~\ref{tab:add_loss_metric}} reports the stability of the two loss metrics under outlier trimming on the 200-sample subset. Log-MAE is the reliable ranking metric for loss. Loss targets span about $16$ orders of magnitude, so NRMSE is dominated by a few high-loss samples and reshuffles without them, whereas Log-MAE weights every order of magnitude equally.

\input{Tables/tab_add_loss_metric}

\paragraph{Sampling Variance and pass@$k$ for Stochastic Methods.}
\textbf{Tab.~\ref{tab:add_passk}} shows Self-Consistency on the GPT-4o-mini backbone as the sample budget $N$ grows from 1 to 10 at $T=0.1$. Stochastic decoding does not close the perception--design gap. Gains plateau by $N\!=\!5$, the value used in the main results, so the bottleneck lies in the physical reasoning of the model rather than in decoding variance.

\input{Tables/tab_add_passk}

\paragraph{Sensitivity to Prompt Paraphrasing.}
\textbf{Tab.~\ref{tab:add_prompt}} shows the results for the original prompt and four rewrites (instruction-reordered, terse, verbose, and independently re-authored) of four representative tasks, each repeated three times at $T=0.1$. Model rankings survive rewording. Wording matters more than decoding noise, most for small open-weight models, but the gaps between perception and design scores are an order of magnitude larger than the paraphrase spread, and leaderboard gaps larger than about .03 exceed paraphrase variation.

\input{Tables/tab_add_prompt}

\paragraph{Natural-to-Photonic Distribution Shift.}
\textbf{Tab.~\ref{tab:add_shift}} shows the Fr\'echet distance of PCFBench images to a COCO reference set in SigLIP~\cite{zhai2023sigmoid} space. PCFBench imagery lies far outside the natural-image distribution, but this shift explains perception errors, not design errors. Families farther from natural images are recognized less accurately (Spearman $\rho \approx -.6$ for Gemini-3-Flash and GPT-5), whereas closed-loop success is unrelated to the distance, so unfamiliar imagery accounts for part of the perception drop but not for the wrong design moves of models that read a fiber correctly.

\input{Tables/tab_add_shift}

\subsection{Propagation Extension}
\label{app:propagation}

This extension carries PCFBench from the transverse structure-to-field problem to propagation-level reasoning, where dispersion and nonlinearity govern pulse dynamics. It adds two binary tasks on 600 index-guiding solid-core samples within the validity range of the empirical effective-index model.

\paragraph{Tasks.}
\textbf{P1 (dispersion regime)} classifies the group-velocity dispersion at 1550\,nm as anomalous or normal. \textbf{P2 (broadening regime)} classifies the broadening of a 100\,fs input pulse as soliton-driven or quasi-linear, with pulse shape, peak power, and propagation length held fixed. Both tasks follow the image-grounded format of the inverse-design tasks and add only the scalar geometric context the model needs. Ground truth combines the empirical effective-index model with scalar GNLSE propagation at 1550\,nm, and the labels are balanced, so chance accuracy is .500. Results are reported in Sec.~\ref{sec:analysis} (\textbf{Tab.~\ref{tab:propagation_main}}).

\section{Diagnostics and Physics Grounding}
\label{app:diagnostics}

This appendix decomposes MAS behavior (Appendix~\ref{app:mas_metrics}, \ref{app:mas_diagnostic}, and \ref{app:mas_prompts}), reports cost trade-offs (Appendix~\ref{app:cost_analysis}), documents the physics rules behind the design ground truth (Appendix~\ref{app:physics_gt}), and grounds the main-text closed-loop claims in 20 family-stratified cases (Appendix~\ref{app:case_catalog}).

\subsection{MAS Baselines and Coordination Patterns}
\label{app:mas_metrics}

The MAS setting decomposes PCF inverse design into perception, understanding, reasoning, and design stages and compares the result with single-model prompting. Each agent receives the original visual evidence and the context accumulated by earlier stages. This subsection lists the 13 baselines of the main text and defines the pipeline-level metrics.

\paragraph{MAS Baselines.}
We implement 13 representative methods with MASLab~\cite{ye2025maslab}, a unified framework for LLM-based multi-agent systems. \textbf{Tab.~\ref{tab:mas_baselines}} groups them by coordination paradigm and sketches the workflow of each. Self-Consistency votes over $N{=}5$ responses at $T{=}0.1$.

\input{Tables/tab_mas_baselines}

The 12 end-to-end methods receive identical inputs (the structure and mode-field images and a unified query) and return a structured design recommendation. The query carries the same information and output constraints as the single-VLM inverse-design prompts (Tasks 20--23 in Appendix~\ref{app:tasks}): the family name, the current confinement loss and parameter values, the parameter candidate set, and the required output format (parameter name, direction, and a concrete numeric proposal).

\begin{taskprompt}[MAS Baseline Query (End-to-End, All 12 Methods)]
\sysrole\ You are an expert in photonic crystal fibers.

\prompthead{Input}\ \imgtoken{structure}\ \imgtoken{field@1550\,nm}

\prompthead{Context}\ This is a \promptvar{family} PCF. The current confinement loss at 1550\,nm is \promptvar{loss:.2e}\,dB/km. Key current parameters: pitch = \promptvar{pitch}\,$\mu$m, hole/rod diameter = \promptvar{hole\_d}\,$\mu$m.

\prompthead{Instruction}\ Your task:
\begin{enumerate}[leftmargin=1.6em, itemsep=0pt, topsep=1pt]
\item \textbf{Parameter Selection}: Which geometric parameter should be adjusted first to reduce loss? Choose from: pitch ($\Lambda$), hole diameter ($d$), strut thickness ($t$), core diameter, ring radius, number of rings, or gap width.
\item \textbf{Direction}: Should that parameter be increased or decreased?
\item \textbf{New Value}: Propose a specific new numeric value for that parameter in $\mu$m.
\end{enumerate}

\prompthead{Output Format}\par
\texttt{Parameter: <name>}\\
\texttt{Direction: <increase or decrease>}\\
\texttt{New Value: <number> $\mu$m}\\
\texttt{Reasoning: <brief physics justification>}
\promptimages{Structure $+$ field 1550\,nm (concatenated).}
\promptextraction{Answers are read from the \texttt{Parameter:}, \texttt{Direction:}, and \texttt{New Value:} lines, and otherwise by keyword matching (same extraction as single-VLM Tasks 20--23).}
\promptmetric{PSel (Parameter Selection Accuracy), CL (Closed-Loop Success Rate), HS (Human Score).}
\end{taskprompt}

\noindent
The MAS baselines coordinate through text, take the unified query without explicit stage boundaries, and are evaluated end to end. The Sequential Pipeline instead splits the task into four stages with vision-enabled agents (prompts in Appendix~\ref{app:mas_prompts}) and runs in two modes: \textbf{Oracle}, where each stage receives ground-truth context from prior stages, and \textbf{Cascaded}, where each stage receives predicted context. Three comparisons follow: the cascaded Sequential Pipeline against the MAS baselines (structured domain decomposition against general-purpose coordination), Oracle against Cascaded (the cost of error propagation), and MAS methods against single-VLM baselines (the value of multi-agent coordination).

\paragraph{Pipeline-Level Metrics.}
Beyond the per-stage metrics (which are identical to the single-task metrics defined in Appendix~\ref{app:tasks}), we introduce four pipeline-level metrics for the Sequential Pipeline:

\underline{\textit{End-to-End Design Success Rate.}} The fraction of samples for which the full pipeline produces a correct design action, requiring both correct perception and correct design direction:
\begin{equation}
\mathrm{E2E\;Success} = \frac{1}{N}\sum_{i=1}^{N}
\mathds{1}\!\big[\hat{y}^{(1)}_i = y^{(1)}_i\big]
\cdot
\mathds{1}\!\big[\hat{d}^{(4)}_i = d^{*}_i\big],
\end{equation}
where $\hat{y}^{(1)}_i$ is the Stage~1 family prediction, $y^{(1)}_i$ is the ground truth, $\hat{d}^{(4)}_i$ is the Stage~4 design direction, and $d^{*}_i$ is the physics-based correct direction.

\underline{\textit{Error Cascade Rate.}} The conditional probability that the design stage fails given that perception was incorrect, quantifying the severity of upstream error propagation:
\begin{equation}
\mathrm{Cascade} = 1 - \frac{
\sum_{i:\,\hat{y}^{(1)}_i \neq y^{(1)}_i}
\mathds{1}\!\big[\hat{d}^{(4)}_i = d^{*}_i\big]
}{
\sum_{i}\mathds{1}\!\big[\hat{y}^{(1)}_i \neq y^{(1)}_i\big]
}.
\end{equation}
A cascade rate of 1.0 means that every perception error leads to a design failure, whereas a rate near 0 indicates that the design agent can compensate for upstream mistakes.

\underline{\textit{Pipeline Consistency.}} The fraction of samples whose proposed design action is logically consistent with the \emph{predicted} (not ground-truth) upstream context. A sample counts as consistent when its direction prediction matches the physics rule for the \emph{predicted family}:
\begin{equation}
\mathrm{Consistency} = \frac{1}{N}\sum_{i=1}^{N}
\mathds{1}\!\Big[\hat{d}^{(4)}_i =
f_\mathrm{phys}\!\big(\hat{y}^{(1)}_i,\;\hat{p}^{(4)}_i,\;
\hat{\mathcal{L}}^{(2)}_i\big)\Big],
\end{equation}
where $f_\mathrm{phys}$ is the physics direction function (Appendix~\ref{app:physics_gt}), $\hat{p}^{(4)}_i$ is the selected parameter, and $\hat{\mathcal{L}}^{(2)}_i$ is the predicted loss from Stage~2. High consistency indicates that the design agent respects the upstream analysis even when that analysis is wrong.

\underline{\textit{Design Accuracy Conditioned on Correct Perception.}} To separate design from perception errors:
\begin{equation}
\mathrm{Acc}_{4|1} =
\frac{\sum_{i:\,\hat{y}^{(1)}_i = y^{(1)}_i}
      \mathds{1}\!\big[\hat{d}^{(4)}_i = d^{*}_i\big]}
     {\sum_{i}\mathds{1}\!\big[\hat{y}^{(1)}_i = y^{(1)}_i\big]}.
\end{equation}
Comparing $\mathrm{Acc}_{4|1}$ (accuracy given correct perception) with unconditional accuracy separates the part of the perception--design gap caused by error propagation from intrinsic design limitations.

\subsection{Sequential Pipeline Decomposition}
\label{app:mas_diagnostic}

\textbf{Tab.~\ref{tab:mas_stages}} reports the Understanding and Reasoning stage accuracies, PSel and CL, and the pipeline-level metrics of the Sequential Pipeline under Oracle and Cascaded modes, and its $\Delta$ row gives the cost of error propagation. It applies to staged methods only and is reported apart from the cross-method leaderboard (\textbf{Tab.~\ref{tab:mas_methods}}).

\input{Tables/tab_mas_stages}

\subsection{Cost-Performance and Token Budget}
\label{app:cost_analysis}

\begin{figure}[H]
  \centering
  \includegraphics[width=\textwidth]{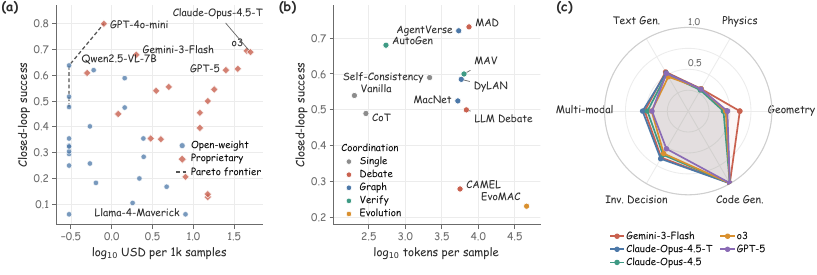}
  \caption{(a) Cost-performance Pareto frontier. (b) MAS token budget vs.\ closed-loop success (GPT-4o-mini backbone). (c) Category profiles of the top-5 VLMs.}
  \label{fig:d3_panels}
\end{figure}

\textbf{Fig.~\ref{fig:d3_panels}} shows design quality against cost, token use, and per-category profile. Higher cost, more tokens, and stronger generalist models do not yield better designs. Inexpensive models set the Pareto frontier while the strongest proprietary models pay up to two orders of magnitude more and stay below it (panel a), AutoGen stays close to the best debate method at a fraction of its tokens while EvoMAC spends the most and designs the worst (panel b), and the top-ranked models owe their global lead to geometry and design-decision tasks rather than to physics competence (panel c). Physical grounding, not model capacity or interaction budget, is therefore the bottleneck.

\subsection{Sequential Pipeline Stage Prompts}
\label{app:mas_prompts}

The four stages share the same system prompt and decoding configuration as the single-task evaluation (Appendix~\ref{app:tasks_shared}). Each stage receives the prior stages' \emph{extracted} answers as context (Oracle mode substitutes ground-truth values). Stage~4's ground-truth direction comes from the physics rules of the single-VLM Design Direction task (Appendix~\ref{app:physics_gt}).

\begin{taskprompt}[MAS Sequential Pipeline: Stage Prompts (Stages 1--4)]
\sysrole\ You are an expert in photonic crystal fibers. \hfill\textit{(shared by all four stages)}
\par\smallskip{\color{promptboxframe!35}\hrule}\smallskip
\textcolor{promptboxframe}{\textbf{Stage 1: Perception}}\par
{\raggedright \prompthead{Input}\ \imgtoken{structure}\quad (at most 256 output tokens)\par}

\prompthead{Instruction}\ Analyze this photonic crystal fiber cross-section image and provide:
\begin{enumerate}[leftmargin=1.6em, itemsep=0pt, topsep=1pt]
\item Family type (choose from: Solid-Core Hexagonal, Hollow-Core PBG, Anti-Resonant ARF, High-Birefringence PM, Suspended Cobweb, Multi-Core, Square/Rectangular, Vortex/Ring-Core, Fractal/Quasi-Crystal, Octagonal Lattice, Dual-Core Coupler, Nested Anti-Resonant, Elliptical Core, Hybrid Plasmonic, All-Solid Bandgap, Twisted Hexagonal, Graded-Index Multi-Ring)
\item Estimated pitch in micrometers
\item Number of cladding rings
\item Lattice topology (hexagonal, square, octagonal, kagome, or other)
\end{enumerate}
\prompthead{Output Format}\ \texttt{Family: <name>} / \texttt{Pitch: <value> um} / \texttt{Rings: <integer>} / \texttt{Topology: <type>} (one field per line).
\par\smallskip{\color{promptboxframe!35}\hrule}\smallskip
\textcolor{promptboxframe}{\textbf{Stage 2: Understanding}}\par
{\raggedright \prompthead{Input}\ \imgtoken{field@1550\,nm}\par \prompthead{Upstream Context}\ \promptvar{perception\_context} (Stage 1)\par}

\prompthead{Instruction}\ Given the perception summary above, analyze the mode-field image at 1550\,nm and provide:
\begin{enumerate}[leftmargin=1.6em, itemsep=0pt, topsep=1pt]
\item Estimated confinement loss in dB/km (use scientific notation)
\item Mode type: single-mode, multi-mode, or leaky
\item Primary guidance mechanism (e.g., modified TIR, photonic bandgap, anti-resonant reflection, inhibited coupling)
\end{enumerate}
\prompthead{Output Format}\ \texttt{Loss: <value> dB/km} / \texttt{Mode: <type>} / \texttt{Mechanism: <name>} (one per line).
\par\smallskip{\color{promptboxframe!35}\hrule}\smallskip
\textcolor{promptboxframe}{\textbf{Stage 3: Reasoning}}\par
{\raggedright \prompthead{Input}\ \imgtoken{structure}\ \imgtoken{field@1550\,nm}\quad (at most 512 output tokens)\par \prompthead{Upstream Context}\ \promptvar{perception\_context} / \promptvar{understanding\_context} (Stages 1--2)\par}

\prompthead{Instruction}\ Using the upstream summaries, examine both the structure and field images together and provide:
\begin{enumerate}[leftmargin=1.6em, itemsep=0pt, topsep=1pt]
\item Which wavelength (1310\,nm or 1550\,nm) likely has higher confinement loss, and why?
\item What is the dominant loss mechanism for this fiber?
\item A brief evidence-based explanation (2--3 sentences) citing visual features.
\end{enumerate}
\prompthead{Output Format}\ \texttt{Higher loss wavelength: <1310 or 1550>} / \texttt{Dominant mechanism: <name>} / \texttt{Explanation: <text>}.
\par\smallskip{\color{promptboxframe!35}\hrule}\smallskip
\textcolor{promptboxframe}{\textbf{Stage 4: Design}}\par
{\raggedright \prompthead{Input}\ \imgtoken{structure}\ \imgtoken{field@1550\,nm}\quad (at most 256 output tokens)\par}

{\raggedright \prompthead{Upstream Context}\ \promptvar{perception\_context} / \promptvar{understanding\_context} / \promptvar{reasoning\_context} (Stages 1--3). The fiber currently has a confinement loss of \promptvar{loss\_value}\,dB/km.\par}

\prompthead{Instruction}\ Propose a design improvement:
\begin{enumerate}[leftmargin=1.6em, itemsep=0pt, topsep=1pt]
\item Which parameter should be adjusted first? (pitch, hole diameter, strut thickness, core diameter, ring radius, gap width, or number of rings)
\item Should it be increased or decreased?
\item Suggest a specific new value for the parameter you selected (if the current value is known).
\end{enumerate}
\prompthead{Output Format}\ \texttt{Parameter: <name>} / \texttt{Direction: <increase or decrease>} / \texttt{Reasoning: <brief physics justification>}.
\end{taskprompt}

\input{sections/app_physics_rules}

\subsection{Per-Case Inverse-Design Findings}
\label{app:case_catalog}

The 20-case closed-loop sweep extends the two examples of the main text (Sec.~\ref{sec:case_study}) across the 18 canonical families, with cases~1--8 forming contrast pairs and cases~9--20 covering the remaining 12 families with one case each. Each case compares single-parameter proposals from 11 systems (10 single VLMs plus the Sequential Pipeline MAS) under matched FDTD conditions ($\lambda = 1550$\,nm), with $\Delta = 10\log_{10}(L_\text{new}/L_\text{base})$\,dB measuring whether the change improves (negative) or degrades (positive) the original design.

\begin{figure}[H]
\centering\includegraphics[width=\textwidth]{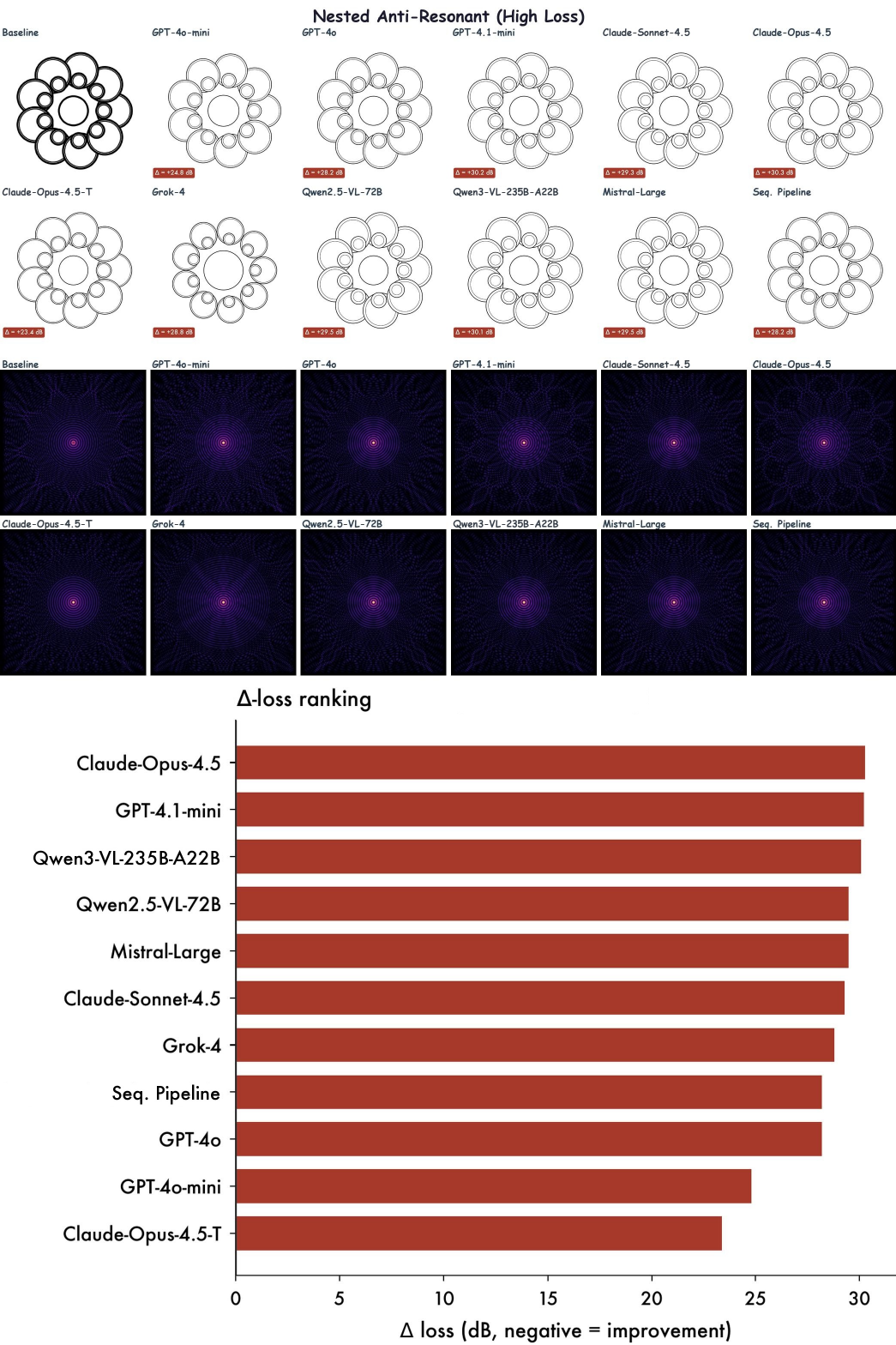}
\caption{Case~2: Nested Anti-Resonant (High Loss). All systems regress because a coordinated dual-gap adjustment is required.}\label{fig:case02}
\end{figure}

\begin{figure}[H]
\centering\includegraphics[width=\textwidth]{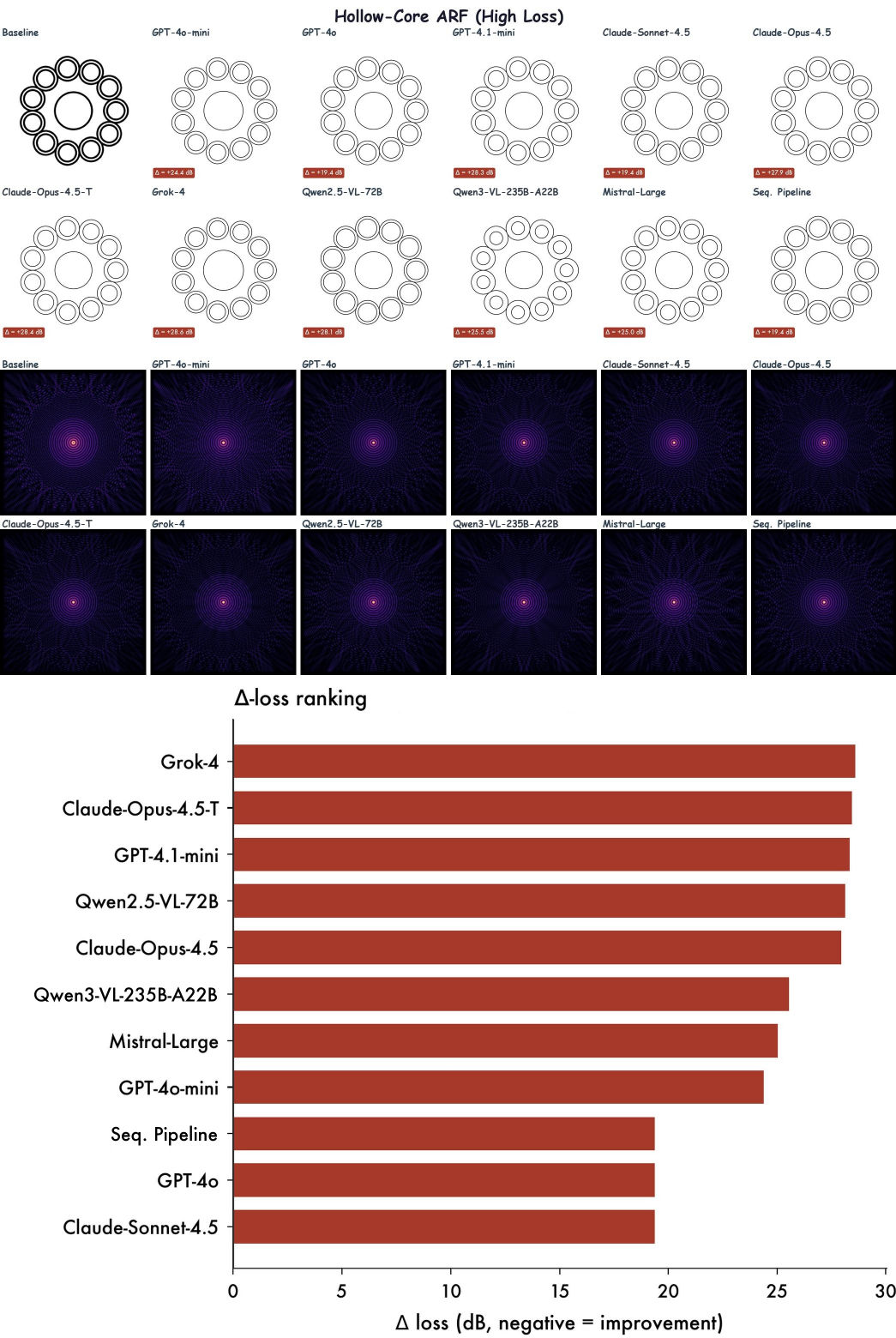}
\caption{Case~3: Hollow-Core ARF (High Loss). A single-ring topology reproduces the case~2 failure, ruling out nested geometry as the sole cause.}\label{fig:case03}
\end{figure}

\begin{figure}[H]
\centering\includegraphics[width=\textwidth]{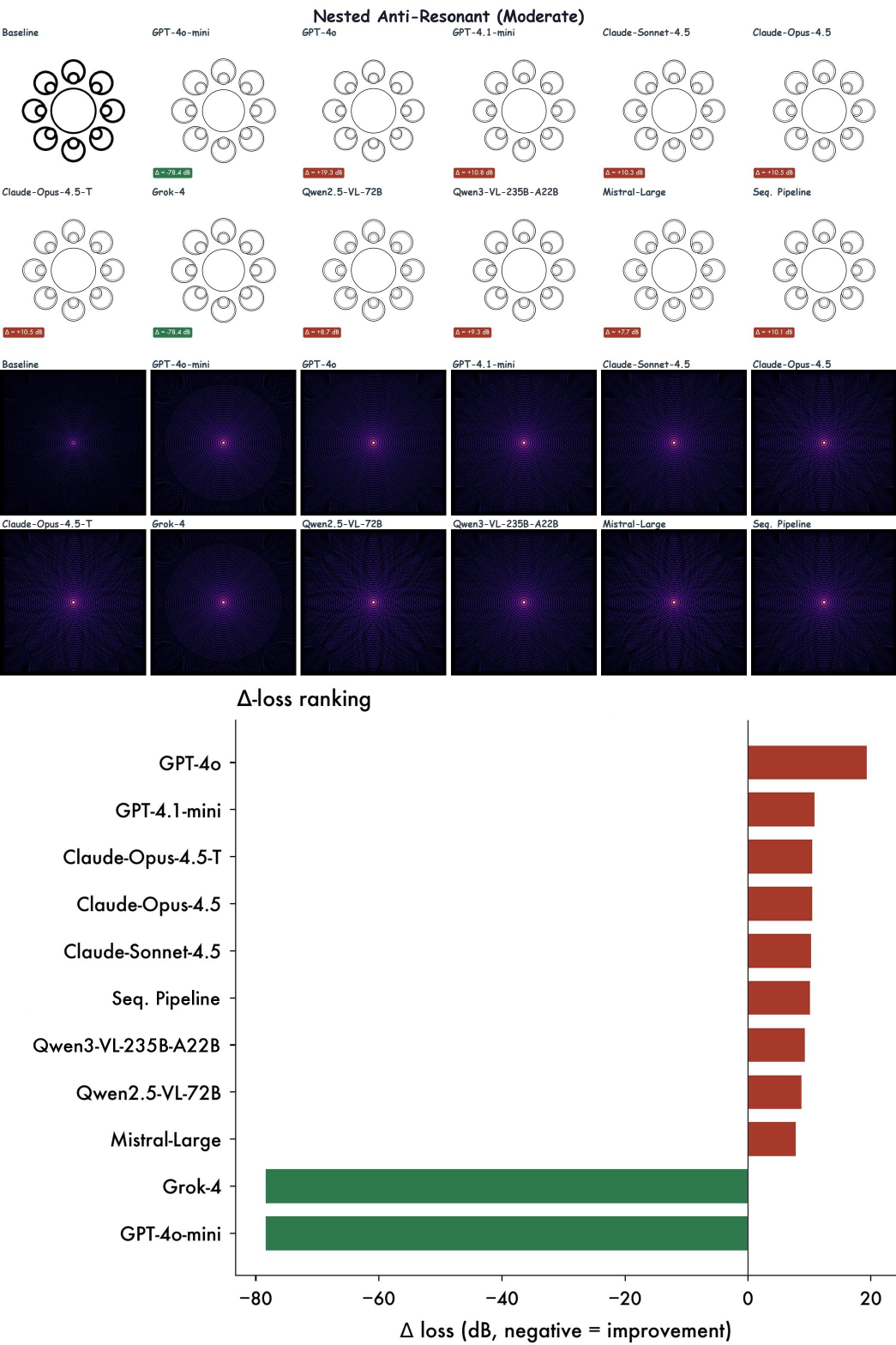}
\caption{Case~4: Nested Anti-Resonant (Moderate Loss). Partial recovery (2/11 improve).}\label{fig:case04}
\end{figure}

\begin{figure}[H]
\centering\includegraphics[width=\textwidth]{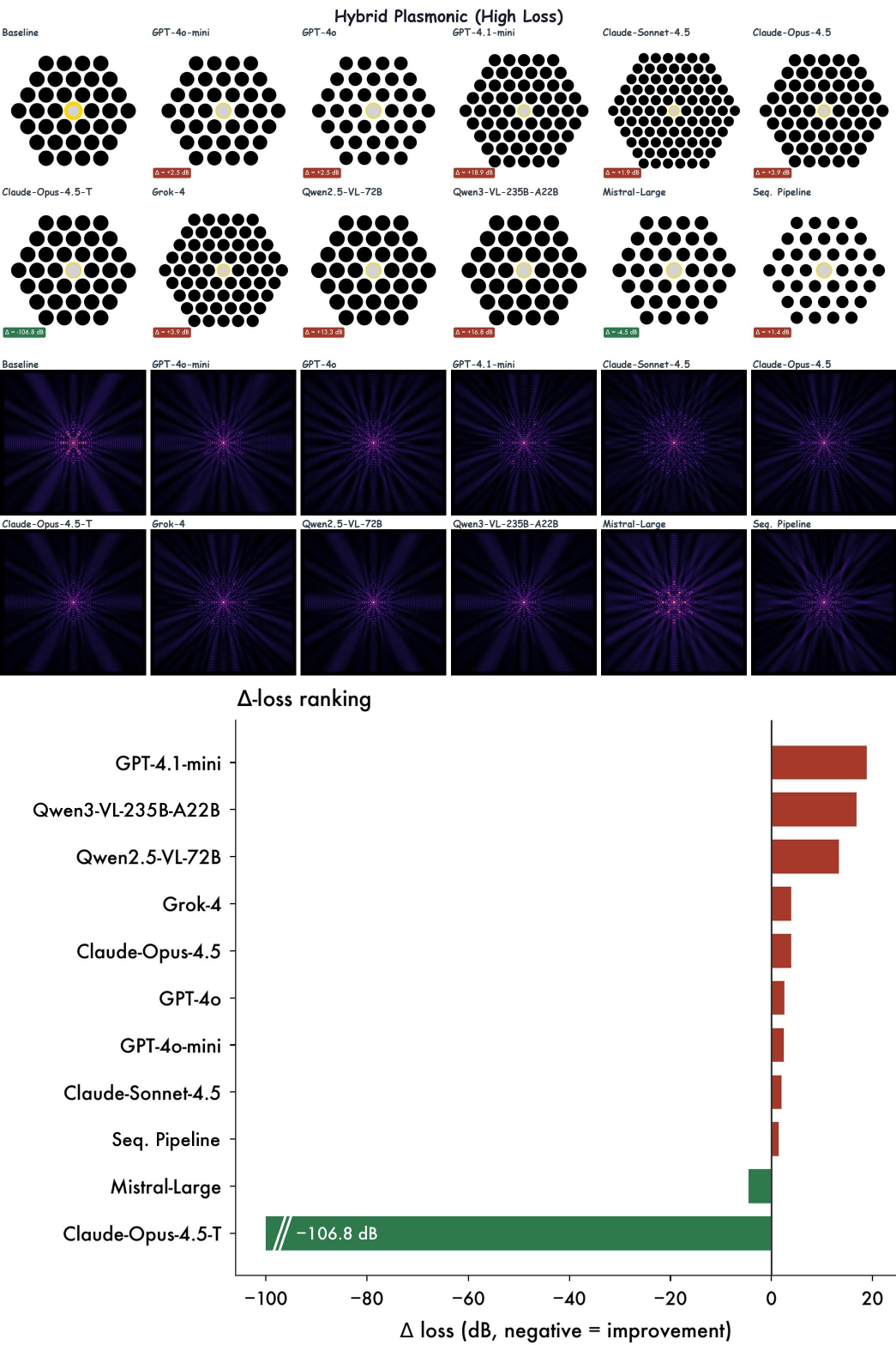}
\caption{Case~5: Hybrid Plasmonic (High Loss). Occasional, unreliable success (2/11).}\label{fig:case05}
\end{figure}

\begin{figure}[H]
\centering\includegraphics[width=\textwidth]{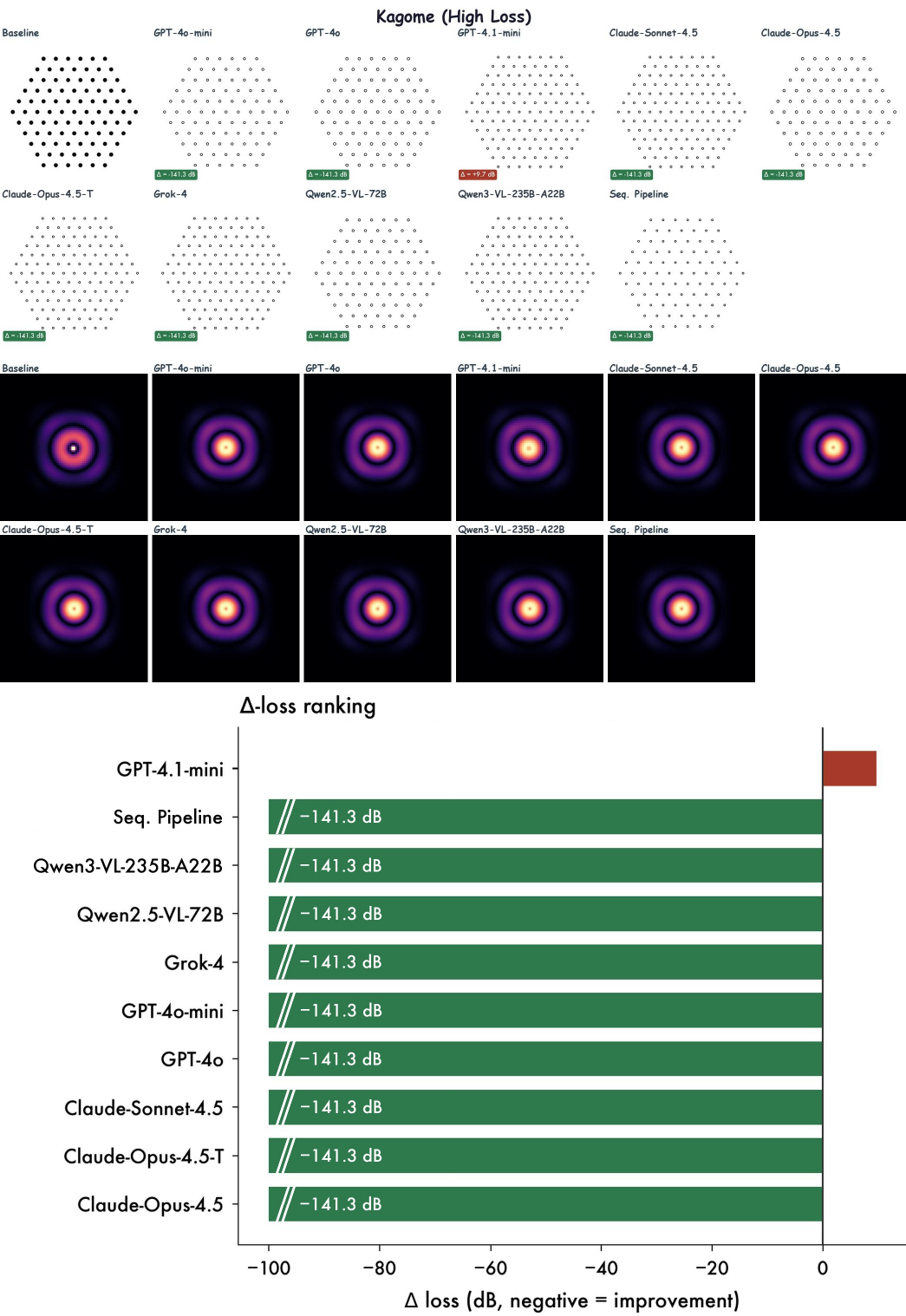}
\caption{Case~6: Kagome (High Loss). Strongest success in the catalog: 9/10 systems reach the numerical floor through different edits.}\label{fig:case06}
\end{figure}

\begin{figure}[H]
\centering\includegraphics[width=\textwidth]{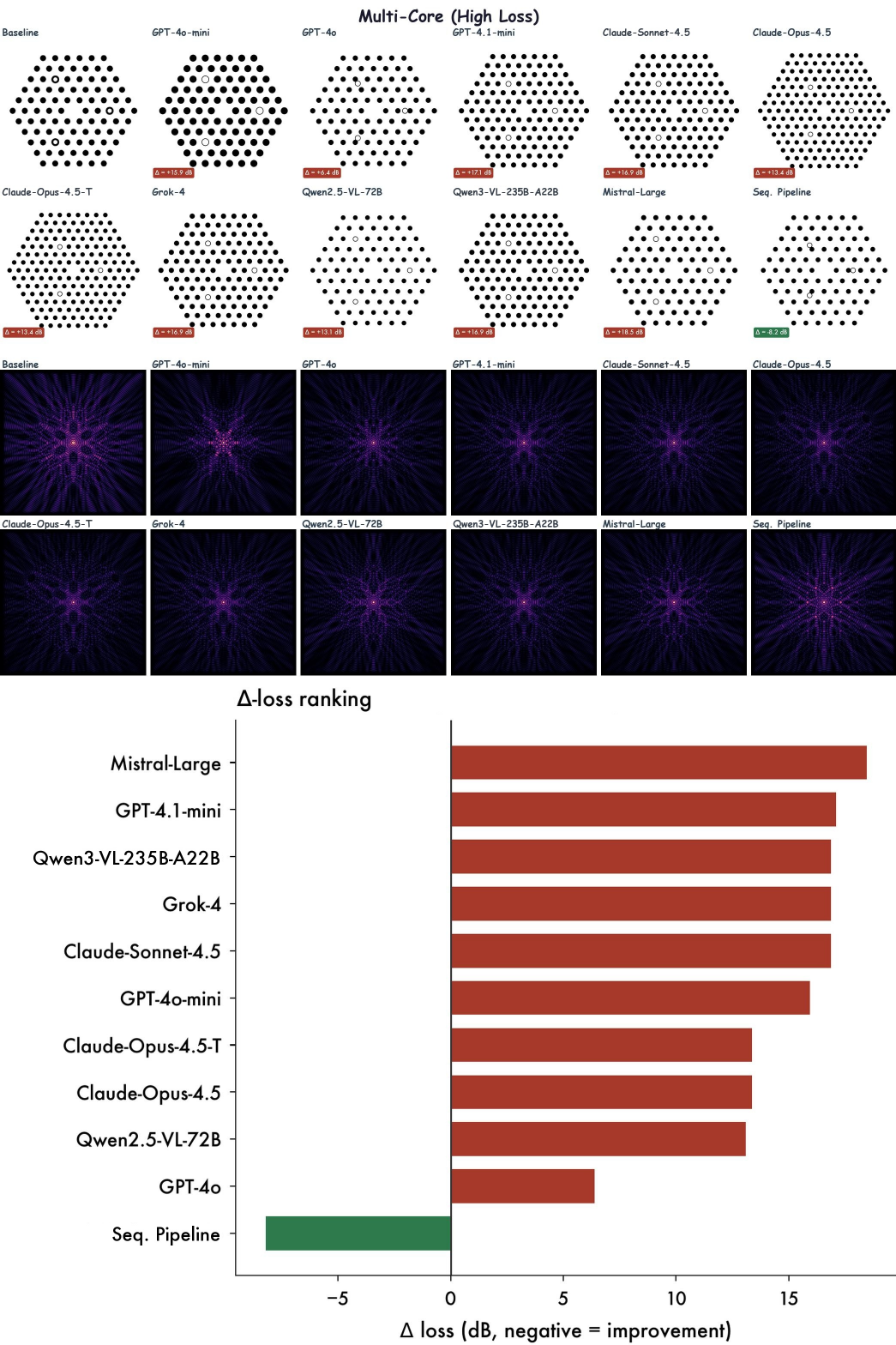}
\caption{Case~7: Multi-Core (High Loss). Coupling reasoning fails (1/11 marginal improvement).}\label{fig:case07}
\end{figure}

\begin{figure}[H]
\centering\includegraphics[width=\textwidth]{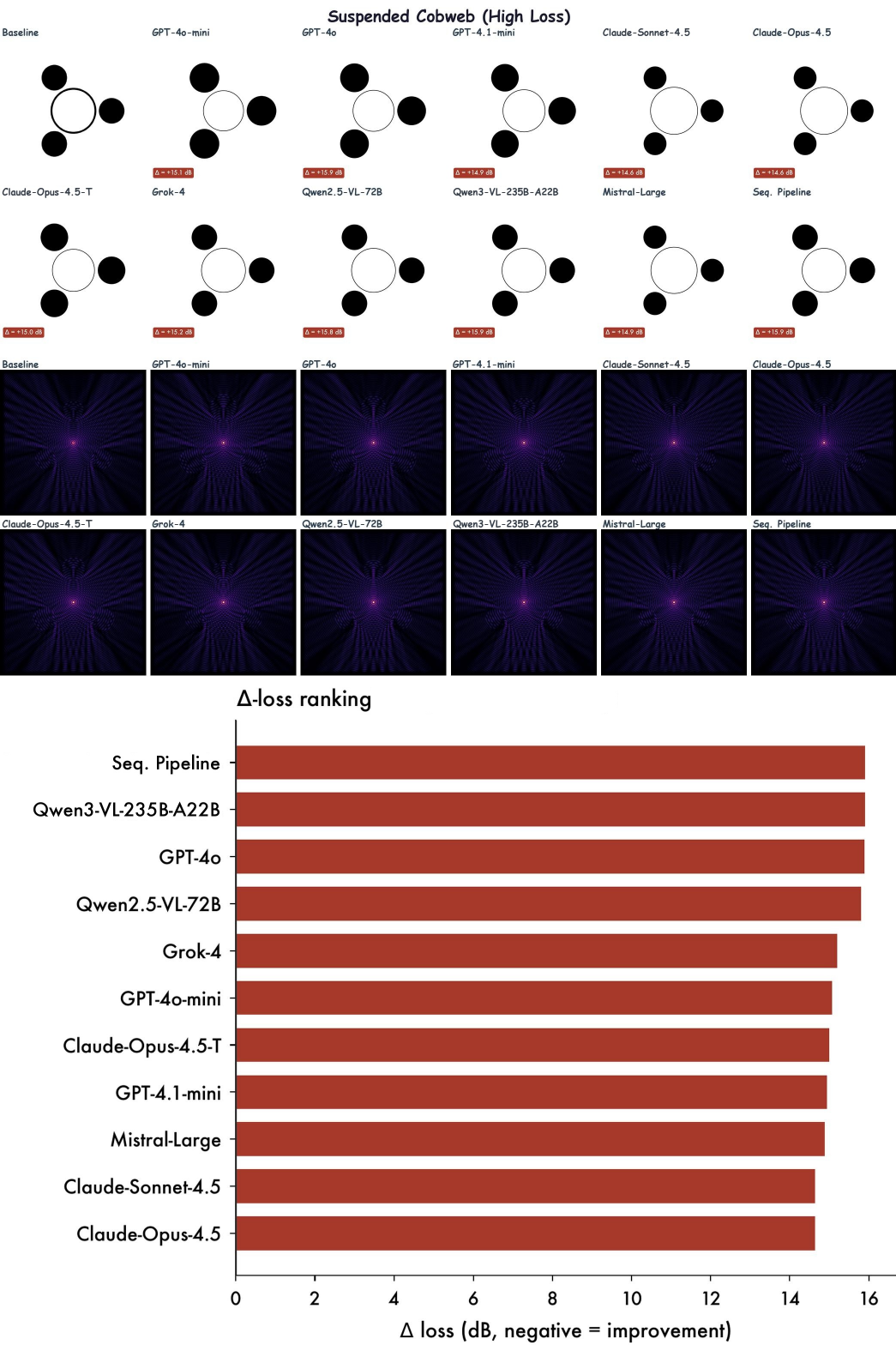}
\caption{Case~8: Suspended Cobweb (High Loss). Largest baseline loss, yet no system improves, and most err by reducing strut thickness in the wrong regime.}\label{fig:case08}
\end{figure}

\begin{figure}[H]
\centering\includegraphics[width=\textwidth]{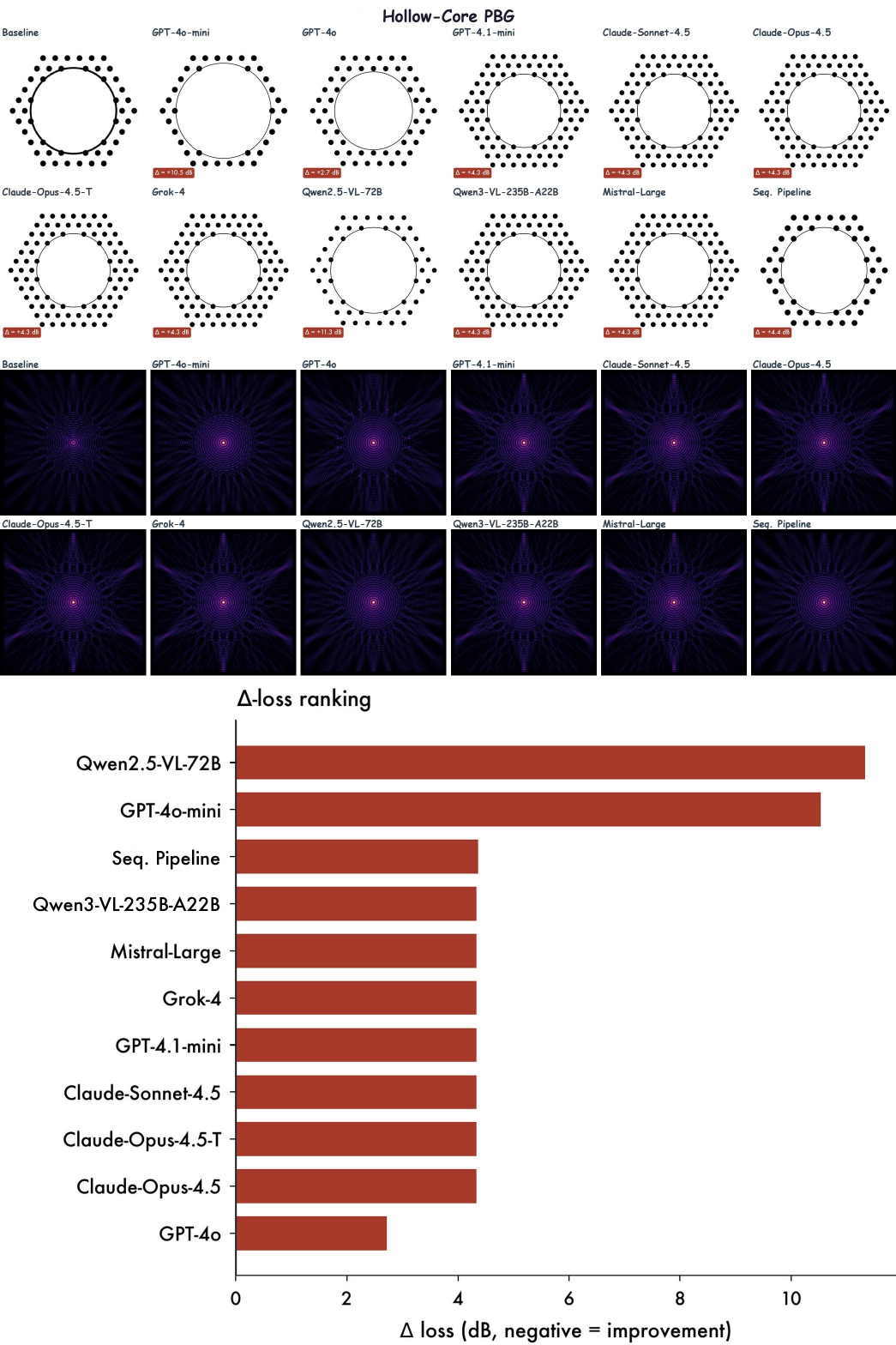}
\caption{Case~9: Hollow-Core PBG. Near-optimal baseline, so all edits regress.}\label{fig:case09}
\end{figure}

\begin{figure}[H]
\centering\includegraphics[width=\textwidth]{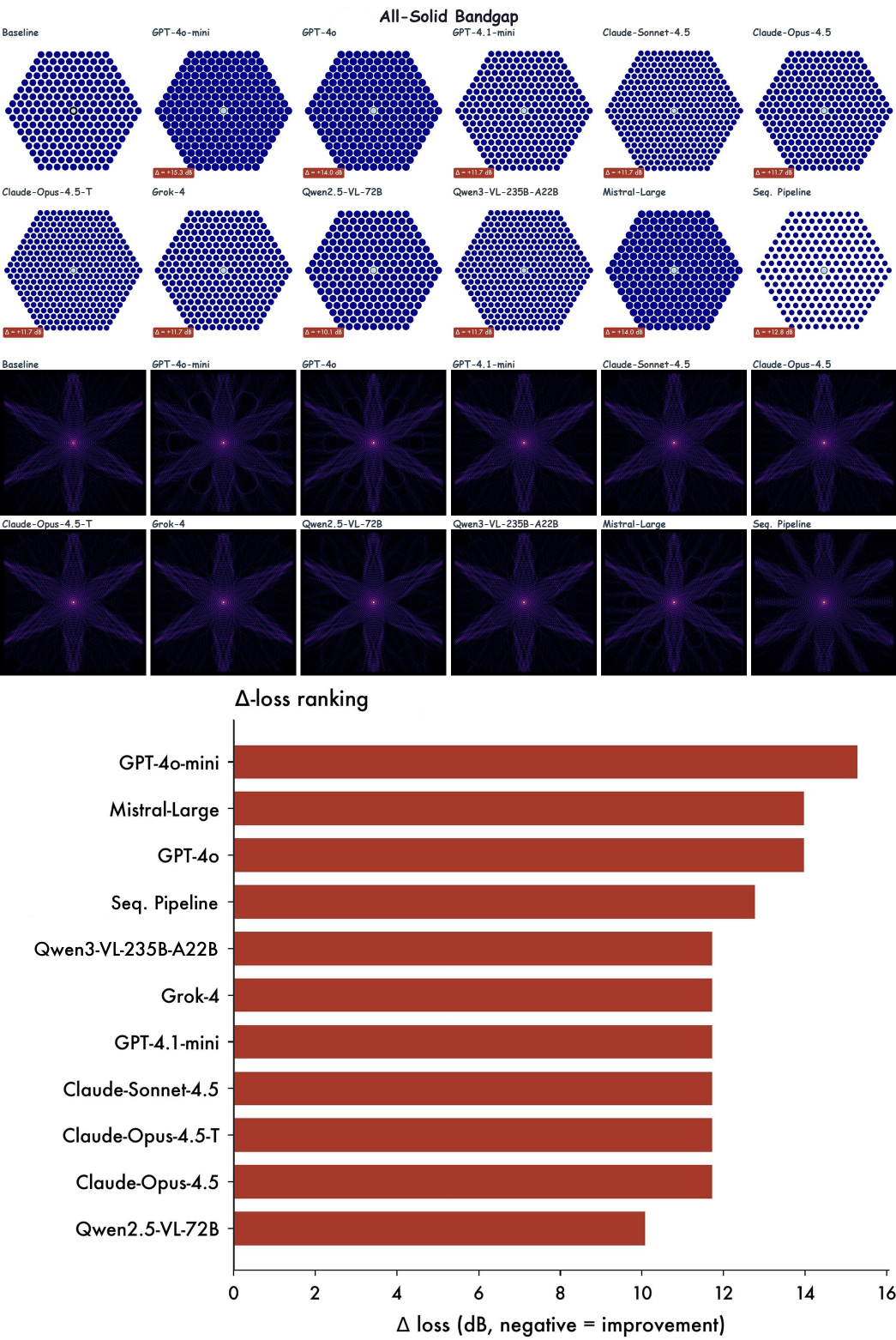}
\caption{Case~10: All-Solid Bandgap. Simulator-floor baseline leaves no headroom.}\label{fig:case10}
\end{figure}

\begin{figure}[H]
\centering\includegraphics[width=\textwidth]{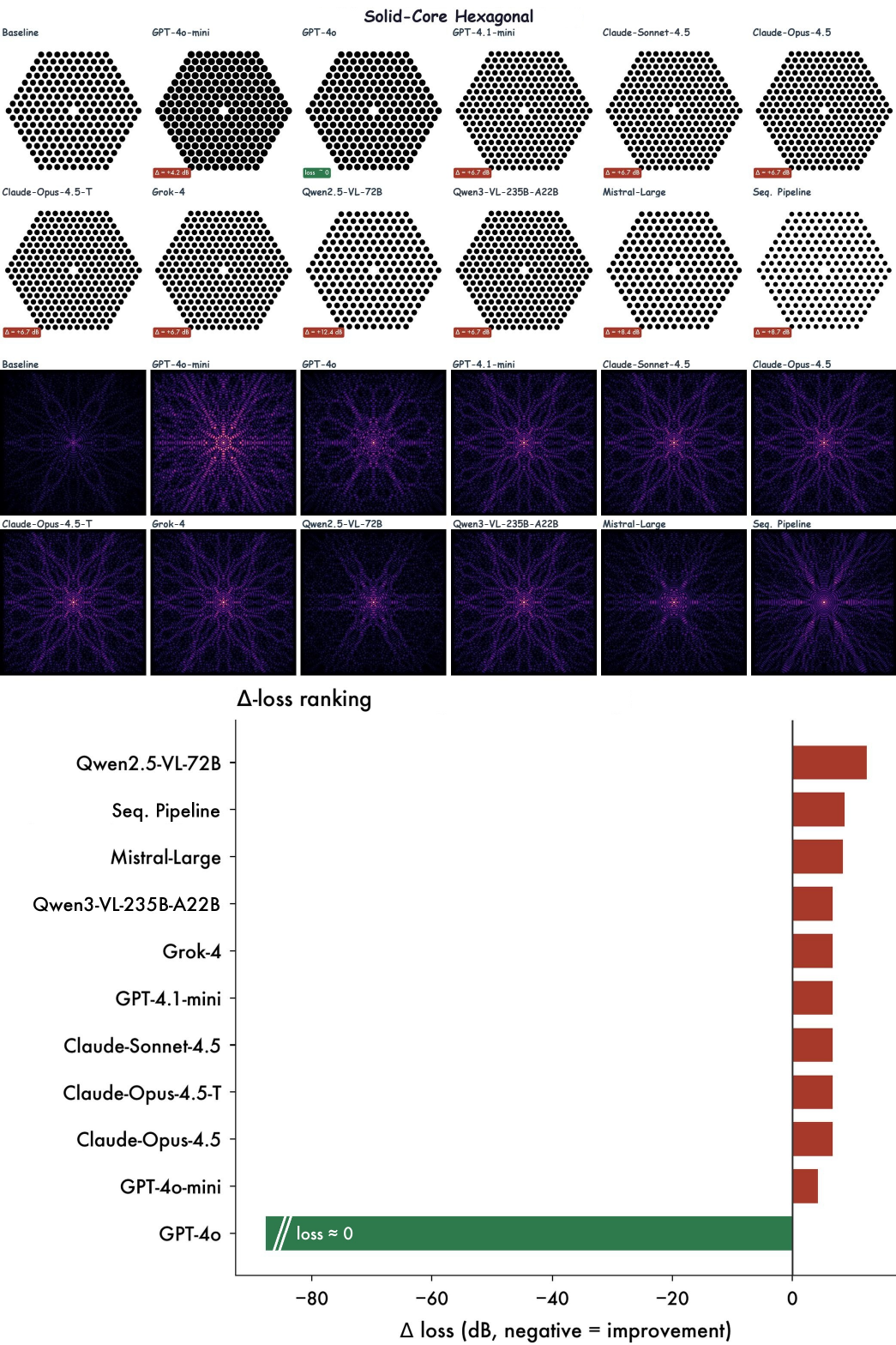}
\caption{Case~11: Solid-Core Hexagonal. Only one anomalous improvement.}\label{fig:case11}
\end{figure}

\begin{figure}[H]
\centering\includegraphics[width=\textwidth]{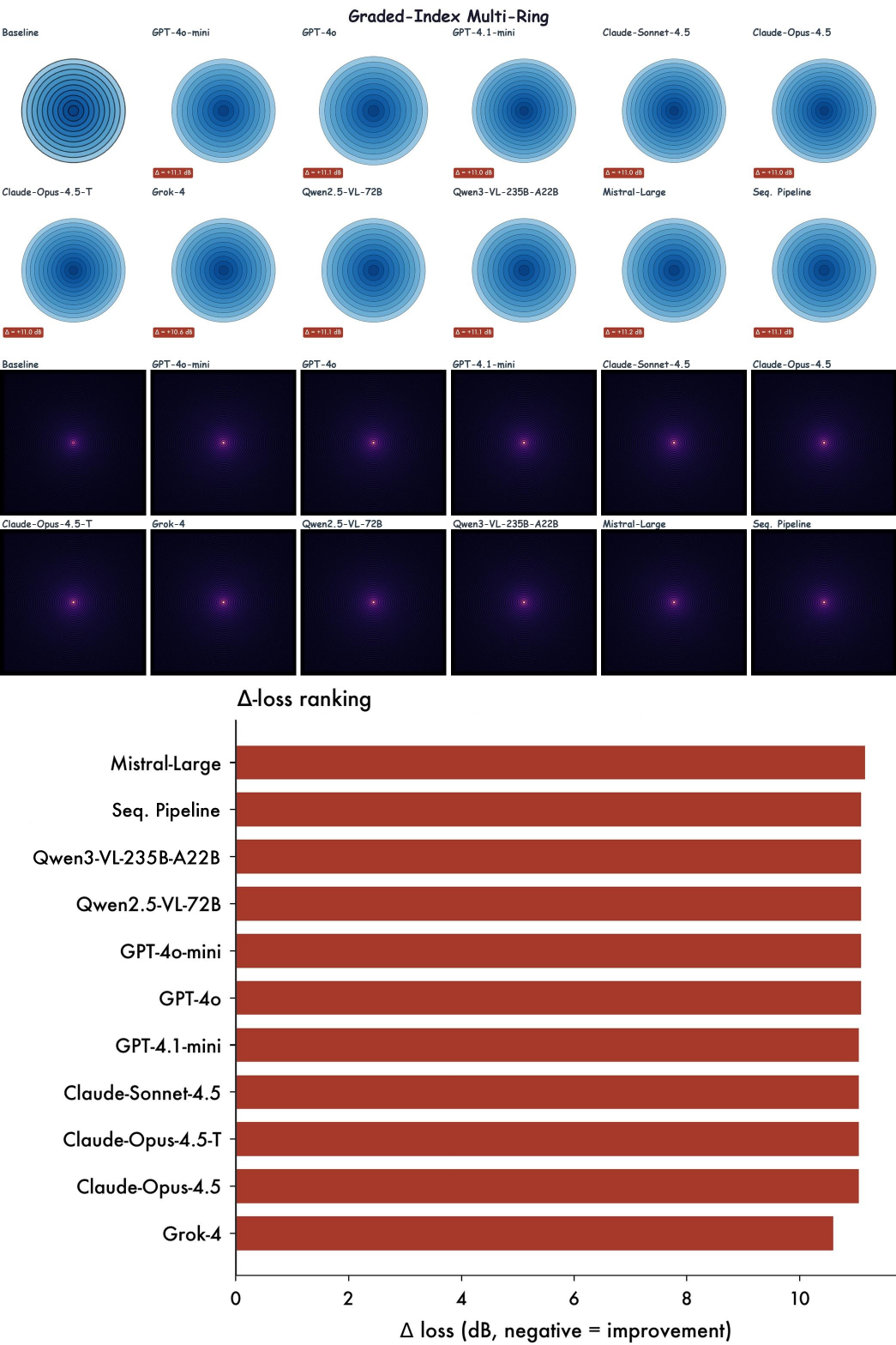}
\caption{Case~12: Graded-Index Multi-Ring. Part of the near-optimal cluster, with 0/11 improving.}\label{fig:case12}
\end{figure}

\begin{figure}[H]
\centering\includegraphics[width=\textwidth]{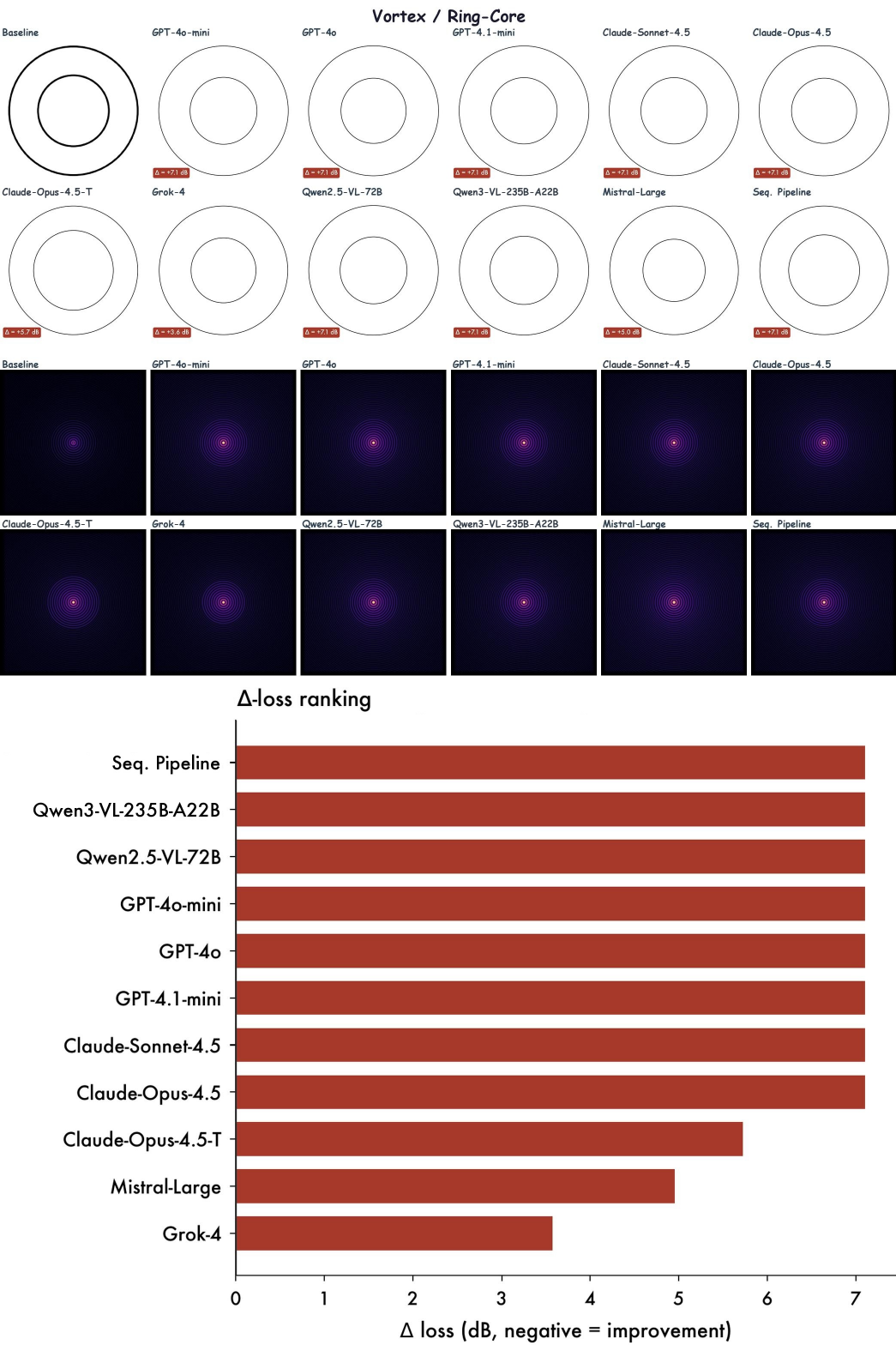}
\caption{Case~14: Vortex / Ring-Core. Near-floor baseline leaves no headroom (0/11 improve).}\label{fig:case14}
\end{figure}

\begin{figure}[H]
\centering\includegraphics[width=\textwidth]{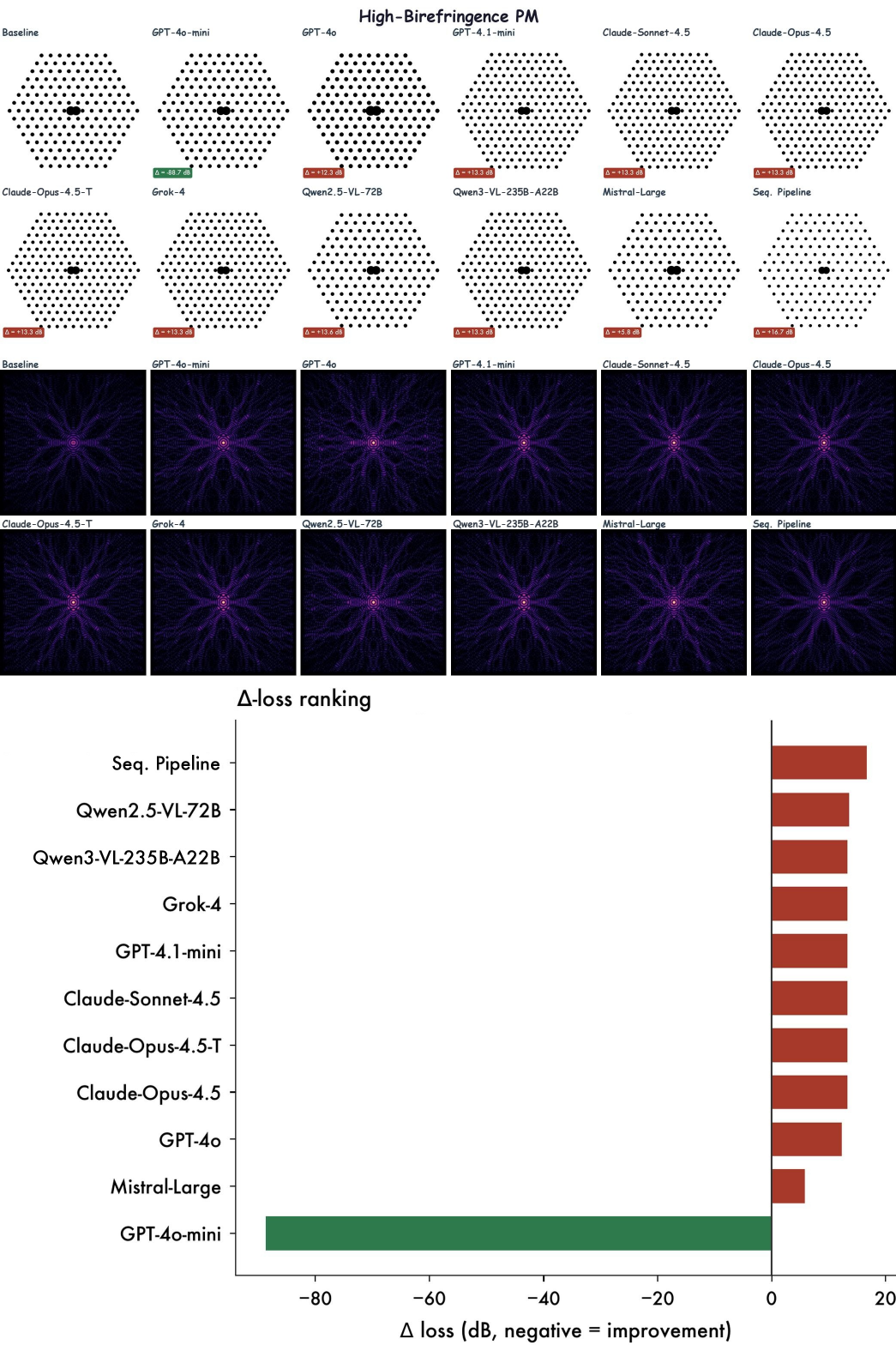}
\caption{Case~15: High-Birefringence PM. Occasional success (1/11) but most systems regress.}\label{fig:case15}
\end{figure}

\begin{figure}[H]
\centering\includegraphics[width=\textwidth]{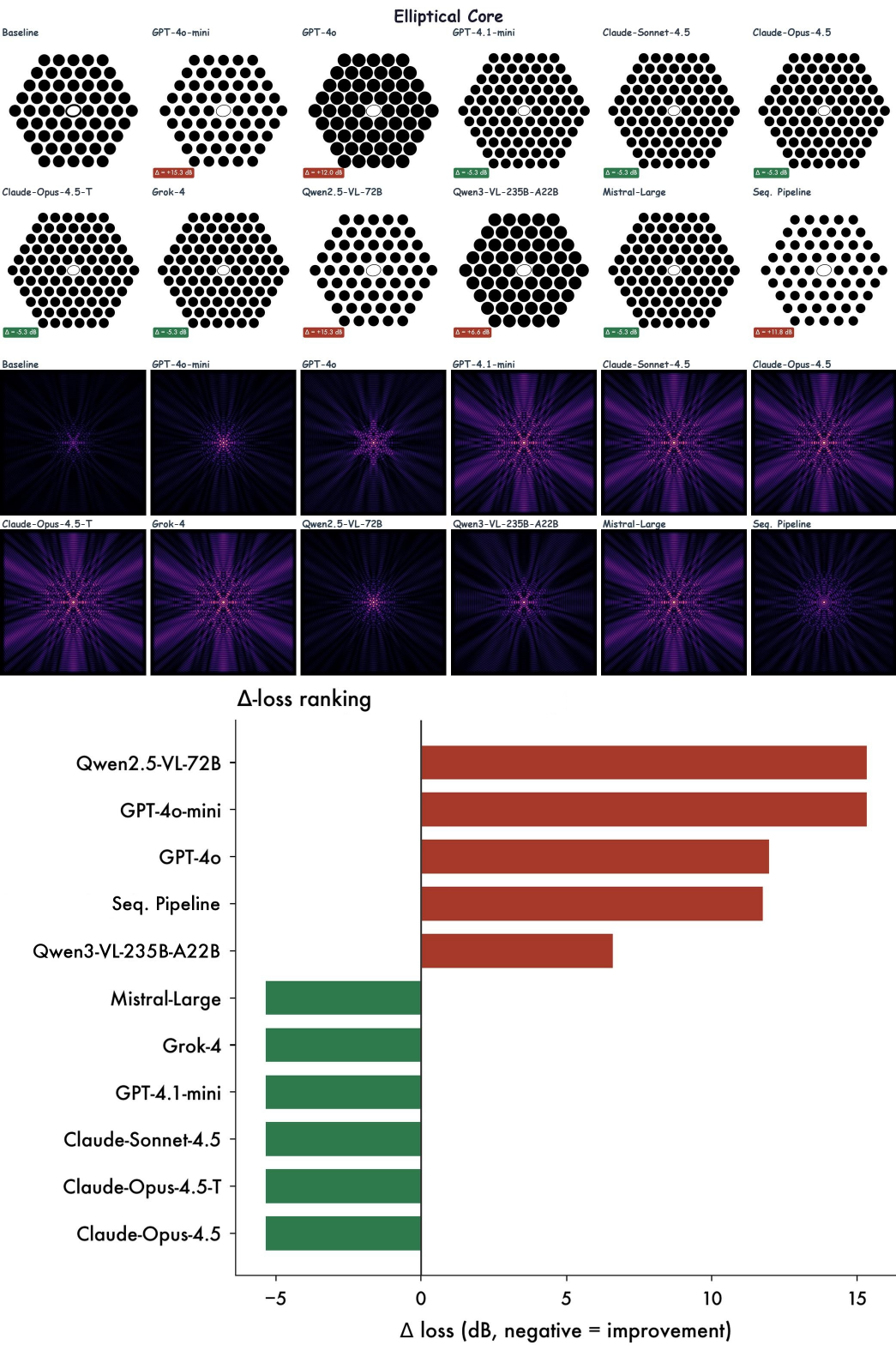}
\caption{Case~16: Elliptical Core. Most evenly contested case (6/11 improve).}\label{fig:case16}
\end{figure}

\begin{figure}[H]
\centering\includegraphics[width=\textwidth]{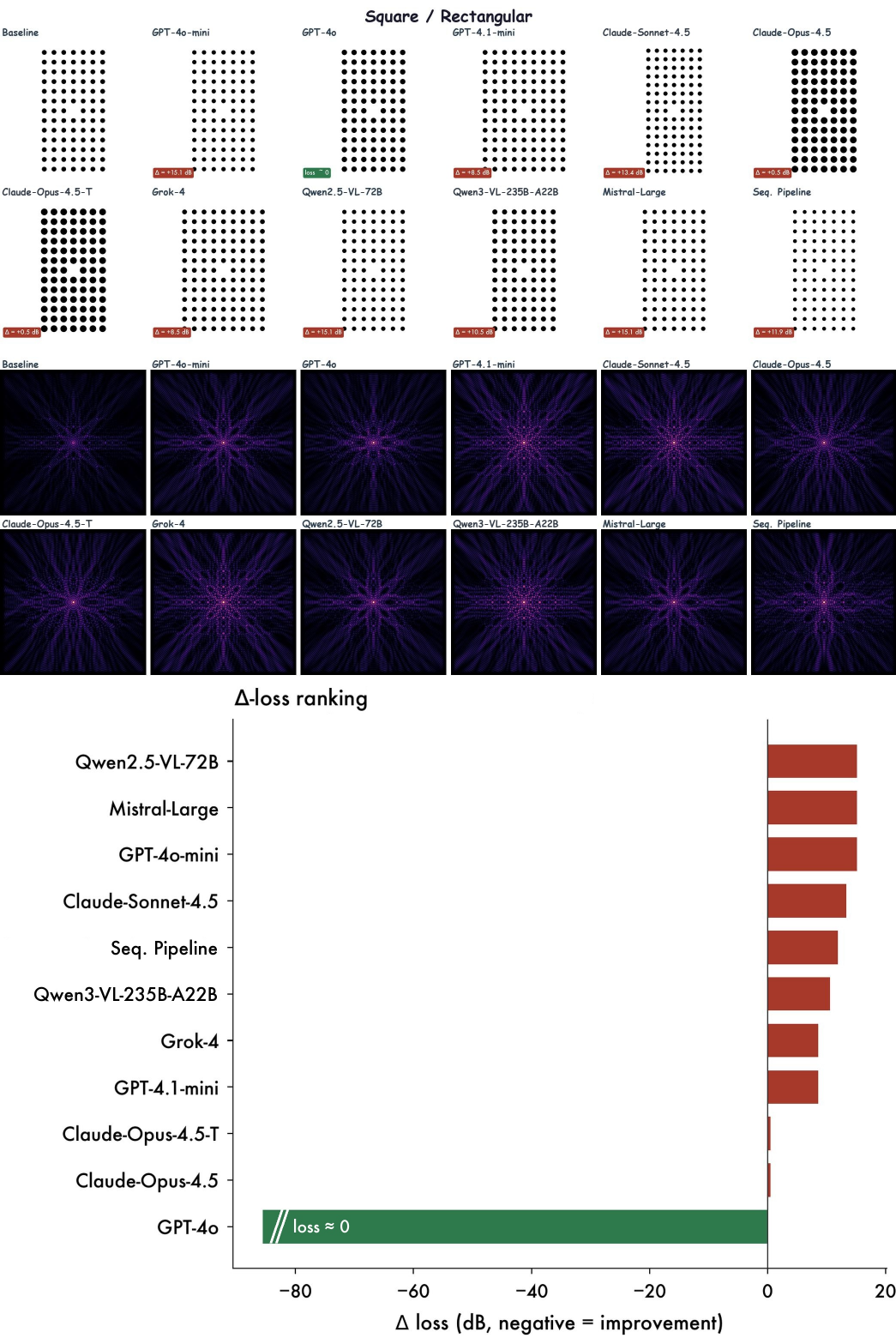}
\caption{Case~17: Square / Rectangular. One improvement, typical of the near-optimal cluster.}\label{fig:case17}
\end{figure}

\begin{figure}[H]
\centering\includegraphics[width=\textwidth]{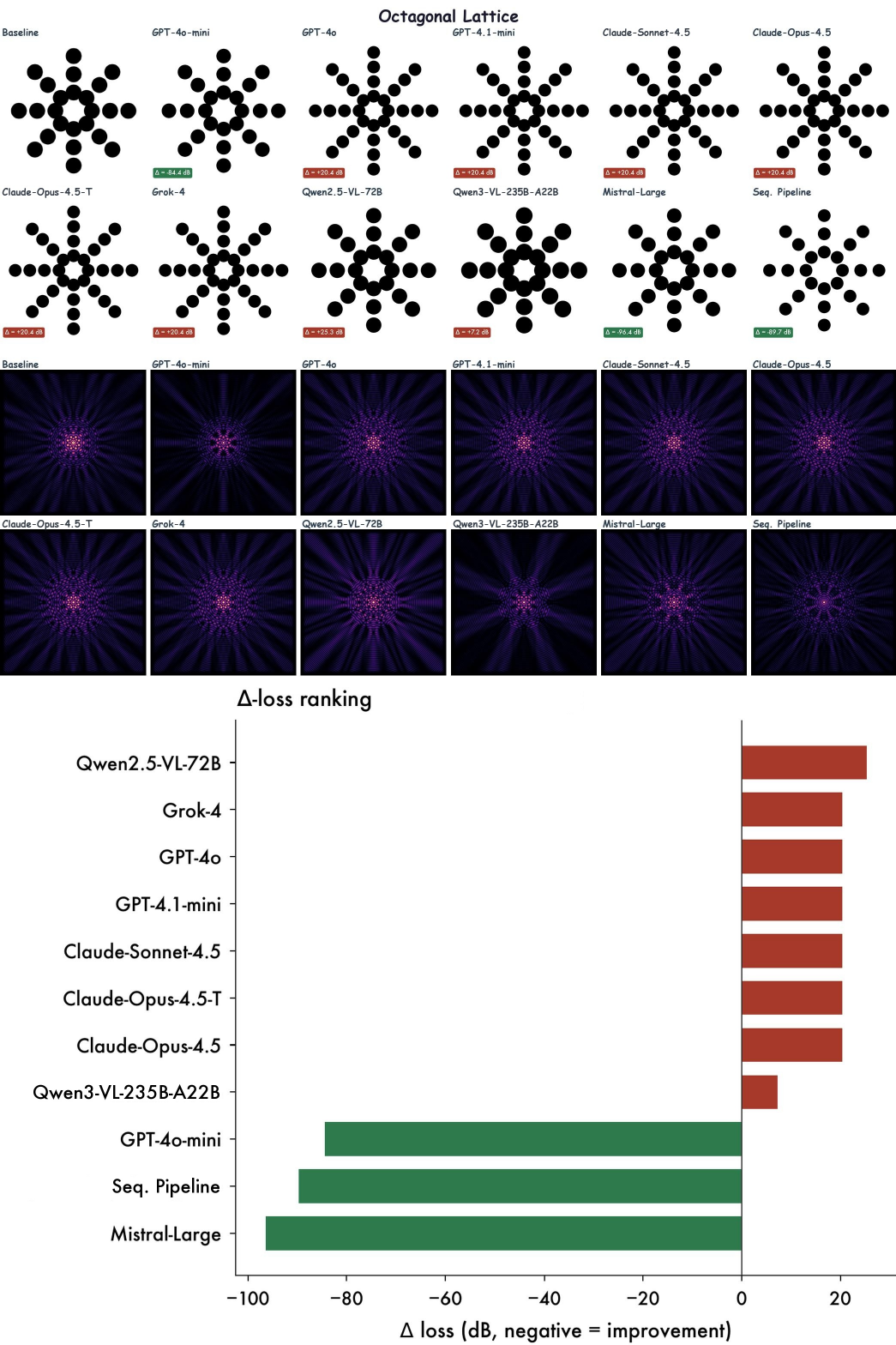}
\caption{Case~18: Octagonal Lattice. Three improvements, the best at $-96.4$\,dB.}\label{fig:case18}
\end{figure}

\begin{figure}[H]
\centering\includegraphics[width=\textwidth]{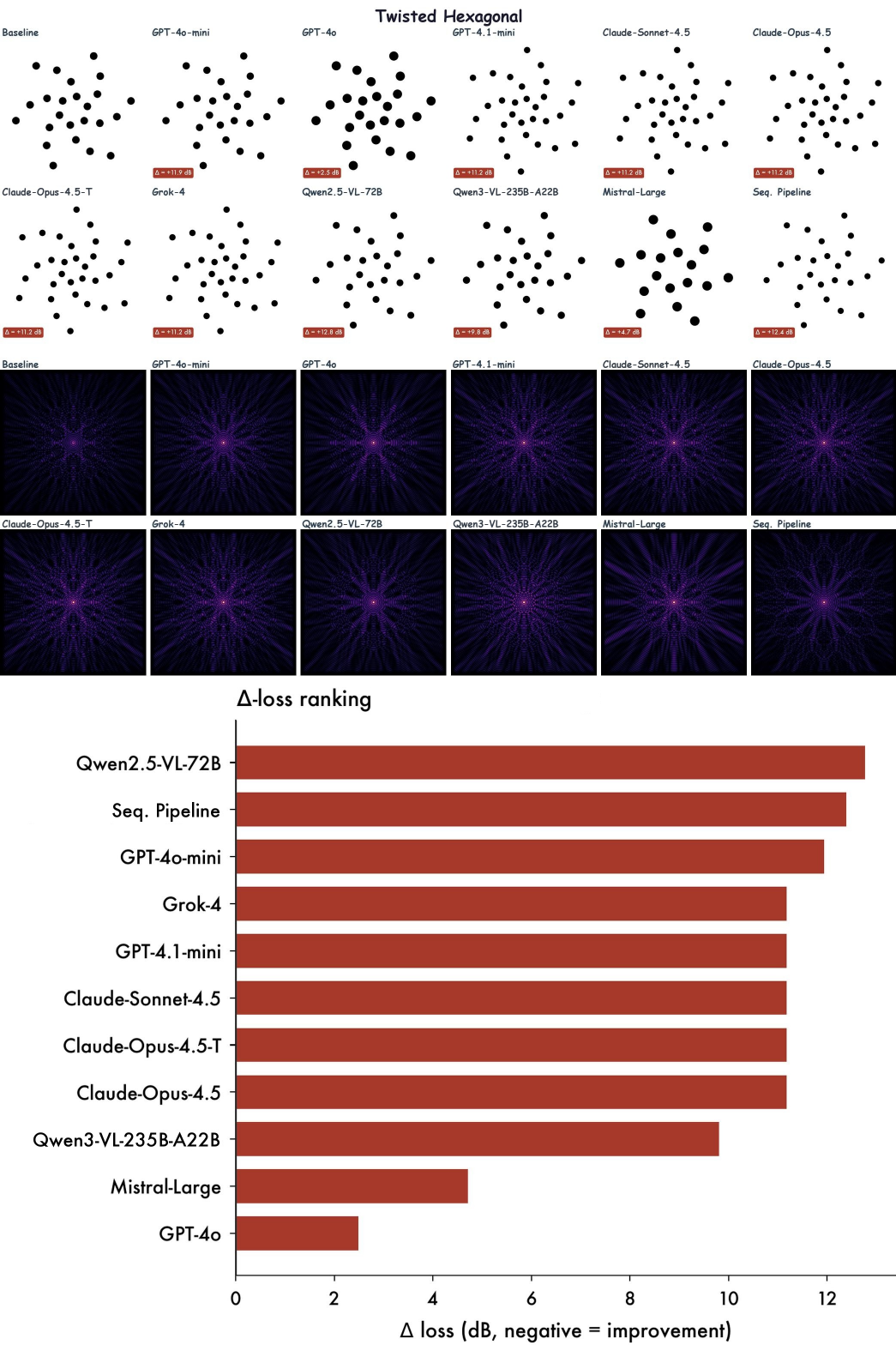}
\caption{Case~19: Twisted Hexagonal. Near-floor baseline, so every system regresses.}\label{fig:case19}
\end{figure}

\begin{figure}[H]
\centering\includegraphics[width=\textwidth]{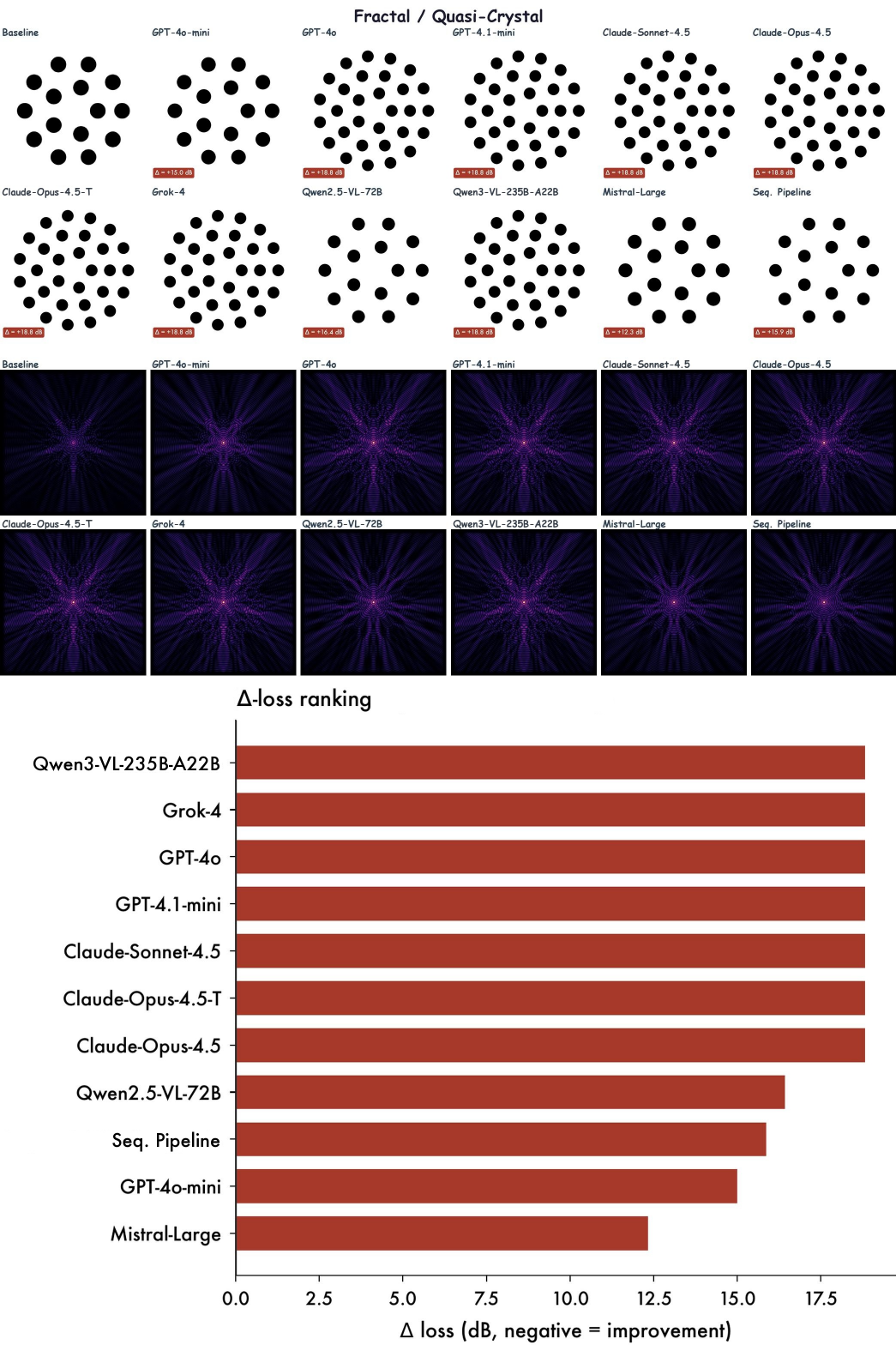}
\caption{Case~20: Fractal / Quasi-Crystal. Without periodicity to exploit, every system regresses.}\label{fig:case20}
\end{figure}

\paragraph{Three Outcome Regimes.}
The 20 cases (Figs.~\ref{fig:case02}--\ref{fig:case20}) separate into three regimes. Models succeed when the correct modification is a monotonic single-parameter adjustment, as in cases~1 (ARF, 9/11 improve, left panel of \textbf{Fig.~\ref{fig:case_study}}), 6 (Kagome, 9/10, \textbf{Fig.~\ref{fig:case06}}), and 13 (Dual-Core Coupler, 10/11, right panel of \textbf{Fig.~\ref{fig:case_study}}). They regress uniformly when the baseline is already near the simulator floor and no improvement is achievable, as in cases~9 (PBG, \textbf{Fig.~\ref{fig:case09}}), 10 (All-Solid Bandgap, \textbf{Fig.~\ref{fig:case10}}), 14 (Vortex Ring-Core, \textbf{Fig.~\ref{fig:case14}}), and 19 (Twisted Hexagonal, \textbf{Fig.~\ref{fig:case19}}), with $\Delta$ between $+2$ and $+15$\,dB. They fail most severely when improvement requires coordinated multi-parameter changes, as in cases~2 (Nested Anti-Resonant, \textbf{Fig.~\ref{fig:case02}}), 3 (Hollow-Core ARF, \textbf{Fig.~\ref{fig:case03}}), 8 (Suspended Cobweb, \textbf{Fig.~\ref{fig:case08}}), and 20 (Fractal Quasi-Crystal, \textbf{Fig.~\ref{fig:case20}}), where every system regresses by $+12$ to $+30$\,dB.

\paragraph{Anti-resonant Family (Cases 1--3).}
Case~1 (ARF, baseline $2.2\!\times\!10^{-2}$\,dB/m) is the prototypical success, with 9/11 systems improving, six by 84 to 97\,dB and three to a numerically zero loss, mostly by retuning the tube wall thickness. Case~2 (Nested Anti-Resonant, $10.3$\,dB/m, \textbf{Fig.~\ref{fig:case02}}) inverts the outcome, and all systems regress ($+23$ to $+30$\,dB) because nested tubes require two gap widths to change together. Case~3 (Hollow-Core ARF, \textbf{Fig.~\ref{fig:case03}}) reproduces this failure in a single-ring topology, implicating multi-resonance-band reasoning rather than geometric complexity.

\paragraph{Plasmonic and Moderate-Loss Variants (Cases 4--5).}
Case~4 (Nested Anti-Resonant, lower loss, \textbf{Fig.~\ref{fig:case04}}) partially recovers (2/11 improve), confirming loss-regime dependence. Case~5 (Hybrid Plasmonic, \textbf{Fig.~\ref{fig:case05}}) shows sporadic success (2/11) with metallic inclusions, suggesting plasmonic resonance tuning is occasionally within reach but unreliable.

\paragraph{Inhibited Coupling, Multi-Core, and Cobweb (Cases 6--8).}
Case~6 (Kagome, $22.0$\,dB/m, \textbf{Fig.~\ref{fig:case06}}) is the strongest success, with 9/10 systems reaching the numerical floor ($\Delta \approx -141$\,dB) through different edits, including thinner struts and an extra cladding ring. Case~7 (Multi-Core, \textbf{Fig.~\ref{fig:case07}}) is markedly harder (1/11 marginal improvement) as inter-core coupling reasoning fails. Case~8 (Suspended Cobweb, \textbf{Fig.~\ref{fig:case08}}) shows the largest baseline loss ($558$\,dB/m) yet no system improves. The dominant error is reducing strut thickness in the wrong loss regime.

\paragraph{Bandgap and Index-Guiding Families (Cases 9--12).}
Cases~9--12 (PBG, All-Solid Bandgap, Solid-Core Hexagonal, and Graded-Index Multi-Ring in Figs.~\ref{fig:case09}--\ref{fig:case12}) form a near-optimal cluster with baselines near the simulator floor ($\sim\!10^{-12}$\,dB/m). Most modifications regress, and 0/11 improve in cases~9, 10, and 12. Case~11 (\textbf{Fig.~\ref{fig:case11}}) has one anomalous improvement where a model lands on an alternative near-floor configuration.

\paragraph{Coupled-Core and Ring-Core (Cases 13--14).}
Case~13 (Dual-Core Coupler) is detailed in the main text. Case~14 (Vortex Ring-Core, \textbf{Fig.~\ref{fig:case14}}) shows 0/11 improvements, as its baseline already sits near the simulator floor.

\paragraph{Birefringent and Elliptical Families (Cases 15--16).}
Case~15 (High-Birefringence PM, \textbf{Fig.~\ref{fig:case15}}) yields occasional success (1/11). Case~16 (Elliptical Core, \textbf{Fig.~\ref{fig:case16}}) is the most evenly contested case (6/11 improve, modest $\Delta$ either way).

\paragraph{Lattice Variants and Quasi-Crystals (Cases 17--20).}
Case~17 (Square/Rectangular, \textbf{Fig.~\ref{fig:case17}}) shows a single improvement and Case~18 (Octagonal, \textbf{Fig.~\ref{fig:case18}}) shows three ($-84$ to $-96$\,dB), with the rest regressing. Case~19 (Twisted Hexagonal, \textbf{Fig.~\ref{fig:case19}}) regresses uniformly (0/11 improve) from a near-floor baseline. Case~20 (Fractal Quasi-Crystal, \textbf{Fig.~\ref{fig:case20}}) sees every system regress ($+12$ to $+19$\,dB), since single-axis design intuition has no lattice periodicity to exploit.

\paragraph{Implications.}
The 20-case sweep shows that aggregate $\Delta$-loss conflates cases with simple single-parameter improvements (cases~1, 6, 13) and cases where failure arises either from geometric coupling (cases~2, 3, 8, 20) or from near-floor baselines (cases~9--12, 14, 17, and 19). Per-family closed-loop success is therefore necessary for meaningful VLM evaluation in photonic inverse design.

\clearpage

\input{sections/app_case_prompts}

\section{Limitations and Broader Impacts}
\label{app:limitations}

\paragraph{Limitations.}
PCFBench is a controlled benchmark for relative, physics-aware evaluation of VLM geometry-to-design reasoning, and it complements downstream full-fidelity design and fabrication verification. The 18 fiber families covered reflect the principal guidance mechanisms in routine practical use. The benchmark covers the transverse structure-to-field problem, which Appendix~\ref{app:propagation} extends to propagation in index-guiding fibers, with hollow-core propagation as the next target.

\paragraph{Broader Impacts.}
PCFBench is a measurement substrate for VLM reasoning in a physics-grounded engineering domain, with primarily indirect broader impacts. Positively, it lowers the barrier to studying physics-aware multimodal reasoning, enables reproducible comparison of design-oriented VLMs, and complements simulation workflows for fiber applications in communications, sensing, and scientific instrumentation. Negatively, users may overinterpret benchmark performance as evidence of fabrication readiness, a risk mitigated by our central finding of a perception--design gap and by our recommendation of simulation-in-the-loop verification before fabrication. PCFBench releases only curated geometric parameters, simulated fields, and expert annotations, so direct misuse risk is low.

%% file: Tables/tab_failure_modes.tex
\begin{table}[!htbp]
\caption{Automated quality-control checks applied to every candidate sample. A sample is rejected when it meets the listed criterion.}
\label{tab:failure_modes}
\centering
\small
\setlength{\tabcolsep}{5pt}
\begin{tabular}{@{}>{\raggedright\arraybackslash}p{\dimexpr(\textwidth-10pt)*28/100\relax} >{\raggedright\arraybackslash}p{\dimexpr(\textwidth-10pt)*72/100\relax}@{}}
\toprule
\textbf{Check} & \textbf{Rejection Criterion} \\
\midrule
Numerical convergence & Confinement loss is undefined or infinite. \\
\addlinespace[3pt]
Mode-pattern validity & The mode-field image shows isotropic speckle without a discernible confined-mode lobe. \\
\addlinespace[3pt]
Geometric validity & Wall thickness $\le 0$, overlapping holes, core radius~$>$~cladding outer radius, or strut count outside the family range. \\
\addlinespace[3pt]
Mode-count consistency & The estimated mode count deviates from the $V$-number prediction by more than one order. \\
\addlinespace[3pt]
Annotation--physics consistency & The L3 diagnosis proposes a parameter change whose direction contradicts the family-specific physics rule (Appendix~\ref{app:physics_gt}) for the L2-reported loss regime. \\
\bottomrule
\end{tabular}
\end{table}

%% file: Tables/tab_bench_comparison.tex
\begin{table}[!htbp]
\caption{PCFBench vs.\ prior multimodal and scientific benchmarks. \textbf{Modality}: I = image, T = text, C = code, D = dialogue. \textbf{Closed-Loop Sim.}: ground truth verified by re-running a numerical solver on the model output. \textbf{Inverse Design}: tasks ask for actionable parameter changes, not only classification or QA. \textbf{Native MAS Eval}: built-in evaluation of multi-agent systems.}
\label{tab:bench_comparison}
\centering
\footnotesize
\renewcommand{\arraystretch}{1.2}
\setlength{\tabcolsep}{3pt}
\begin{tabular}{@{}l >{\raggedright\arraybackslash}p{1.95cm} l c r c c c@{}}
\toprule
\textbf{Benchmark} & \textbf{Domain} & \textbf{Modality} & \textbf{Tasks} &
\textbf{Samples} & \textbf{\shortstack{Closed-Loop\\Sim.}} &
\textbf{\shortstack{Inverse\\Design}} & \textbf{\shortstack{Native\\MAS Eval}} \\
\midrule
MathVista~\cite{mathvista2023}      & Math Reasoning           & I + T       &  6 (subtypes) &  6{,}141 & \cmpN & \cmpN & \cmpN \\
MMStar~\cite{mmstar2024}            & General Multimodal       & I + T       &  6 (subtypes) &  1{,}500 & \cmpN & \cmpN & \cmpN \\
GeoBench~\cite{geobench2025}        & Geospatial Reasoning     & I + T       & 12             &  2{,}000 & \cmpN & \cmpN & \cmpN \\
PICBench~\cite{wu2025picbench}        & Photonic IC Layout       & T + C       &  4             &    420   & \cmpY (partial) & \cmpY & \cmpN \\
ScienceAgentBench~\cite{chen2025scienceagentbench} & Scientific Workflows & T + C + I & \begin{tabular}[t]{@{}c@{}}102\\(workflows)\end{tabular} & 102 & \cmpY (per task) & \cmpN & \cmpY \\
NewtonBench~\cite{zheng2026newtonbench}  & Physical-Law Inference   & T + I       & 12             &  1{,}200 & \cmpN & \cmpN & \cmpN \\
DSBench~\cite{jing2025dsbench}          & Data-Science Automation  & T + C       & 466            &    466   & \cmpY (per task) & \cmpN & \cmpY \\
MedAgentGym~\cite{chen2024gmai}  & Medical Reasoning        & I + T       & 15             &  1{,}800 & \cmpN & \cmpN & \cmpY \\
\midrule
\textbf{PCFBench (ours)} & \textbf{Photonic Crystal Fiber} & \textbf{I + T + C + D} &
\textbf{26}     & \textbf{220{,}000} & \cmpY (FDTD) & \cmpY & \cmpY \\
\bottomrule
\end{tabular}
\end{table}

%% file: Tables/tab_fdtd_config.tex
\begin{table}[!htbp]
\caption{Unified simulation configuration shared by all 220K PCFBench samples. Values are held constant across the 18 fiber families so that loss and mode-field outputs are directly comparable.}
\label{tab:fdtd_config}
\centering
\small
\setlength{\tabcolsep}{6pt}
\renewcommand{\arraystretch}{1.05}
\resizebox{\textwidth}{!}{
\begin{tabular}{l l}
\toprule
\textbf{Component} & \textbf{Setting} \\
\midrule
Engine & Meep FDTD (time-domain, 2D) \\
Spatial resolution & 40 pixels per $\mu$m (uniform Yee grid, $\Delta x \approx 25$\,nm) \\
Cell size & $50 \times 50\,\mu$m square (cell radius $25\,\mu$m) \\
Boundary conditions & Perfectly Matched Layer (PML) on all four lateral edges \\
PML thickness & $2.0\,\mu$m \\
Materials & Fused silica $n_{\mathrm{SiO_2}} = 1.444$ (fixed across both telecom bands), air holes $n = 1.000$ \\
Wavelengths & $\lambda \in \{1310, 1550\}$\,nm, one independent simulation per wavelength \\
Source & Continuous-wave point source ($E_z$ component) at the cell center \\
Run time & 200 Meep time units per wavelength \\
Mode field & $|E_z|^2$ on the full cell at $z = 0$, normalized to its maximum \\
Loss proxy & Total outward flux through four near-PML monitors, normalized by mean intensity, in nominal dB/km \\
$n_\text{eff}$ proxy & Family-dependent constant (e.g., 1.42 for solid-core lattices, 1.01 for hollow-core, 1.35 for plasmonic) \\
Difficulty label & Loss at 1550\,nm $> 10$\,dB/km tagged \emph{high-loss}, otherwise \emph{low-loss} \\
\bottomrule
\end{tabular}
}
\end{table}

%% file: Tables/tab_add_sensitivity.tex
\begin{table}[H]
\caption{Per-cluster numerical-sensitivity check on the simulator. Smaller is better.}
\label{tab:add_sensitivity}
\centering\small
\setlength{\tabcolsep}{8pt}
\begin{tabular}{l c}
\toprule
\textbf{Cluster} & $|\Delta\log\mathcal{L}|$ (dex) \\
\midrule
Index-guided     & 0.022 \\
ARROW            & 0.038 \\
PBG / FSM        & 0.051 \\
Plasmonic        & 0.084 \\
\bottomrule
\end{tabular}
\end{table}

%% file: sections/app_task_prompts.tex
\subsection{Shared Evaluation Protocol}
\label{app:tasks_shared}

Each PCFBench task uses a fixed prompt template and decoding setting, a rule-based answer extractor, and a predefined metric. \textbf{Tab.~\ref{tab:task_overview}} lists the 26 tasks across six categories.

\paragraph{System Prompt.}
All tasks use ``\texttt{You are an expert in photonic crystal fibers.}''

\paragraph{Decoding Configuration.}
All tasks use temperature $T\,{=}\,0.1$ (near-deterministic for reproducibility) with task-specific output-token budgets calibrated to each task's expected output length. \textbf{Tab.~\ref{tab:decoding_config}} lists the full per-task configuration and the images given to each task. Budgets range from $64$ tokens for single-label classifications up to $2048$ for full Meep simulation scripts.

\input{Tables/tab_decoding_config}

\paragraph{Image Notation.}
Prompt templates below use the placeholders \imgtoken{structure}, \imgtoken{field@1310\,nm}, and \imgtoken{field@1550\,nm} to mark image slots. Multi-image prompts concatenate images horizontally with captioned separators. The \textit{Images} line of each prompt box lists the images used by that task.

\paragraph{Evaluation Subset.}
All reported results use a stratified subset of 200 samples per task drawn with a fixed random seed, except Mechanism Diagnosis (Tasks~16/17), which is scored on the 15 samples with reliable mechanism annotations. Spot checks on the full 1000-sample evaluation pool keep the model ranking unchanged and primary-metric scores within 1\%.

\paragraph{Shared Metric Formulas.}
Tasks reuse a small set of metrics, defined once here:
\begin{align}
\mathrm{Accuracy} &= \tfrac{1}{N}\sum_{i=1}^N \mathds{1}[\hat{y}_i = y_i], \\
\mathrm{Macro\text{-}F1} &= \tfrac{1}{|\mathcal{Y}|}\sum_{c \in \mathcal{Y}} F_{1,c}, \\
\mathrm{NRMSE} &= \frac{\sqrt{\tfrac{1}{N}\sum_i (\hat{\theta}_i - \theta_i)^2}}
                       {\theta_{\max} - \theta_{\min}},
\quad
\mathrm{Log\text{-}MAE} = \tfrac{1}{N}\sum_i \big|\log_{10}\hat{\mathcal{L}}_i - \log_{10}\mathcal{L}_i\big|,
\end{align}
plus BERTScore-F1 (RoBERTa-large) and ROUGE-L for generation tasks.

\paragraph{Answer Extraction Rules.}
Three shared extraction rules cover most tasks. (i) \emph{Substring matching} lowercases the response, normalizes hyphens and spaces to underscores, and picks the longest class label that appears as a substring (longest-first prevents, e.g., ``telecom'' from prematurely matching ``ultra\_low\_loss\_telecom''). (ii) \emph{First-number parsing} reads the first integer or decimal number in the response, and loss predictions use the dedicated extraction rule of Tasks~6/7. (iii) \emph{Keyword-coverage scoring} tokenizes the response, intersects it with a task-specific keyword list, and reports $|\mathcal{K}_\cap|/|\mathcal{K}_\mathrm{req}|$.

\subsection{Task Definitions and Prompts}
\label{app:tasks}

The following subsections, one per task category, give the prompt template, extraction rule, and metric of every task.

\subsubsection{Geometry Perception}

The geometry-perception category covers extracting basic structural information from a cross-section image, the foundation of the evaluation hierarchy.

\paragraph{Task~1: Family Classification.}
17-way classification over all families except Kagome, from a $336\!\times\!336$ structure image.

\begin{taskprompt}[Task 1: Family Classification]
\prompthead{Input}\ \imgtoken{structure}

\prompthead{Instruction}\ What is the family type of this photonic crystal fiber structure? Choose from: Solid-Core Hexagonal, Hollow-Core PBG, Anti-Resonant (ARF), High-Birefringence PM, Suspended Cobweb, Multi-Core, Square/Rectangular, Vortex/Ring-Core, Fractal/Quasi-Crystal, Octagonal Lattice, Dual-Core Coupler, Nested Anti-Resonant, Elliptical Core, Hybrid Plasmonic, All-Solid Bandgap, Twisted Hexagonal, Graded-Index Multi-Ring.

\prompthead{Output Format}\ Answer with only the family name (one of the 17 labels above).
\promptimages{Structure only.}
\promptextraction{After lowercasing and underscore normalization, longest-substring match to the 17 family labels.}
\promptmetric{Accuracy and Macro-F1.}
\end{taskprompt}

\paragraph{Task~2 / Task~3: Parameter Prediction (Pitch~$\Lambda$ / Hole~Diameter~$d$).}
Two scalar-regression tasks evaluated separately: $\Lambda \in [0.8, 50]\,\mu$m and $d \in [0.2, 7.5]\,\mu$m. The family name is given in the prompt so the model can condition on canonical geometry.

\begin{taskprompt}[Task 2 / 3: Parameter Prediction (Pitch~$\Lambda$ / Hole Diameter~$d$)]
\prompthead{Input}\ \imgtoken{structure}

\prompthead{Instruction}\ This is a \promptvar{family} photonic crystal fiber. What is the value of \promptvar{parameter} in this PCF structure?

\prompthead{Output Format}\ Provide only the numerical value in micrometers (e.g., \texttt{1.35}).
\promptimages{Structure only.}
\promptextraction{First number in the response, parsed as the scalar prediction.}
\promptmetric{MAE and NRMSE.}
\end{taskprompt}

\paragraph{Task~4: Structure Counting (Cladding Rings).}
Exact integer count of cladding rings, with the family name given in the prompt.

\begin{taskprompt}[Task 4: Structure Counting (Cladding Rings)]
\prompthead{Input}\ \imgtoken{structure}

\prompthead{Instruction}\ This is a \promptvar{family} photonic crystal fiber. How many cladding rings are in this PCF structure?

\prompthead{Output Format}\ Provide only the integer number (e.g., \texttt{4}).
\promptimages{Structure only.}
\promptextraction{First integer in the response, taken as the ring count.}
\promptmetric{Exact-match accuracy (strict) and MAE (partial credit).}
\end{taskprompt}

\paragraph{Task~5: Topology Recognition.}
5-way lattice-symmetry classification.

\begin{taskprompt}[Task 5: Topology Recognition]
\prompthead{Input}\ \imgtoken{structure}

\prompthead{Instruction}\ What is the lattice topology of this PCF structure? Choose from: hexagonal, square, octagonal, kagome, other.

\prompthead{Output Format}\ Answer with only the topology type.
\promptimages{Structure only.}
\promptextraction{Case-insensitive exact match against \{hexagonal, square, octagonal, kagome, other\}.}
\promptmetric{Accuracy and Macro-F1.}
\end{taskprompt}

\subsubsection{Physics Understanding}

This category covers inferring physical properties, mainly from mode-field images, the key capability between perception and design.

\paragraph{Task~6 / Task~7: Loss Prediction (1310\,nm / 1550\,nm).}
Confinement loss spans $10^{-13}$ to $10^3$\,dB/km, so the prompt explicitly permits scientific notation. Both ground-truth and predicted values are clamped to a floor of $10^{-9}$\,dB/km before log-space scoring. Variation below this floor carries no design information and would otherwise let a few sub-floor samples dominate Log-MAE.

\begin{taskprompt}[Task 6 / 7: Loss Prediction (1310\,nm / 1550\,nm)]
\prompthead{Input}\ \imgtoken{field@\promptvar{wavelength}\,nm}

\prompthead{Instruction}\ Estimate the confinement loss of this PCF at \promptvar{wavelength}\,nm from the field distribution pattern.

\prompthead{Output Format}\ Provide a single numerical value in dB/km. Use scientific notation if needed (e.g., \texttt{1.5e-3}).
\promptimages{Mode-field at the queried wavelength only.}
\promptextraction{Four-stage parsing in priority order: (1) a value tagged dB/km or dB/m, with dB/m converted to dB/km, (2) scientific notation not followed by nm, (3) the first bare number that is neither followed by nm nor an echo of the wavelength (1310 or 1550), and (4) zero if nothing matches. Errors are then computed in log space.}
\promptmetric{NRMSE and Log-MAE (loss spans 16 orders of magnitude, so Log-MAE is primary).}
\end{taskprompt}

\paragraph{Task~8: Application Classification.}
13-way classification (12 application categories plus \texttt{unknown}) mapping mode-field features to functional categories.

\begin{taskprompt}[Task 8: Application Classification]
\prompthead{Input}\ \imgtoken{field@1550\,nm}

\prompthead{Instruction}\ Based on the performance characteristics of this PCF structure, what is its most suitable application? Choose from: telecom, sensing, experimental, polarization\_maintaining, high\_power\_delivery, space\_division\_multiplexing, nonlinear\_optics, plasmonic\_sensing, orbital\_angular\_momentum, optical\_switching, robust\_bandgap\_guidance, ultra\_low\_loss\_telecom, unknown.

\prompthead{Output Format}\ Answer with only the category name.
\promptimages{Field 1550\,nm only.}
\promptextraction{Longest-substring match against the category list (longest-first prevents premature matches, e.g., ``telecom'' inside ``ultra\_low\_loss\_telecom'').}
\promptmetric{Accuracy and Macro-F1.}
\end{taskprompt}

\paragraph{Task~9: Mode Classification.}
3-way classification into single-mode, multi-mode, or leaky.

\begin{taskprompt}[Task 9: Mode Classification]
\prompthead{Input}\ \imgtoken{field@1550\,nm}

\prompthead{Instruction}\ What is the guidance mode of this PCF structure? Choose from: single\_mode (supports only the fundamental mode), multi\_mode (supports multiple modes), or leaky (high loss, poor confinement).

\prompthead{Output Format}\ Answer with only the mode type.
\promptimages{Field 1550\,nm only.}
\promptextraction{After lowercasing and normalizing hyphens and spaces to underscores, substring match against \{single\_mode, multi\_mode, leaky\}.}
\promptmetric{Accuracy and Macro-F1.}
\end{taskprompt}

\paragraph{Task~10: Mechanism Reasoning.}
Free-text explanation of the guidance principle in 3--5 sentences.

\begin{taskprompt}[Task 10: Mechanism Reasoning]
\prompthead{Input}\ \imgtoken{structure}

\prompthead{Instruction}\ Explain the physical mechanism of light guidance in this PCF structure in 3--5 sentences. What principle enables light confinement? Mention the specific guidance mechanism (e.g., mTIR, photonic bandgap, anti-resonant reflection).

\prompthead{Output Format}\ 3--5 sentences of free-text physics explanation.
\promptimages{Structure only.}
\promptextraction{Keyword-coverage scoring against a 40-term dictionary (mTIR, PBG, ARF, NANF, confinement, resonance, evanescent, \ldots).}
\promptmetric{Mean keyword recall $|\mathcal{K}_\cap|/|\mathcal{K}_\mathrm{req}|$.}
\end{taskprompt}

\subsubsection{Text Generation}

These tasks bridge visual perception and technical communication. Ground-truth references come from the L1/L2 expert annotations (Sec.~\ref{sec:dataset}).

\paragraph{Task~11: Geometry Description.}
The model describes the fiber geometry in three to five sentences from the structure image.

\begin{taskprompt}[Task 11: Geometry Description]
\prompthead{Input}\ \imgtoken{structure}

\prompthead{Instruction}\ Describe the geometric structure of this photonic crystal fiber in 3--5 sentences. Include: lattice type, core structure, number of rings, and key geometric parameters.

\prompthead{Output Format}\ 3--5 sentences of scientific prose.
\promptimages{Structure only.}
\promptextraction{Raw generation scored directly against the L1 ground-truth description, with no special parsing.}
\promptmetric{BERTScore-F1 (primary), ROUGE-L, domain-term recall.}
\end{taskprompt}

\paragraph{Task~12: Conversational QA.}
The question is taken verbatim from the first Q\&A pair of the sample's L1 annotation, and the model answers with the mode-field image attached.

\begin{taskprompt}[Task 12: Conversational QA]
\prompthead{Input}\ \imgtoken{field@1550\,nm}

\prompthead{Instruction}\ \promptvar{question\_from\_L1\_annotation} (answered in natural language).

\prompthead{Output Format}\ Free-text scientific answer.
\promptimages{Field 1550\,nm only.}
\promptextraction{Raw text scored directly without parsing. Auxiliary domain-term overlap is computed against a curated glossary.}
\promptmetric{BERTScore-F1 (primary), ROUGE-L, domain-term overlap.}
\end{taskprompt}

\paragraph{Task~13: Report Generation.}
Four-section structured technical report.

\begin{taskprompt}[Task 13: Report Generation]
\prompthead{Input}\ \imgtoken{structure}\ \imgtoken{field@1550\,nm}

\prompthead{Instruction}\ Generate a structured analysis report (300--500 words) for this photonic crystal fiber covering four sections:
\begin{itemize}[leftmargin=1.4em, itemsep=0pt, topsep=1pt]
  \item \textit{Geometry}: lattice type, core structure, key parameters.
  \item \textit{Guidance mechanism}: physical principle of light confinement.
  \item \textit{Performance}: loss, mode properties, dispersion characteristics.
  \item \textit{Applications}: suitable use cases based on the design.
\end{itemize}

\prompthead{Output Format}\ A structured report of 300--500 words with clearly labeled sections.
\promptimages{Structure $+$ field 1550\,nm (concatenated).}
\promptextraction{Section-wise keyword recall against four curated term lists (geometry / mechanism / performance / application).}
\promptmetric{$\mathrm{Completeness} = \tfrac{1}{4}\sum_{k=1}^{4} c_k$, where $c_k \in [0,1]$ is the keyword-coverage score for section $k$.}
\end{taskprompt}

\subsubsection{Multi-modal Reasoning}

Each of the three reasoning tasks has two input settings, \textbf{Paired} (structure $+$ field, full prompt) and \textbf{Field-only} (field alone, reduced prompt), which isolates the contribution of the structure image.

\paragraph{Task~14 (Paired) / Task~15 (Field-only): Wavelength Comparison.}
The model decides whether 1310\,nm or 1550\,nm has the higher confinement loss, with or without the structure image.

\begin{taskprompt}[Task 14 / 15: Wavelength Comparison (Paired / Field-only)]
\prompthead{Paired (Task 14) Input}\ \imgtoken{structure}\ \imgtoken{field@1310\,nm}\ \imgtoken{field@1550\,nm}

\prompthead{Paired Instruction}\ Compare the optical field distributions at 1310\,nm and 1550\,nm. Which wavelength exhibits higher confinement loss? Answer with the wavelength number (1310 or 1550) first, then briefly explain why.

\prompthead{Field-only (Task 15) Input}\ \imgtoken{field@1310\,nm}\ \imgtoken{field@1550\,nm}

\prompthead{Field-only Instruction}\ Which wavelength exhibits higher confinement loss: 1310\,nm or 1550\,nm?

\prompthead{Output Format}\ Wavelength number (\texttt{1310} or \texttt{1550}). Paired responses may additionally include a short explanation.
\promptimages{Structure $+$ both fields (Paired), both fields only (Field-only).}
\promptextraction{Pattern matching on phrasings such as ``1310 has higher loss'', and otherwise the first wavelength mentioned.}
\promptmetric{Accuracy and balanced accuracy, $\mathrm{BalAcc} = \tfrac{1}{2}(\mathrm{TP}_{1310}/\mathrm{P}_{1310} + \mathrm{TP}_{1550}/\mathrm{P}_{1550})$.}
\end{taskprompt}

\paragraph{Task~16 (Paired) / Task~17 (Field-only): Mechanism Diagnosis.}
7-way classification of the dominant loss mechanism.

\begin{taskprompt}[Task 16 / 17: Mechanism Diagnosis (Paired / Field-only)]
\prompthead{Paired (Task 16) Input}\ \imgtoken{structure}\ \imgtoken{field@1550\,nm}

\prompthead{Paired Instruction}\ Diagnose the dominant loss mechanism from the structure and field pattern. Choose from: material absorption, macro-bending, resonant coupling, tunneling leakage, surface scattering, mode mismatch, structural irregularity. State the mechanism first, then briefly explain.

\prompthead{Field-only (Task 17) Input}\ \imgtoken{field@1550\,nm}

\prompthead{Field-only Instruction}\ Diagnose the dominant loss mechanism. Choose from the same seven mechanisms. Answer with only the mechanism name.

\prompthead{Output Format}\ Mechanism name. Paired responses may additionally include a short explanation.
\promptimages{Structure $+$ field 1550\,nm (Paired), field 1550\,nm only (Field-only).}
\promptextraction{Whole-word matching of the 7 labels in underscored, hyphenated, and base forms.}
\promptmetric{Macro-F1 (primary), top-1 and top-3 accuracy.}
\end{taskprompt}

\paragraph{Task~18 (Paired) / Task~19 (Field-only): Evidence-Based Explanation.}
Free-text justification anchored on visual features.

\begin{taskprompt}[Task 18 / 19: Evidence-Based Explanation (Paired / Field-only)]
\prompthead{Paired (Task 18) Input}\ \imgtoken{structure}\ \imgtoken{field@1550\,nm}

\prompthead{Paired Instruction}\ The mode is classified as \promptvar{mode\_label}. Provide an evidence-based explanation (3--5 sentences) citing specific visual features in the cladding region.

\prompthead{Field-only (Task 19) Input}\ \imgtoken{field@1550\,nm}

\prompthead{Field-only Instruction}\ Why is this mode considered a Higher-Order Mode (HOM)? Point out the visual evidence in the field distribution in 2--3 sentences.

\prompthead{Output Format}\ Free-text explanation (2--5 sentences).
\promptimages{Structure $+$ field 1550\,nm (Paired), field 1550\,nm only (Field-only).}
\promptextraction{Raw text scored directly, plus an auxiliary count of visual-evidence keywords (cladding, intensity, leakage, lobe, symmetric, penetration, \ldots).}
\promptmetric{BERTScore-F1 (primary), ROUGE-L, evidence-keyword count.}
\end{taskprompt}

\subsubsection{Inverse-Design Decision}

Inverse-design tasks directly probe the \emph{perception--design gap}. Ground truth for the direction and closed-loop tasks is derived from family-aware physics rules (Appendix~\ref{app:physics_gt}).

\paragraph{Task~20: Parameter Selection.}
The model chooses which geometric parameter to optimize first.

\begin{taskprompt}[Task 20: Parameter Selection]
\prompthead{Input}\ \imgtoken{structure}\ \imgtoken{field@1550\,nm}

\prompthead{Instruction}\ To reduce the confinement loss observed here, which geometric parameter should be prioritized for optimization? Choose from: pitch ($\Lambda$), hole diameter ($d$), strut thickness ($t$), core diameter, ring radius, number of rings, or gap width.

\prompthead{Output Format}\ Parameter name first, then a brief explanation of why.
\promptimages{Structure $+$ field 1550\,nm.}
\promptextraction{Longest-substring match to \{pitch, hole diameter, core diameter, strut thickness, ring radius, number of rings, gap width\}, with aliases (e.g., ``$\Lambda$''$\to$pitch, ``wall''$\to$strut).}
\promptmetric{Accuracy and Macro-F1.}
\end{taskprompt}

\paragraph{Task~21 / Task~22: Design Direction (Pitch / Hole~Diameter).}
Binary increase-or-decrease decision, one task per parameter.

\begin{taskprompt}[Task 21 / 22: Design Direction (Pitch / Hole Diameter)]
\prompthead{Input}\ \imgtoken{structure}\ \imgtoken{field@1550\,nm}

\prompthead{Instruction}\ The confinement loss is \promptvar{loss:.2e}\,dB/km. Should the \promptvar{param\_display} be increased or decreased to improve confinement?

\prompthead{Output Format}\ \texttt{increase} or \texttt{decrease} first, then a brief explanation.
\promptimages{Structure $+$ field 1550\,nm.}
\promptextraction{Keyword match for \texttt{increase} versus \texttt{decrease} or \texttt{reduce} (case-insensitive). Ground truth is the family-aware physics direction (Appendix~\ref{app:physics_gt}).}
\promptmetric{Accuracy, plus binary F1, precision, and recall computed for the positive class (\texttt{increase}). The directional decision is one-sided, so the positive-class score is reported instead of a macro average.}
\end{taskprompt}

\paragraph{Task~23: Closed-Loop Improvement.}
The model proposes a new numeric value, scored by whether the change has the physics-consistent sign and a relative step between 2\% and 50\%.

\begin{taskprompt}[Task 23: Closed-Loop Improvement]
\prompthead{Input}\ \imgtoken{structure}\ \imgtoken{field@1550\,nm}

\prompthead{Instruction}\ The current loss is \promptvar{current\_loss:.2e}\,dB/km. Suggest a new value for \promptvar{param\_display} (currently \promptvar{current\_value:.2f}\,$\mu$m) to reduce the loss.

\prompthead{Output Format}\ Provide only the new numeric value in $\mu$m (e.g., \texttt{2.73}).
\promptimages{Structure $+$ field 1550\,nm.}
\promptextraction{First number in the response $\to$ $\hat{\theta}'$. Success requires $0.02 \leq |\hat{\theta}' - \theta|/|\theta| \leq 0.50$ and $\mathrm{sign}(\hat{\theta}' - \theta) = d^{*}$ (physics direction).}
\promptmetric{Closed-loop success rate and $\Delta$-proxy $= \big|\,|\hat{\theta}' - \theta| - 0.10\cdot\theta\,\big|/|\theta|$ (deviation from a 10\% reference step, lower is better).}
\end{taskprompt}

\subsubsection{Code Generation}

These tasks evaluate the translation of visual understanding into executable Meep FDTD code.

\paragraph{Task~24: Simulation Code Generation.}
The model writes a complete Meep script for the given family and parameters.

\begin{taskprompt}[Task 24: Simulation Code Generation]
\prompthead{Input}\ \imgtoken{structure}

\prompthead{Instruction}\ Generate a complete, executable Python script using Meep to simulate this \promptvar{family} PCF structure with parameters \promptvar{params\_dict}. The script must include:
\begin{enumerate}[leftmargin=1.6em, itemsep=0pt, topsep=1pt]
\item All necessary imports (\texttt{meep}, \texttt{numpy}, \texttt{matplotlib} if needed)
\item Material definitions (silica, air, etc.)
\item Geometry setup with proper PCF structure
\item Simulation cell and boundary conditions
\item Light source configuration
\item Simulation run with appropriate runtime
\item Field extraction and result saving
\item Proper error handling
\end{enumerate}

\prompthead{Output Format}\ Provide ONLY the complete Python code in a \texttt{\textasciigrave\textasciigrave\textasciigrave python} code block, ready to execute.
\promptimages{Structure only.}
\promptextraction{Code-block extraction $\to$ Python syntax check $\to$ import and completeness checks.}
\promptmetric{$Q = 0.3\cdot\mathds{1}[\mathrm{syntax}] + 0.3\cdot|\mathcal{I}_\cap|/|\mathcal{I}_\mathrm{req}| + 0.4\cdot|\mathcal{C}_\cap|/|\mathcal{C}_\mathrm{req}|$, where $\mathcal{I}_\mathrm{req} = \{\texttt{meep}, \texttt{numpy}\}$ and $\mathcal{C}_\mathrm{req} = \{\text{geometry, simulation, source, material, output}\}$.}
\end{taskprompt}

\paragraph{Task~25: Parameter Modification.}
The model applies an expert design suggestion to the current parameter dictionary in code.

\begin{taskprompt}[Task 25: Parameter Modification]
\prompthead{Input}\ \imgtoken{structure}

\prompthead{Instruction}\ Based on the expert design suggestion ``\promptvar{suggestion}'' applied to current parameters \promptvar{current\_params}, generate Python code that:
\begin{enumerate}[leftmargin=1.6em, itemsep=0pt, topsep=1pt]
\item Defines the current parameters as a dictionary
\item Implements the suggested modifications
\item Validates the new parameters
\item Prints the changes
\end{enumerate}

\prompthead{Output Format}\ Complete, executable Python code in a \texttt{\textasciigrave\textasciigrave\textasciigrave python} code block.
\promptimages{Structure only.}
\promptextraction{Code-block extraction and syntax check, with completeness defined as the fraction of \{parameter dictionary, assignment, validation, output\} present.}
\promptmetric{Syntax-correctness rate and completeness.}
\end{taskprompt}

\paragraph{Task~26: Code Refactoring.}
The model refactors a reference Meep script for error handling, readability, documentation, and performance.

\begin{taskprompt}[Task 26: Code Refactoring]
\prompthead{Input}\ \imgtoken{structure}

\prompthead{Instruction}\ Refactor and optimize this PCF simulation code: \promptvar{reference\_code}

Improvements to make:
\begin{enumerate}[leftmargin=1.6em, itemsep=0pt, topsep=1pt]
\item Add proper error handling
\item Improve code structure and readability
\item Add comments and documentation
\item Optimize performance where possible
\item Follow Python best practices
\end{enumerate}

\prompthead{Output Format}\ Provide the refactored code in a \texttt{\textasciigrave\textasciigrave\textasciigrave python} code block.
\promptimages{Structure only.}
\promptextraction{Code-block extraction, then structural comparison of the generated script with the reference.}
\promptmetric{Syntax-correctness rate and $Q_{\mathrm{refactor}} = \tfrac{1}{5}\big(\mathds{1}[\mathrm{syntax}] + \mathds{1}[\mathrm{comments}] + \mathds{1}[\mathrm{docstrings}] + \mathds{1}[\mathrm{error\text{-}handling}] + \mathds{1}[\Delta\mathrm{functions}\geq 0]\big)$, the mean of five binary checks (valid syntax, comments, docstrings, exception handling, and a function count no lower than the reference). Unlike $Q$ of Task~24, it scores style and structure.}
\end{taskprompt}

%% file: Tables/tab_decoding_config.tex
\begin{table}[tbh]
\caption{Per-task decoding configuration. All tasks share temperature $T\,{=}\,0.1$ and the system prompt ``\texttt{You are an expert in photonic crystal fibers.}'', with a task-specific output-token budget. Multi-image rows list all images given to the model. Field-only rows are ablations of the corresponding Paired task.}
\label{tab:decoding_config}
\centering
\resizebox{\textwidth}{!}{%
\begin{tabular}{llcll}
\toprule
\textbf{Category} & \textbf{Task} & \textbf{Max Tokens} & \textbf{Image Input} & \textbf{Output Form} \\
\midrule
\multirow{5}{*}{Geometry Perception}
  & Family Classification           &  64  & \imgtoken{structure}              & 1 family label \\
  & Parameter Pred.\ (Pitch)        & 128  & \imgtoken{structure}              & 1 numeric ($\mu$m) \\
  & Parameter Pred.\ (Hole D)       & 128  & \imgtoken{structure}              & 1 numeric ($\mu$m) \\
  & Structure Counting (Rings)      & 128  & \imgtoken{structure}              & 1 integer \\
  & Topology Recognition            &  64  & \imgtoken{structure}              & 1 topology label \\
\midrule
\multirow{5}{*}{Physics Understanding}
  & Loss Prediction (1310\,nm)      & 128  & \imgtoken{field@1310\,nm}         & 1 numeric (dB/km) \\
  & Loss Prediction (1550\,nm)      & 128  & \imgtoken{field@1550\,nm}         & 1 numeric (dB/km) \\
  & Application Classification      &  64  & \imgtoken{field@1550\,nm}         & 1 category label \\
  & Mode Classification             &  64  & \imgtoken{field@1550\,nm}         & 1 mode label \\
  & Mechanism Reasoning             & 512  & \imgtoken{structure}              & free text \\
\midrule
\multirow{3}{*}{Text Generation}
  & Geometry Description            & 512  & \imgtoken{structure}              & 3--5 sentences \\
  & Conversational QA               & 256  & \imgtoken{field@1550\,nm}         & 1 short answer \\
  & Report Generation               & 512  & \imgtoken{structure}\,+\,\imgtoken{field@1550} & structured report \\
\midrule
\multirow{6}{*}{Multi-modal Reasoning}
  & Wavelength Comparison (Paired)  & 256  & \imgtoken{structure}\,+\,\imgtoken{field@1310}\,+\,\imgtoken{field@1550}  & label\,+\,reasoning \\
  & Wavelength Comparison (Field-only) &  64  & \imgtoken{field@1310}\,+\,\imgtoken{field@1550}  & label only \\
  & Mechanism Diagnosis (Paired)          & 512  & \imgtoken{structure}\,+\,\imgtoken{field@1550}  & label\,+\,evidence \\
  & Mechanism Diagnosis (Field-only)          & 128  & \imgtoken{field@1550\,nm}         & label only \\
  & Evidence Explanation (Paired)         & 512  & \imgtoken{structure}\,+\,\imgtoken{field@1550}  & free-text rationale \\
  & Evidence Explanation (Field-only)         & 256  & \imgtoken{field@1550\,nm}         & shorter rationale \\
\midrule
\multirow{4}{*}{Inverse-Design Decision}
  & Parameter Selection             & 256  & \imgtoken{structure}\,+\,\imgtoken{field@1550}  & 1 parameter + reasoning \\
  & Design Direction (Pitch)        & 256  & \imgtoken{structure}\,+\,\imgtoken{field@1550}  & up/down + reasoning \\
  & Design Direction (Hole D)       & 256  & \imgtoken{structure}\,+\,\imgtoken{field@1550}  & up/down + reasoning \\
  & Closed-Loop Improvement (Pitch) & 128  & \imgtoken{structure}\,+\,\imgtoken{field@1550}  & 1 numeric ($\mu$m) \\
\midrule
\multirow{3}{*}{Code Generation}
  & Simulation Code Generation      & 2048 & \imgtoken{structure}              & full Meep script \\
  & Parameter Modification          & 1024 & \imgtoken{structure}              & modified script \\
  & Code Refactoring                & 2048 & \imgtoken{structure}              & refactored script \\
\midrule
\multirow{4}{*}{\shortstack[l]{MAS Sequential Pipeline\\(per-stage)}}
  & Stage 1 (Perception)            & 256  & \imgtoken{structure}              & family\,+\,rings \\
  & Stage 2 (Understanding)         & 512  & \imgtoken{field@1550\,nm}         & loss\,+\,mode \\
  & Stage 3 (Reasoning)             & 512  & \imgtoken{structure}\,+\,\imgtoken{field@1550}  & wavelength judgment \\
  & Stage 4 (Design)                & 256  & \imgtoken{structure}\,+\,\imgtoken{field@1550}  & parameter\,+\,direction \\
\bottomrule
\end{tabular}}
\end{table}

%% file: Tables/tab_model_catalogue.tex
\begin{table}[!t]
\caption{Evaluated model catalog. Icons match the leaderboard tables. \textit{Access}: Prop.\ = proprietary API, Open = open-weight download. ``-T'' marks a thinking variant.}
\label{tab:model_catalogue}
\centering
\small
\setlength{\tabcolsep}{3pt}
\renewcommand{\arraystretch}{1.15}
\resizebox{\textwidth}{!}{%
\begin{tabular}{l c c l}
\toprule
\textbf{Model} & \textbf{Size} & \textbf{Access} & \textbf{Description} \\
\midrule
\model{claude}{Claude-3.5-Sonnet}{anthropic2024claude35}   & --- & Prop. & Mid-2024 multimodal anchor predating the 4.5 generation \\
\model{claude}{Claude-Sonnet-4.5}{anthropic2025claude45}   & --- & Prop. & Late-2025 Sonnet with balanced multimodal reasoning \\
\model{claude}{Claude-Opus-4.5}{anthropic2025opus45}       & --- & Prop. & 4.5-generation flagship and Anthropic's largest public VLM \\
\model{claude}{Claude-Sonnet-4.5-T}{anthropic2025claude45} & --- & Prop. & Sonnet-4.5 with extended-thinking mode \\
\model{claude}{Claude-Opus-4.5-T}{anthropic2025opus45}     & --- & Prop. & Opus-4.5 with extended-thinking mode \\
\cmidrule(lr){1-4}
\model{openai}{GPT-4o}{hurst2024gpt}         & --- & Prop. & 2024 native-multimodal flagship (vision, speech, and text) \\
\model{openai}{GPT-4o-mini}{hurst2024gpt}     & --- & Prop. & Cost-optimized GPT-4o sibling \\
\model{openai}{GPT-4.1}{openai2025gpt41}         & --- & Prop. & 2025 successor with expanded context and stronger coding performance \\
\model{openai}{GPT-4.1-mini}{openai2025gpt41}     & --- & Prop. & Cost-optimized GPT-4.1 sibling \\
\model{openai}{GPT-5}{singh2025openai}           & --- & Prop. & 2025 unified frontier release (merging the GPT and reasoning lines) \\
\model{openai}{GPT-5-mini}{singh2025openai}       & --- & Prop. & Cost-optimized GPT-5 sibling \\
\model{openai}{o3}{openai2025o3}                 & --- & Prop. & Reasoning-specialized, extended test-time deliberation \\
\model{openai}{o4-mini}{openai2025o3}             & --- & Prop. & Cost-optimized o-series reasoning model \\
\cmidrule(lr){1-4}
\model{gemini}{Gemini-2.5-Flash-Lite}{comanici2025gemini} & --- & Prop. & Cost-optimized 2.5-generation Gemini \\
\model{gemini}{Gemini-3-Flash}{gemini3-2025}          & --- & Prop. & Late-2025 next-generation multimodal, fastest Gemini~3 tier \\
\cmidrule(lr){1-4}
\model{grok}{Grok-4}{grok4-2025}     & --- & Prop. & 2025 frontier multimodal with agentic tool-use objectives \\
\model{grok}{Grok-4.1}{grok4-2025}   & --- & Prop. & Iterative update of Grok-4 \\
\cmidrule(lr){1-4}
\model{zhipu}{GLM-4.6V}{hong2025glm}   & --- & Prop. & Zhipu's bilingual (CN--EN) multimodal GLM variant \\
\cmidrule(lr){1-4}
\model{gemma}{Gemma-3-4B}{team2025gemma}   & 4B  & Open & 2025 open-weight, shared vision encoder and long context \\
\model{gemma}{Gemma-3-12B}{team2025gemma}  & 12B & Open & Mid-scale Gemma-3 \\
\model{gemma}{Gemma-3-27B}{team2025gemma}  & 27B & Open & Largest Gemma-3 \\
\cmidrule(lr){1-4}
\model{meta}{Llama-3.2-11B-V}{grattafiori2024llama}  & 11B  & Open  & 2024 vision-augmented dense Llama-3 \\
\model{meta}{Llama-3.2-90B-V}{grattafiori2024llama}  & 90B  & Open  & Large-scale dense vision Llama-3 \\
\model{meta}{Llama-4-Scout}{llama4-2025}    & 109B & Open  & 2025 natively multimodal MoE \\
\model{meta}{Llama-4-Maverick}{llama4-2025} & 400B & Open  & Largest Llama-4 MoE \\
\cmidrule(lr){1-4}
\model{qwen}{Qwen2-VL-72B-Instruct}{wang2024qwen2}            & 72B  & Open  & 2024 dense VLM with dynamic-resolution input \\
\model{qwen}{Qwen2.5-VL-7B-Instruct}{bai2025qwen25vltechnicalreport}          & 7B   & Open  & 2025 2.5-gen dense VLM, 7B tier \\
\model{qwen}{Qwen2.5-VL-32B-Instruct}{bai2025qwen25vltechnicalreport}         & 32B  & Open  & 2025 2.5-gen dense VLM, mid-tier \\
\model{qwen}{Qwen2.5-VL-72B-Instruct}{bai2025qwen25vltechnicalreport}         & 72B  & Open  & 2025 2.5-gen dense VLM, largest tier \\
\model{qwen}{Qwen3-VL-8B-Instruct}{bai2025qwen3vltechnicalreport}               & 8B   & Open  & 2025 3-gen dense VLM \\
\model{qwen}{Qwen3-VL-30B-A3B-Instruct}{bai2025qwen3vltechnicalreport}          & 30B  & Open  & 3-gen MoE (30B total, 3B active) \\
\model{qwen}{Qwen3-VL-235B-A22B-Instruct}{bai2025qwen3vltechnicalreport}        & 235B & Open  & 3-gen MoE (235B total, 22B active) \\
\model{qwen}{Qwen3.5-9B}{qwenteam2026qwen35omnitechnicalreport}                        & 9B   & Open  & 2026 3.5-gen dense VLM (small tier, multimodal hybrid reasoning) \\
\model{qwen}{Qwen3.5-27B}{qwenteam2026qwen35omnitechnicalreport}                       & 27B  & Open  & 2026 3.5-gen dense VLM (mid tier) \\
\model{qwen}{Qwen3.5-122B-A10B}{qwenteam2026qwen35omnitechnicalreport}                 & 122B & Open  & 2026 3.5-gen MoE (122B total, 10B active, hybrid attention and sparse MoE) \\
\model{qwen}{Qwen-VL-Max}{bai2025qwen25vltechnicalreport}                     & ---  & Prop. & Alibaba API-only flagship, size undisclosed \\
\cmidrule(lr){1-4}
\model{mistral}{Ministral-3-8B-Instruct}{ministral-2024}    & 8B   & Open  & Edge-scale, latency-optimized dense VLM \\
\model{mistral}{Ministral-3-14B-Instruct}{ministral-2024}   & 14B  & Open  & Mid-scale Ministral VLM \\
\model{mistral}{Pixtral-12B}{agrawal2024pixtral12b}                  & 12B  & Open  & First natively multimodal Mistral release \\
\model{mistral}{Mistral-Large-3-Instruct}{mistrallarge-2024} & 675B & Open  & 2025 flagship multimodal release \\
\cmidrule(lr){1-4}
\model{ai2}{Molmo-2-8B}{deitke2025molmo}             & 8B   & Open  & Fully open (weights, data, and recipe) multimodal model \\
\bottomrule
\end{tabular}}%
\end{table}

%% file: Tables/tab_human_eval_rubric.tex
\begin{table}[tbh]
\caption{Human evaluation rubric for the Inverse-Design Decision category. Each of six dimensions is scored on $\{0.0,\,0.5,\,1.0\}$ by two photonics experts. The composite Human Score is their arithmetic mean across dimensions and raters.}
\label{tab:human_eval_rubric}
\centering
\small
\setlength{\tabcolsep}{5pt}
\begin{tabular}{@{}>{\raggedright\arraybackslash}p{\dimexpr(\textwidth-30pt)*37/100\relax} *{3}{>{\raggedright\arraybackslash}p{\dimexpr(\textwidth-30pt)*21/100\relax}}@{}}
\toprule
\textbf{Dimension} & \textbf{Score 0.0} & \textbf{Score 0.5} & \textbf{Score 1.0} \\
\midrule
\textbf{Physics Validity}\newline Does the proposed parameter change align with waveguide physics (anti-resonance theory, photonic bandgap, or index-guiding principles)? & Contradicts established physics. & Direction correct but reasoning flawed. & Direction correct and reasoning physically sound. \\
\addlinespace[5pt]
\textbf{Quantitative Correctness}\newline Is the proposed value within a physically plausible range for the identified family (e.g., ARF strut~$\in$~$0.3$--$1.2\,\mu$m)? & Outside known bounds. & Within bounds but misaligned with the loss regime. & Appropriate for both family and loss regime. \\
\addlinespace[5pt]
\textbf{Fabrication Feasibility}\newline Can the proposed structure be fabricated with current stack-and-draw or CVD-based processes? & Violates minimum-feature or geometric continuity constraints. & Requires a non-standard process. & Standard fabrication route. \\
\addlinespace[5pt]
\textbf{Performance Improvement}\newline Would the proposed change plausibly reduce confinement loss as claimed? & No improvement or makes loss worse. & Marginal improvement. & Substantial improvement ($>$1 order of magnitude for high-loss, $>$3$\times$ for moderate-loss). \\
\addlinespace[5pt]
\textbf{Design Coherence}\newline Are the parameter choice, direction, and magnitude internally consistent with the diagnosed loss mechanism? & Contradictory across stages. & Partial consistency. & Fully coherent end-to-end reasoning. \\
\addlinespace[5pt]
\textbf{Justification Quality}\newline Does the textual explanation reference the relevant physical principle (Maxwell's equations, resonance condition, bandgap shift)? & No or incorrect justification. & Partially correct reasoning. & Explicit and precise physical reasoning. \\
\bottomrule
\end{tabular}
\end{table}

%% file: Tables/tab_add_kappa.tex
\begin{table}[!htbp]
\caption{Inter-rater reliability of the Human Score on the 120-sample subset.}
\label{tab:add_kappa}
\centering\small
\setlength{\tabcolsep}{8pt}
\begin{tabular}{l c}
\toprule
\textbf{Statistic} & \textbf{Value} \\
\midrule
Cohen's $\kappa$                          & 0.72 \\
Weighted $\kappa$                          & 0.68 \\
Exact agreement                            & 81.4\% \\
Within-one-point agreement                 & 96.2\% \\
Adjudicated cases ($>$1-point disagreement) & 12.5\% \\
Leaderboard Kendall $\tau$ (pre vs.\ post adj.) & 0.97 \\
\bottomrule
\end{tabular}
\end{table}

%% file: Tables/tab_geo_phys.tex
\begin{table}[t]
\caption{\textbf{Geometry Perception and Physics Understanding.} \textit{Fam.}: family classification, \textit{Topo.}: topology recognition, \textit{App.} and \textit{Mode}: application and mode classification, \textit{Mech.}: mechanism reasoning. Top-3 per column are \rankkey{rankgold}{\textbf{Gold}} / \rankkey{ranksilver}{\underline{Blue}} / \rankkey{rankbronze}{Gray}.}
\label{tab:geo_phys}
\centering
\small
\setlength{\tabcolsep}{3.5pt}
\resizebox{\textwidth}{!}{%
\begin{tabular}{l ccccc ccccc}
\toprule
& \multicolumn{5}{c}{\textbf{Geometry Perception}} & \multicolumn{5}{c}{\textbf{Physics Understanding}} \\
\cmidrule(lr){2-6}\cmidrule(lr){7-11}
\multirow{2}{*}{\textbf{Model}} & Fam. & Pitch & Hole\,D & Count & Topo. & Loss@1310 & Loss@1550 & App. & Mode & Mech. \\
& Acc\,$\uparrow$ & NRMSE\,$\downarrow$ & NRMSE\,$\downarrow$ & Acc\,$\uparrow$ & Acc\,$\uparrow$ & Log-MAE\,$\downarrow$ & Log-MAE\,$\downarrow$ & Acc\,$\uparrow$ & Acc\,$\uparrow$ & KwdRec\,$\uparrow$ \\
\midrule
\model{gemma}{Gemma-3-4B}{team2025gemma} & .020 & 2.950 & 2.850 & .145 & .120 & 2.750 & 2.850 & .035 & .120 & .245 \\
\model{qwen}{Qwen2.5-VL-7B}{bai2025qwen25vltechnicalreport} & .096 & 1.820 & 1.395 & .176 & .230 & 2.315 & 2.655 & .055 & .285 & .295 \\
\model{qwen}{Qwen3-VL-8B}{bai2025qwen3vltechnicalreport} & .192 & .276 & .826 & .174 & .293 & 2.290 & 2.435 & .050 & .265 & .265 \\
\model{mistral}{Ministral-8B}{ministral-2024} & .054 & .158 & .209 & .200 & .185 & 2.595 & 2.745 & .040 & .155 & .225 \\
\model{ai2}{Molmo-2-8B}{deitke2025molmo} & .093 & .310 & .536 & .194 & .246 & 2.795 & 2.895 & .030 & .135 & .215 \\
\model{meta}{Llama-3.2-11B-V}{grattafiori2024llama} & .034 & .332 & .209 & .248 & .020 & .600 & 2.491 & \first{.245} & .005 & .300 \\
\cmidrule(lr){1-11}
\model{gemma}{Gemma-3-12B}{team2025gemma} & .075 & 2.185 & 1.945 & .115 & .165 & 2.245 & 2.350 & .050 & .185 & .295 \\
\model{mistral}{Pixtral-12B}{agrawal2024pixtral12b} & .133 & .172 & .252 & .143 & .339 & 2.385 & 2.485 & .055 & .205 & .305 \\
\model{mistral}{Ministral-14B}{ministral-2024} & .115 & .179 & .244 & .199 & .287 & 2.215 & 2.315 & .055 & .245 & .285 \\
\model{gemma}{Gemma-3-27B}{team2025gemma} & .145 & 1.895 & 1.465 & .195 & .245 & 1.785 & 1.885 & .065 & .275 & .335 \\
\model{qwen}{Qwen3-VL-30B-A3B}{bai2025qwen3vltechnicalreport} & .115 & .288 & .705 & .204 & .291 & 1.685 & 1.745 & .080 & .345 & .375 \\
\model{qwen}{Qwen2.5-VL-32B}{bai2025qwen25vltechnicalreport} & .180 & .285 & .324 & .113 & .260 & .625 & 2.671 & .065 & \second{.965} & .352 \\
\cmidrule(lr){1-11}
\model{qwen}{Qwen2-VL-72B}{wang2024qwen2} & .155 & 2.115 & 1.545 & .195 & .385 & 2.485 & 2.590 & .105 & .325 & .285 \\
\model{qwen}{Qwen2.5-VL-72B}{bai2025qwen25vltechnicalreport} & .185 & .270 & .194 & .234 & .430 & .568 & 2.448 & .135 & .085 & .304 \\
\model{meta}{Llama-3.2-90B-V}{grattafiori2024llama} & .025 & 2.565 & 2.215 & .275 & .065 & 2.015 & 2.095 & \second{.185} & .075 & .325 \\
\model{meta}{Llama-4-Scout}{llama4-2025} & .105 & .292 & 22.545 & .048 & .100 & .516 & 2.350 & .045 & .520 & .348 \\
\model{mistral}{Mistral-Large}{mistrallarge-2024} & .060 & .137 & \second{.170} & .226 & .198 & 1.825 & 1.895 & .050 & .345 & .325 \\
\model{qwen}{Qwen3-VL-235B-A22B}{bai2025qwen3vltechnicalreport} & .225 & .295 & .265 & .230 & \second{.482} & 1.165 & 1.225 & .090 & .375 & .405 \\
\model{qwen}{Qwen3.5-9B}{qwenteam2026qwen35omnitechnicalreport} & .175 & .270 & .625 & .195 & .315 & 2.045 & 2.180 & .075 & .305 & .310 \\
\model{qwen}{Qwen3.5-27B}{qwenteam2026qwen35omnitechnicalreport} & .205 & .282 & .488 & .218 & .355 & 1.495 & 1.580 & .115 & .485 & .395 \\
\model{qwen}{Qwen3.5-122B-A10B}{qwenteam2026qwen35omnitechnicalreport} & .242 & .265 & .225 & .265 & \first{.495} & 1.025 & 1.085 & \third{.140} & .495 & .425 \\
\model{meta}{Llama-4-Maverick}{llama4-2025} & .190 & .292 & .221 & .128 & .370 & .617 & 2.576 & .065 & .550 & .234 \\
\model{qwen}{Qwen-VL-Max}{bai2025qwen25vltechnicalreport} & .210 & .265 & .285 & .220 & .372 & 1.345 & 1.420 & .085 & .395 & .385 \\
\cmidrule(lr){1-11}
\model{openai}{GPT-4o-mini}{hurst2024gpt} & .130 & .142 & .230 & .228 & .348 & .845 & .895 & .060 & .465 & .395 \\
\model{openai}{GPT-4o}{hurst2024gpt} & .185 & .183 & .337 & .205 & .461 & .355 & 1.339 & .083 & .835 & .310 \\
\model{openai}{GPT-4.1-mini}{openai2025gpt41} & .305 & .165 & .220 & .397 & .370 & .204 & .676 & .055 & .690 & .376 \\
\model{openai}{GPT-4.1}{openai2025gpt41} & .360 & .145 & .235 & .440 & .420 & .475 & .520 & .075 & .545 & .410 \\
\model{openai}{GPT-5-mini}{singh2025openai} & .385 & .135 & .215 & .405 & .425 & .325 & .355 & .090 & .565 & .445 \\
\model{openai}{GPT-5}{singh2025openai} & \second{.455} & \third{.125} & .195 & \second{.480} & .470 & .195 & .225 & .110 & .605 & \third{.500} \\
\model{openai}{o4-mini}{openai2025o3} & .335 & .155 & .235 & .390 & .415 & .355 & .385 & .085 & .515 & .420 \\
\model{openai}{o3}{openai2025o3} & .420 & \second{.115} & \first{.165} & .450 & \third{.475} & \first{.145} & \first{.170} & .115 & .615 & \second{.505} \\
\cmidrule(lr){1-11}
\model{claude}{Claude-3.5-Sonnet}{anthropic2024claude35} & .065 & .209 & .537 & .199 & .210 & .282 & .832 & .045 & .130 & .377 \\
\model{claude}{Claude-Sonnet-4.5}{anthropic2025claude45} & .265 & .215 & .305 & .295 & .380 & .445 & .485 & .065 & .485 & .445 \\
\model{claude}{Claude-Opus-4.5}{anthropic2025opus45} & .395 & .135 & .215 & .415 & .455 & .245 & .275 & .095 & .555 & .485 \\
\model{claude}{Claude-Sonnet-4.5-T}{anthropic2025claude45} & .305 & .195 & .285 & .335 & .410 & .365 & .395 & .070 & .505 & .470 \\
\model{claude}{Claude-Opus-4.5-T}{anthropic2025opus45} & \third{.445} & \first{.105} & \third{.175} & \third{.455} & \third{.475} & .185 & .205 & .105 & .585 & \third{.500} \\
\cmidrule(lr){1-11}
\model{gemini}{Gemini-3-Flash}{gemini3-2025} & \first{.668} & \first{.105} & \first{.165} & \first{.757} & .425 & \second{.155} & \second{.175} & .110 & .585 & \second{.505} \\
\model{gemini}{Gemini-2.5-Flash-Lite}{comanici2025gemini} & .264 & .243 & .301 & .176 & .434 & \second{.155} & \third{.202} & .065 & .813 & .194 \\
\model{grok}{Grok-4}{grok4-2025} & .215 & .185 & .201 & .220 & .405 & \third{.170} & .203 & .025 & \first{.990} & \first{.510} \\
\model{grok}{Grok-4.1}{grok4-2025} & .240 & .211 & .211 & .184 & .385 & \third{.170} & .203 & .020 & \third{.920} & .452 \\
\model{zhipu}{GLM-4.6V}{hong2025glm} & .145 & .195 & .345 & .210 & .325 & 1.845 & 1.925 & .085 & .345 & .295 \\
\bottomrule
\end{tabular}}
\end{table}

%% file: Tables/tab_text_mm.tex
\begin{table}[t]
\caption{\textbf{Text Generation and Multi-modal Reasoning.} Text metrics are BERTScore-F1 (BS-F1), ROUGE-L (RL), term recall (TR), and completeness (Comp.). WC, MD, and EE denote wavelength comparison, mechanism diagnosis, and evidence explanation under \textit{Paired} (structure and field images) or \textit{Field}-only inputs. Top-3 per column are \rankkey{rankgold}{\textbf{Gold}} / \rankkey{ranksilver}{\underline{Blue}} / \rankkey{rankbronze}{Gray}.}
\label{tab:text_mm}
\centering
\small
\setlength{\tabcolsep}{3.2pt}
\resizebox{\textwidth}{!}{%
\begin{tabular}{l ccc cc c cc cc cc}
\toprule
& \multicolumn{6}{c}{\textbf{Text Generation}} & \multicolumn{6}{c}{\textbf{Multi-modal Reasoning}} \\
\cmidrule(lr){2-7}\cmidrule(lr){8-13}
& \multicolumn{3}{c}{Geo.~Desc.} & \multicolumn{2}{c}{Conv.~QA} & Report & \multicolumn{2}{c}{WC (Acc)} & \multicolumn{2}{c}{MD (F1)} & \multicolumn{2}{c}{EE (BS-F1)} \\
\cmidrule(lr){2-4}\cmidrule(lr){5-6}\cmidrule(lr){7-7}\cmidrule(lr){8-9}\cmidrule(lr){10-11}\cmidrule(lr){12-13}
\textbf{Model} & BS-F1 & RL & TR & BS-F1 & RL & Comp. & Paired & Field & Paired & Field & Paired & Field \\
\midrule
\model{gemma}{Gemma-3-4B}{team2025gemma} & .270 & .109 & .438 & .106 & .073 & .465 & .247 & \second{.759} & .063 & \first{1.000} & .387 & \first{.395} \\
\model{qwen}{Qwen2.5-VL-7B}{bai2025qwen25vltechnicalreport} & .334 & .153 & .404 & .112 & .056 & .482 & \second{.758} & \second{.759} & \first{.696} & \first{1.000} & .340 & .297 \\
\model{qwen}{Qwen3-VL-8B}{bai2025qwen3vltechnicalreport} & .317 & .146 & .566 & .096 & .068 & .554 & .663 & .738 & \second{.500} & .421 & .387 & .307 \\
\model{mistral}{Ministral-8B}{ministral-2024} & .285 & .121 & .587 & .098 & .062 & .625 & .457 & .279 & .125 & .571 & .410 & .348 \\
\model{meta}{Llama-3.2-11B-V}{grattafiori2024llama} & .339 & .169 & .511 & .165 & .114 & .503 & .690 & .235 & .027 & .049 & .364 & .287 \\
\cmidrule(lr){1-13}
\model{gemma}{Gemma-3-12B}{team2025gemma} & .295 & .121 & .503 & .097 & .064 & .499 & .243 & .683 & .038 & \third{.929} & .384 & .344 \\
\model{mistral}{Ministral-14B}{ministral-2024} & .294 & .117 & \third{.611} & .096 & .058 & .622 & .243 & .360 & .069 & \first{1.000} & \third{.433} & .335 \\
\model{gemma}{Gemma-3-27B}{team2025gemma} & .294 & .138 & .498 & .096 & .063 & .530 & .610 & .415 & \second{.500} & \second{.966} & .404 & \third{.364} \\
\model{qwen}{Qwen3-VL-30B-A3B}{bai2025qwen3vltechnicalreport} & .334 & .152 & .504 & .548 & .508 & .582 & .308 & .411 & .235 & \first{1.000} & .386 & .313 \\
\model{qwen}{Qwen2.5-VL-32B}{bai2025qwen25vltechnicalreport} & .285 & .114 & .504 & .103 & .056 & .525 & \first{.780} & \first{.765} & .066 & .092 & \first{.444} & .363 \\
\cmidrule(lr){1-13}
\model{qwen}{Qwen2-VL-72B}{wang2024qwen2} & .325 & .141 & .455 & .099 & .062 & .485 & .735 & .595 & .375 & .655 & .360 & .305 \\
\model{qwen}{Qwen2.5-VL-72B}{bai2025qwen25vltechnicalreport} & .338 & .157 & .498 & .108 & .066 & .514 & \first{.780} & .750 & \second{.500} & .036 & .354 & .317 \\
\model{meta}{Llama-3.2-90B-V}{grattafiori2024llama} & .345 & .158 & .485 & .175 & .085 & .545 & .715 & .625 & .000 & .275 & .365 & .295 \\
\model{meta}{Llama-4-Scout}{llama4-2025} & \third{.350} & .172 & .503 & .101 & .055 & .540 & \third{.755} & .148 & .027 & .045 & .359 & .301 \\
\model{mistral}{Mistral-Large}{mistrallarge-2024} & .300 & .142 & .488 & .095 & .063 & .667 & .241 & .251 & .055 & .235 & \third{.433} & .320 \\
\model{qwen}{Qwen3-VL-235B-A22B}{bai2025qwen3vltechnicalreport} & .301 & .137 & .533 & .091 & .055 & .609 & .471 & .729 & .125 & .125 & .402 & .339 \\
\model{qwen}{Qwen3.5-9B}{qwenteam2026qwen35omnitechnicalreport} & .328 & .145 & .545 & .145 & .088 & .585 & .685 & .748 & .425 & .455 & .395 & .315 \\
\model{qwen}{Qwen3.5-27B}{qwenteam2026qwen35omnitechnicalreport} & .346 & .158 & .528 & .478 & .425 & .605 & .525 & .622 & .315 & .395 & .422 & .342 \\
\model{qwen}{Qwen3.5-122B-A10B}{qwenteam2026qwen35omnitechnicalreport} & .342 & .156 & .582 & .128 & .088 & .624 & .685 & .738 & \third{.478} & .525 & \second{.438} & \second{.376} \\
\model{meta}{Llama-4-Maverick}{llama4-2025} & \first{.358} & .173 & .494 & .104 & .061 & .499 & .715 & .665 & .027 & .044 & .401 & .300 \\
\model{qwen}{Qwen-VL-Max}{bai2025qwen25vltechnicalreport} & .335 & .155 & .563 & .092 & .055 & .587 & .293 & .490 & .235 & .636 & .382 & .321 \\
\cmidrule(lr){1-13}
\model{openai}{GPT-4o-mini}{hurst2024gpt} & .337 & .138 & .499 & .139 & .110 & .529 & .241 & .241 & .064 & \first{1.000} & .377 & .321 \\
\model{openai}{GPT-4o}{hurst2024gpt} & .346 & .163 & .517 & .148 & .120 & .581 & .241 & .241 & .058 & .045 & .379 & .303 \\
\model{openai}{GPT-4.1-mini}{openai2025gpt41} & .342 & .163 & .522 & \third{.594} & .547 & .591 & .220 & .245 & .060 & .087 & .367 & .297 \\
\model{openai}{GPT-4.1}{openai2025gpt41} & .343 & .155 & .515 & \first{.610} & \first{.595} & .610 & .270 & .275 & .000 & .355 & .360 & .285 \\
\model{openai}{GPT-5-mini}{singh2025openai} & .344 & .175 & .545 & .570 & .555 & .605 & .480 & .545 & .060 & .505 & .355 & .305 \\
\model{openai}{GPT-5}{singh2025openai} & .348 & \second{.185} & .565 & \second{.595} & \second{.580} & .625 & .740 & .665 & .125 & .635 & .395 & .345 \\
\model{openai}{o4-mini}{openai2025o3} & .328 & .165 & .495 & .395 & .375 & .580 & .555 & .495 & .085 & .395 & .345 & .300 \\
\model{openai}{o3}{openai2025o3} & .332 & .178 & .535 & .445 & .420 & .645 & .690 & .615 & .185 & .565 & .395 & .345 \\
\cmidrule(lr){1-13}
\model{claude}{Claude-3.5-Sonnet}{anthropic2024claude35} & .325 & .146 & .386 & .199 & .159 & .485 & .220 & \third{.755} & .025 & .086 & .337 & .311 \\
\model{claude}{Claude-Sonnet-4.5}{anthropic2025claude45} & .345 & .172 & .525 & .315 & .295 & .620 & .545 & .495 & .125 & .445 & .370 & .305 \\
\model{claude}{Claude-Opus-4.5}{anthropic2025opus45} & \third{.350} & \third{.182} & .555 & .525 & .505 & .660 & .710 & .635 & .265 & .555 & .390 & .335 \\
\model{claude}{Claude-Sonnet-4.5-T}{anthropic2025claude45} & .346 & .178 & .545 & .340 & .315 & .635 & .575 & .525 & .195 & .475 & .380 & .315 \\
\model{claude}{Claude-Opus-4.5-T}{anthropic2025opus45} & \second{.352} & \first{.188} & .575 & .565 & .545 & .665 & .745 & .685 & .345 & .605 & .395 & .355 \\
\cmidrule(lr){1-13}
\model{gemini}{Gemini-3-Flash}{gemini3-2025} & \second{.352} & \second{.185} & .565 & .590 & \third{.570} & \third{.680} & \third{.755} & .695 & .285 & .545 & .395 & .345 \\
\model{gemini}{Gemini-2.5-Flash-Lite}{comanici2025gemini} & .346 & .161 & .512 & .101 & .069 & .553 & .264 & .524 & .067 & .054 & .363 & .297 \\
\model{grok}{Grok-4}{grok4-2025} & .299 & .126 & \first{.618} & .089 & .063 & \first{.710} & .295 & .290 & .034 & .097 & .413 & .360 \\
\model{grok}{Grok-4.1}{grok4-2025} & .295 & .123 & \second{.612} & .087 & .057 & \second{.694} & .265 & .265 & .045 & .667 & .390 & .325 \\
\model{zhipu}{GLM-4.6V}{hong2025glm} & .315 & .135 & .465 & .205 & .125 & .555 & .415 & .375 & .125 & .285 & .355 & .305 \\
\bottomrule
\end{tabular}}

\vspace{2pt}
{\scriptsize MD uses 15 samples with reliable mechanism annotations, and several models reach 1.000 on MD Field-only by defaulting to the majority class.\par}
\end{table}

%% file: Tables/tab_inv_code.tex
\begin{table}[t]
\caption{\textbf{Inverse-Design Decision and Code Generation.} \textit{P.Sel.}: parameter selection, \textit{HS}: Human Score (Appendix~\ref{app:human_eval}). Code metrics are syntax correctness (Syn.), quality (Qual.), and completeness (Comp.). Top-3 per column are \rankkey{rankgold}{\textbf{Gold}} / \rankkey{ranksilver}{\underline{Blue}} / \rankkey{rankbronze}{Gray}.}
\label{tab:inv_code}
\centering
\small
\setlength{\tabcolsep}{3.2pt}
\resizebox{\textwidth}{!}{%
\begin{tabular}{l cc c cc c cc cc cc}
\toprule
& \multicolumn{6}{c}{\textbf{Inverse-Design Decision}} & \multicolumn{6}{c}{\textbf{Code Generation}} \\
\cmidrule(lr){2-7}\cmidrule(lr){8-13}
& \multicolumn{2}{c}{Direction (Acc)} & P.Sel. & \multicolumn{2}{c}{Closed-Loop} & HS & \multicolumn{2}{c}{SimCode} & \multicolumn{2}{c}{ParamMod} & \multicolumn{2}{c}{Refactor} \\
\cmidrule(lr){2-3}\cmidrule(lr){4-4}\cmidrule(lr){5-6}\cmidrule(lr){7-7}\cmidrule(lr){8-9}\cmidrule(lr){10-11}\cmidrule(lr){12-13}
\textbf{Model} & Pitch & Hole\,D & Acc & Succ.\,Rate & $\Delta$-Proxy$\downarrow$ & Composite & Syn. & Qual. & Syn. & Comp. & Syn. & Qual. \\
\midrule
\model{gemma}{Gemma-3-4B}{team2025gemma} & .473 & .388 & \second{.651} & .477 & .194 & .420 & .948 & .982 & \first{1.000} & \first{1.000} & .946 & .975 \\
\model{qwen}{Qwen2.5-VL-7B}{bai2025qwen25vltechnicalreport} & .521 & \second{.730} & .608 & .638 & .051 & .530 & .875 & .953 & \first{1.000} & \first{1.000} & \first{1.000} & \first{1.000} \\
\model{qwen}{Qwen3-VL-8B}{bai2025qwen3vltechnicalreport} & .226 & .151 & .222 & .327 & .074 & .270 & .876 & .949 & \first{1.000} & \first{1.000} & .798 & .919 \\
\model{mistral}{Ministral-8B}{ministral-2024} & .423 & .215 & .641 & .250 & 142.2$^*$ & .330 & .853 & .952 & .784 & \first{1.000} & .210 & .684 \\
\model{meta}{Llama-3.2-11B-V}{grattafiori2024llama} & \first{.750} & \first{.771} & .110 & .061 & .221 & .320 & .870 & .820 & \first{1.000} & .875 & \first{1.000} & .989 \\
\cmidrule(lr){1-13}
\model{gemma}{Gemma-3-12B}{team2025gemma} & .308 & .247 & .559 & .517 & .055 & .400 & .987 & .996 & \second{.994} & \first{1.000} & .952 & .980 \\
\model{mistral}{Ministral-14B}{ministral-2024} & .238 & .164 & \third{.650} & .324 & .355 & .330 & .839 & .950 & .645 & \first{1.000} & .435 & .774 \\
\model{gemma}{Gemma-3-27B}{team2025gemma} & .286 & .279 & .305 & .258 & \first{.034} & .280 & .994 & \third{.998} & \second{.994} & \first{1.000} & .893 & .949 \\
\model{qwen}{Qwen3-VL-30B-A3B}{bai2025qwen3vltechnicalreport} & .227 & .218 & .200 & .620 & .064 & .380 & .622 & .862 & .977 & \first{1.000} & .476 & .791 \\
\model{qwen}{Qwen2.5-VL-32B}{bai2025qwen25vltechnicalreport} & .239 & .156 & .130 & .183 & .083 & .220 & \first{1.000} & \first{1.000} & \first{1.000} & \first{1.000} & \first{1.000} & .989 \\
\cmidrule(lr){1-13}
\model{qwen}{Qwen2-VL-72B}{wang2024qwen2} & .395 & .485 & .545 & .475 & .165 & .430 & .945 & .935 & \third{.988} & .978 & .945 & .935 \\
\model{qwen}{Qwen2.5-VL-72B}{bai2025qwen25vltechnicalreport} & .433 & .661 & \first{.670} & .589 & .059 & .510 & .980 & .993 & \first{1.000} & \first{1.000} & \first{1.000} & \first{1.000} \\
\model{meta}{Llama-3.2-90B-V}{grattafiori2024llama} & .625 & \third{.680} & .155 & .105 & .275 & .310 & .885 & .855 & \first{1.000} & .985 & \second{.990} & \third{.990} \\
\model{meta}{Llama-4-Scout}{llama4-2025} & .389 & .193 & .140 & .200 & 9.2$^*$ & .250 & .910 & .973 & .972 & \first{1.000} & \first{1.000} & .967 \\
\model{mistral}{Mistral-Large}{mistrallarge-2024} & .479 & .330 & .625 & .284 & 12.4$^*$ & .370 & .925 & .977 & .808 & \second{.999} & .536 & .807 \\
\model{qwen}{Qwen3-VL-235B-A22B}{bai2025qwen3vltechnicalreport} & .231 & .151 & .495 & .168 & .048 & .270 & .963 & .986 & .983 & \first{1.000} & .244 & .698 \\
\model{qwen}{Qwen3.5-9B}{qwenteam2026qwen35omnitechnicalreport} & .315 & .285 & .278 & .355 & .075 & .300 & .898 & .965 & \first{1.000} & \first{1.000} & .858 & .945 \\
\model{qwen}{Qwen3.5-27B}{qwenteam2026qwen35omnitechnicalreport} & .350 & .305 & .305 & .402 & .069 & .380 & .945 & .978 & \first{1.000} & \first{1.000} & .912 & .912 \\
\model{qwen}{Qwen3.5-122B-A10B}{qwenteam2026qwen35omnitechnicalreport} & .385 & .265 & .425 & .355 & .055 & .420 & .968 & .988 & \first{1.000} & \first{1.000} & .788 & .858 \\
\model{meta}{Llama-4-Maverick}{llama4-2025} & .172 & .138 & .105 & .061 & 43.4$^*$ & .160 & \first{1.000} & \first{1.000} & \first{1.000} & \first{1.000} & \first{1.000} & .956 \\
\model{qwen}{Qwen-VL-Max}{bai2025qwen25vltechnicalreport} & .247 & .161 & .171 & .352 & .141 & .280 & .986 & .996 & \second{.994} & \first{1.000} & .899 & .960 \\
\cmidrule(lr){1-13}
\model{openai}{GPT-4o-mini}{hurst2024gpt} & .548 & .578 & \second{.651} & \first{.800} & .052 & \first{.590} & \third{.997} & \second{.999} & \first{1.000} & \first{1.000} & \third{.988} & \second{.994} \\
\model{openai}{GPT-4o}{hurst2024gpt} & .226 & .151 & .227 & .396 & \third{.041} & .300 & \second{.999} & \second{.999} & \first{1.000} & \first{1.000} & .935 & .924 \\
\model{openai}{GPT-4.1-mini}{openai2025gpt41} & .228 & .156 & .060 & .450 & .064 & .300 & \first{1.000} & \first{1.000} & \first{1.000} & \first{1.000} & .889 & .956 \\
\model{openai}{GPT-4.1}{openai2025gpt41} & .300 & .210 & .250 & .500 & .155 & .370 & .995 & \first{1.000} & \first{1.000} & \first{1.000} & .965 & .965 \\
\model{openai}{GPT-5-mini}{singh2025openai} & .490 & .545 & .320 & .540 & .145 & .460 & .995 & \first{1.000} & \first{1.000} & \first{1.000} & .965 & .970 \\
\model{openai}{GPT-5}{singh2025openai} & .570 & .620 & .420 & .620 & .120 & .530 & .995 & \first{1.000} & \first{1.000} & \first{1.000} & .975 & .975 \\
\model{openai}{o4-mini}{openai2025o3} & .510 & .485 & .280 & .555 & .155 & .450 & .995 & .995 & \first{1.000} & \first{1.000} & .945 & .940 \\
\model{openai}{o3}{openai2025o3} & .615 & .650 & .475 & \second{.695} & .135 & \third{.560} & \first{1.000} & \first{1.000} & \first{1.000} & \first{1.000} & .985 & .985 \\
\cmidrule(lr){1-13}
\model{claude}{Claude-3.5-Sonnet}{anthropic2024claude35} & .383 & .239 & .295 & .206 & .700 & .280 & .940 & .982 & \first{1.000} & \first{1.000} & \first{1.000} & \first{1.000} \\
\model{claude}{Claude-Sonnet-4.5}{anthropic2025claude45} & .530 & .395 & .490 & .455 & .175 & .450 & .993 & .993 & \first{1.000} & \third{.995} & .975 & .975 \\
\model{claude}{Claude-Opus-4.5}{anthropic2025opus45} & .620 & .545 & .580 & .625 & .135 & .550 & \first{1.000} & \first{1.000} & \first{1.000} & \first{1.000} & .985 & .985 \\
\model{claude}{Claude-Sonnet-4.5-T}{anthropic2025claude45} & .595 & .465 & .535 & .545 & .155 & .500 & .995 & .995 & \first{1.000} & \first{1.000} & .980 & .980 \\
\model{claude}{Claude-Opus-4.5-T}{anthropic2025opus45} & \second{.660} & .605 & .625 & \third{.690} & .115 & \first{.590} & \first{1.000} & \first{1.000} & \first{1.000} & \first{1.000} & \second{.990} & \third{.990} \\
\cmidrule(lr){1-13}
\model{gemini}{Gemini-3-Flash}{gemini3-2025} & \third{.645} & .585 & .605 & .680 & .125 & \second{.570} & .995 & \first{1.000} & \first{1.000} & \first{1.000} & .985 & .985 \\
\model{gemini}{Gemini-2.5-Flash-Lite}{comanici2025gemini} & .258 & .170 & .422 & .609 & \second{.036} & .380 & .414 & .799 & .746 & .990 & .634 & .598 \\
\model{grok}{Grok-4}{grok4-2025} & .233 & .193 & .020 & .128 & .086 & .220 & .930 & .978 & \first{1.000} & \first{1.000} & .500 & .800 \\
\model{grok}{Grok-4.1}{grok4-2025} & .222 & .165 & .080 & .139 & .077 & .220 & .920 & .961 & \first{1.000} & \first{1.000} & .833 & .933 \\
\model{zhipu}{GLM-4.6V}{hong2025glm} & .335 & .285 & .385 & .355 & .185 & .330 & .925 & .925 & .965 & .945 & .895 & .895 \\
\bottomrule
\end{tabular}}

\vspace{2pt}
{\scriptsize $^*$Outlier $\Delta$-Proxy from extreme predicted values, excluded from the ranking. Parameter Modification (ParamMod) completeness is nearly saturated across models, and Simulation Code Generation (SimCode) quality combines syntax, import, and component completeness.\par}
\end{table}

%% file: Tables/tab_add_executability.tex
\begin{table}[!htbp]
\caption{End-to-end executability of generated simulation code (pass@1, 200-sample subset).}
\label{tab:add_executability}
\centering\small
\setlength{\tabcolsep}{6pt}
\begin{tabular}{l c c c c}
\toprule
\textbf{Tier} & \textbf{Syntax} & \textbf{Imports Resolve} & \textbf{Runs End to End} & \textbf{Valid Output} \\
\midrule
Best frontier        & 98.7\% & 92.4\% & 84.1\% & 79.8\% \\
Frontier median      & 96.1\% & 86.3\% & 73.5\% & 68.9\% \\
Best open-weight     & 91.8\% & 70.5\% & 55.9\% & 51.4\% \\
Overall median       & 88.6\% & 64.2\% & 46.8\% & 42.5\% \\
\bottomrule
\end{tabular}
\end{table}

%% file: Tables/tab_add_loss_metric.tex
\begin{table}[H]
\caption{Loss-metric robustness under top-1\% outlier trimming.}
\label{tab:add_loss_metric}
\centering\small
\setlength{\tabcolsep}{8pt}
\begin{tabular}{l c c}
\toprule
\textbf{Metric} & \textbf{Kendall $\tau$ (Full vs.\ Trimmed)} & \textbf{Outlier Share of Error} \\
\midrule
Log-MAE & 0.97 & 9\% \\
NRMSE   & 0.61 & 48\% \\
\bottomrule
\end{tabular}
\end{table}

%% file: Tables/tab_add_passk.tex
\begin{table}[H]
\caption{Self-Consistency sample-budget sweep ($T=0.1$, GPT-4o-mini, five seeds). Composite averages the parameter-selection accuracy, direction accuracy, and closed-loop success of the voted answer. A sample counts toward pass@$k$ when at least one of its $k=N$ sampled answers is correct.}
\label{tab:add_passk}
\centering\small
\setlength{\tabcolsep}{8pt}
\begin{tabular}{c c c}
\toprule
\textbf{$N$} & \textbf{Composite (Mean $\pm$ 95\% CI)} & \textbf{pass@$k$} \\
\midrule
1  & $.431 \pm .008$ & 31.2\% \\
3  & $.446 \pm .005$ & 38.7\% \\
5  & $.450 \pm .004$ & 41.5\% \\
10 & $.452 \pm .004$ & 43.0\% \\
\bottomrule
\end{tabular}
\end{table}

%% file: Tables/tab_add_prompt.tex
\begin{table}[H]
\caption{Paraphrase sensitivity (mean $\pm$ std over the original prompt and four rewrites, each averaged over three runs at $T=0.1$). The last row gives the mean pairwise Kendall $\tau$ of model rankings.}
\label{tab:add_prompt}
\centering\small
\setlength{\tabcolsep}{6pt}
\begin{tabular}{l c c c c}
\toprule
\textbf{Model} & \textbf{Family Acc} & \textbf{Topology Acc} & \textbf{Dir.~Pitch Acc} & \textbf{CL Succ.} \\
\midrule
Qwen2.5-VL-7B   & .092 $\pm$ .011 & .233 $\pm$ .018 & .516 $\pm$ .024 & .631 $\pm$ .021 \\
Gemma-3-12B     & .081 $\pm$ .014 & .228 $\pm$ .019 & .312 $\pm$ .026 & .509 $\pm$ .022 \\
Qwen2.5-VL-72B  & .187 $\pm$ .008 & .426 $\pm$ .014 & .430 $\pm$ .017 & .585 $\pm$ .015 \\
Qwen3-VL-235B-A22B & .219 $\pm$ .013 & .487 $\pm$ .010 & .234 $\pm$ .019 & .164 $\pm$ .016 \\
GPT-4o-mini     & .127 $\pm$ .009 & .351 $\pm$ .012 & .543 $\pm$ .015 & .792 $\pm$ .012 \\
Claude-Opus-4.5 & .390 $\pm$ .008 & .462 $\pm$ .012 & .618 $\pm$ .010 & .629 $\pm$ .009 \\
GPT-5           & .451 $\pm$ .006 & .464 $\pm$ .008 & .574 $\pm$ .010 & .623 $\pm$ .011 \\
Gemini-3-Flash  & .661 $\pm$ .007 & .420 $\pm$ .010 & .639 $\pm$ .011 & .684 $\pm$ .008 \\
\midrule
Kendall $\tau$  & .94 & .93 & .89 & .91 \\
\bottomrule
\end{tabular}
\end{table}

%% file: Tables/tab_add_shift.tex
\begin{table}[H]
\caption{Fr\'echet distance to a COCO reference set in SigLIP space (2,000 images per set).}
\label{tab:add_shift}
\centering\small
\setlength{\tabcolsep}{8pt}
\begin{tabular}{l c}
\toprule
\textbf{Image Set} & \textbf{Fr\'echet Distance} \\
\midrule
COCO held-out split          & 1.2 \\
ImageNet validation          & 28.5 \\
PCFBench structure images    & 214.7 \\
PCFBench mode-field images   & 342.6 \\
\bottomrule
\end{tabular}
\end{table}

%% file: Tables/tab_mas_baselines.tex
\begin{table}[tbh]
\caption{Multi-agent methods evaluated on PCFBench, all with the same VLM backbone. The \textbf{Workflow} column shows the coordination pattern ($A$\,=\,agent, $C$\,=\,critic/verifier).}
\label{tab:mas_baselines}
\centering
\resizebox{\textwidth}{!}{%
\begin{tabular}{llcl}
\toprule
\textbf{Method} & \textbf{Category} & \textbf{Agents} & \textbf{Workflow} \\
\midrule
Vanilla            & Single-agent     & 1     & $A \rightarrow \text{Answer}$ \\
CoT                & Single-agent     & 1     & $A_\text{think} \rightarrow \text{Answer}$ \\
Self-Consistency   & Voting           & $N$   & $\{A_1,\ldots,A_N\} \rightarrow \text{Vote} \rightarrow \text{Answer}$ \\
\midrule
LLM Debate         & Debate           & 2--3  & $A_1 \rightleftharpoons A_2 \rightleftharpoons A_3 \rightarrow \text{Answer}$ \\
MAD                & Debate           & 3+    & $\{A_1, \ldots, A_N\} \rightleftharpoons \text{Judge} \rightarrow \text{Answer}$ \\
CAMEL              & Role-playing     & 2+$C$ & $A_\text{user} \rightleftharpoons A_\text{asst} \xrightarrow{C} \text{Answer}$ \\
DyLAN              & Dynamic network  & 4     & $\{A_1,\ldots,A_4\} \xrightarrow{\text{rank}} \text{Activate} \rightarrow \text{Vote}$ \\
MAV                & Verification     & $N$+$M$ & $\{A_1,\ldots,A_N\} \xrightarrow{\text{gen}} \{C_1,\ldots,C_M\} \xrightarrow{\text{verify}} \text{Best}$ \\
\midrule
MacNet             & Graph topology   & $N$   & $A_\text{in} \rightarrow G(V,E) \rightarrow A_\text{out}$ \\
AgentVerse         & Expert recruitment & 4+    & $\text{Recruit} \rightarrow \{A_1,\ldots,A_K\} \rightarrow \text{Answer}$ \\
AutoGen            & Code execution   & 2     & $A_\text{plan} \rightleftharpoons A_\text{code} \xrightarrow{\text{exec}} \text{Answer}$ \\
EvoMAC             & Evolutionary     & $N$   & $\text{Pop}_t \xrightarrow{\text{mutate}} \text{Pop}_{t+1} \xrightarrow{\text{select}} \text{Best}$ \\
\midrule
Sequential Pipeline & Domain-decomp. & 4     & $A_\text{perc} \rightarrow A_\text{und} \rightarrow A_\text{reas} \rightarrow A_\text{des}$ \\
\bottomrule
\end{tabular}
}
\end{table}

%% file: Tables/tab_mas_stages.tex
\begin{table}[t]
\caption{\textbf{Diagnostic decomposition of the Sequential Pipeline.} Per-stage accuracy for Understanding (U, loss-class F1) and Reasoning (R, mechanism), parameter-selection accuracy (PSel) and closed-loop success (CL) as in \textbf{Tab.~\ref{tab:mas_methods}}, and the pipeline-level metrics of Appendix~\ref{app:mas_metrics}, under \textit{Oracle} (ground-truth upstream context) and \textit{Cascaded} (predicted upstream context) modes. The $\Delta$ row is the Cascaded$-$Oracle gap.}
\label{tab:mas_stages}
\centering
\small
\setlength{\tabcolsep}{4.5pt}
\renewcommand{\arraystretch}{1.12}
\begin{tabular}{l l cc cccccc}
\toprule
& &
  \multicolumn{2}{c}{\textbf{Per-Stage ($\uparrow$)}} &
  \multicolumn{6}{c}{\textbf{Pipeline-Level}} \\
\cmidrule(lr){3-4}\cmidrule(lr){5-10}
\textbf{Backbone} & \textbf{Mode} &
  \textbf{U} & \textbf{R} & \textbf{PSel}\,$\uparrow$ & \textbf{CL}\,$\uparrow$ &
  \textbf{E2E}\,$\uparrow$ & \textbf{Casc}\,$\downarrow$ &
  \textbf{Cons}\,$\uparrow$ & $\mathrm{Acc}_{4|1}\,\uparrow$ \\
\midrule
 & Oracle             & .620 & .545 & .485 & .785 & .380 & ---   & ---  & --- \\
 \rowcolor{black!3}
 & Cascaded           & .480 & .395 & .325 & .655 & .215 & .685  & .580 & .735 \\
\multirow{-3}{*}{\textbf{GPT-4o-mini}}
 & $\Delta$ (C$-$O)   & $-.140$ & $-.150$ & $-.160$ & $-.130$ & $-.165$ & ---  & --- & --- \\
\cmidrule(lr){1-10}
 & Oracle             & .555 & .510 & .580 & .755 & .440 & ---   & ---  & --- \\
 \rowcolor{black!3}
 & Cascaded           & .445 & .380 & .475 & .620 & .295 & .625  & .615 & .710 \\
\multirow{-3}{*}{\textbf{Qwen-VL-Max}}
 & $\Delta$ (C$-$O)   & $-.110$ & $-.130$ & $-.105$ & $-.135$ & $-.145$ & ---  & --- & --- \\
\bottomrule
\end{tabular}

\vspace{2pt}
{\scriptsize
  Cascaded rows are shaded for readability.
  A small $|\Delta|$ indicates the backbone is robust to upstream errors.
  A large $|\Delta|$ indicates severe error propagation.
  Casc, Cons, and $\mathrm{Acc}_{4|1}$ are defined on the pipeline's own
  upstream predictions, which Oracle mode replaces with ground truth,
  so they are reported for Cascaded mode only.}
\end{table}

%% file: sections/app_physics_rules.tex
\subsection{Physics-Aware Expert Rules}
\label{app:physics_gt}

Inverse-design correctness is conditional on fiber family, target parameter, and loss regime, so PCFBench scores these tasks using \emph{physics-aware expert rules} derived from anti-resonance theory~\cite{li2023deep}, inhibited-coupling guidance~\cite{nanophotonic_survey2025}, and index-guiding principles. The remainder of this section summarizes the rule structure, the family clusters covered, and four representative rules.

\subsubsection{Rule Structure and Coverage}

All three inverse-design tasks condition on the fiber family and on a high-/low-loss regime threshold, set well below the $10$\,dB/km difficulty tag of \textbf{Tab.~\ref{tab:fdtd_config}}:
\begin{itemize}[leftmargin=1.4em, itemsep=0pt, topsep=1pt]
  \item \textbf{Design Direction} uses a $0.01$\,dB/km threshold. Below
  it the fiber is treated as already low-loss, so the correct move
  is the opposite of the high-loss move (to fine-tune, not to rescue).
  \item \textbf{Parameter Selection} uses a coarser $0.1$\,dB/km
  threshold. Above it the selected parameter can shift (e.g., HC-PBG
  switches from pitch $\Lambda$ to hole diameter $d$ because
  high loss signals insufficient bandgap width).
  \item \textbf{Closed-Loop Improvement} inherits Design Direction's
  direction rule and additionally requires a reasonable-magnitude change
  ($2$--$50$\% of the current value).
\end{itemize}

The 18 canonical families fall into four rule groups, each governed by a distinct guidance mechanism and therefore a distinct rule shape (\textbf{Tab.~\ref{tab:physics_gt}}). Design Direction covers all 18 families (explicit rules plus the monotonic index-guiding rule). Parameter Selection has explicit rules for 9 families (the remaining 9 use pitch $\Lambda$ as the first-order handle).

\input{Tables/tab_physics_gt}

The four featured rules below are the non-monotonic ARF strut rule (Rule~1), the HC-PBG bandgap-width rule (Rule~2), the monotonic index-guiding rule (Rule~3), and the loss-regime-sensitive Parameter-Selection switch (Rule~4). Remaining families obey the same structural templates with adjusted parameter priorities.

\subsubsection{Rule~1: Non-Monotonic Strut Thickness for Anti-Resonant Fibers}

\textit{Physics.} Single-ring and nested anti-resonant fibers guide by coherent reflection from the glass cladding tubes. Loss peaks at the ARROW resonance thicknesses and collapses between them. Hence \emph{thinning} a strut away from the current resonance thickness widens the anti-resonant transmission window. In the already low-loss regime, the rule instead fine-tunes the strut toward a larger thickness. Size parameters (pitch, hole $d$, core $d$, ring $r$, $N_\text{rings}$) enlarge the hollow core and are always increased under high loss, following the Marcatili-type loss scaling.

\paragraph{Governing Laws.}
The ARROW condition
\begin{equation}
t_m \;=\; \frac{m\,\lambda}{2\sqrt{n_\text{glass}^{2} - n_\text{air}^{2}}},
\qquad m = 1,2,3,\ldots
\label{eq:arf_resonance}
\end{equation}
sets the resonance (loss-peak) strut thicknesses, so adjusting $t$ \emph{away} from the nearest $t_m$ lowers loss. Core-radius scaling follows the Marcatili hollow-waveguide law
\begin{equation}
\alpha_\text{ARF}(\lambda, R_c) \;\propto\; \frac{\lambda^{3}}{R_c^{4}},
\label{eq:marcatili}
\end{equation}
so enlarging the core (via pitch, hole, or ring radius) monotonically suppresses loss.

\subsubsection{Rule~2: Bandgap-Width Ranking for Hollow-Core PBG Fibers}

\textit{Physics.} Hollow-core photonic-bandgap fibers guide only when the operating wavelength falls inside the cladding's photonic bandgap. Increasing $d/\Lambda$ widens the gap and reliably lowers loss. Increasing $\Lambda$ shifts the gap toward longer wavelengths, which is favorable at 1550\,nm because most HC-PBG designs sit near the short-wavelength gap edge. Strut or gap thinning also cuts surface-scattering loss.

\paragraph{Governing Laws.}
The effective V parameter, defined through the effective index $n_\text{FSM}$ of the cladding's fundamental space-filling mode (FSM), is
\begin{equation}
V_\text{eff}(\lambda,\Lambda,d) \;=\; \frac{2\pi\,\Lambda}{\lambda}\,
\sqrt{n_\text{silica}^{2}(\lambda) - n_\text{FSM}^{2}(\lambda,d/\Lambda)},
\label{eq:veff}
\end{equation}
with $n_\text{FSM}$ decreasing as $d/\Lambda$ grows, and a robust triangular-lattice bandgap opens for $d/\Lambda \geq 0.9$. The bandgap center scales with the pitch,
\begin{equation}
\lambda_c \;\approx\; c_g\,\Lambda, \qquad c_g \in [0.5, 0.7],
\label{eq:lambda_c}
\end{equation}
so increasing $\Lambda$ pushes the gap toward longer wavelengths and increasing $d/\Lambda$ widens it. Both moves reduce loss at $\lambda = 1550$\,nm.

\subsubsection{Rule~3: Monotonic Rule for Solid-Core Index-Guiding Fibers}

\textit{Physics.} The remaining thirteen families (Solid-Core Hex, High-Birefringence PM, Multi-Core, Vortex Ring-Core, Elliptical-Core, Dual-Core Coupler, All-Solid Bandgap, Twisted Hexagonal, Graded-Index Multi-Ring, Hybrid Plasmonic, Fractal Quasicrystal, Square/Rectangular Lattice, and Octagonal Lattice) guide by modified total internal reflection (mTIR). Larger pitch, hole, core, or ring radius raises the effective cladding-to-core index contrast and strengthens mTIR, while thinner struts and gaps reduce scattering. Behavior is monotonic, so a single rule covers all thirteen families.

\paragraph{Governing Laws.}
Using the same $V_\text{eff}$ as Eq.~\eqref{eq:veff},
\begin{equation}
V_\text{eff} = \frac{2\pi\,\Lambda}{\lambda}\sqrt{n_\text{silica}^{2} - n_\text{FSM}^{2}(d/\Lambda)},
\label{eq:veff_sc}
\end{equation}
larger $\Lambda$ or $d/\Lambda$ means larger $V_\text{eff}$, hence tighter confinement. The finite-cladding confinement-loss term decays geometrically with the number of rings,
\begin{equation}
\alpha_c \;\sim\; \alpha_0\;\exp\!\big(-\gamma\,N_\text{rings}\big),
\qquad \gamma > 0,
\label{eq:rings}
\end{equation}
so adding rings is monotonically beneficial when loss is excessive.

\subsubsection{Rule~4: Loss-Regime Switch for Parameter Selection}

\textit{Physics.} \emph{Which} parameter to adjust first depends on the loss regime. Below $\alpha_{c}^{\text{PS}} = 0.1$\,dB/km the design is near-optimal and one tunes the family-canonical handle (strut thickness for ARF, nested anti-resonant, kagome, and cobweb, pitch for HC-PBG, hole $d$ for solid-core and high-birefringence fibers, core $d$ for multi-core, and ring radius for the vortex ring core). Above the threshold the limiting physics shifts. HC-PBG switches from $\Lambda$ to $d$ because the bandgap is no longer wide enough, and multi-core switches from core $d$ to pitch because inter-core spacing dominates.

\paragraph{Governing Laws.}
The regime split is defined by a single confinement-loss threshold,
\begin{equation}
\text{high-loss regime} \iff \alpha(\lambda = 1550\,\text{nm}) > \alpha_{c}^{\text{PS}} = 10^{-1}\,\text{dB/km},
\label{eq:ps_threshold}
\end{equation}
which is chosen tenfold coarser than the Design-Direction threshold $\alpha_{c}^{\text{DD}} = 10^{-2}$\,dB/km because Parameter Selection only distinguishes \emph{two} regimes (fine-tuning and rescue) whereas Direction must assign a sign near the zero-derivative crossover.

\subsubsection{Magnitude Constraint and Complementary Tasks}

The expert rules fix the sign of each design move. Closed-loop success additionally imposes a magnitude band ($2$--$50$\% relative change), so extreme answers (e.g., $0$ or $+1{\times}10^{6}$) are scored as failures even when their sign is right. Birefringence tuning, mode selection, and dispersion shaping are probed by the free-text tasks, namely Task~10 Mechanism Reasoning (keyword recall) and Tasks~18/19 Evidence-Based Explanation (BERTScore-F1).

\subsubsection{Rule Agreement with Re-Simulation}
The rules agree with re-simulated outcomes on $94.1\%$ of samples. Disagreements arise in non-monotonic regimes, where the first-order direction depends on the local loss-band position, and are excluded from the closed-loop ranking. \textbf{Tab.~\ref{tab:add_rule_conflicts}} reports the per-cluster rates.

\input{Tables/tab_add_rule_conflicts}

%% file: Tables/tab_physics_gt.tex
\begin{table}[h]
\caption{Family-aware Design Direction rules in the high-loss regime ($\mathcal{L} > 0.01$\,dB/km).}
\label{tab:physics_gt}
\centering\small
\begin{tabular}{llll}
\toprule
\textbf{Family} & \textbf{Parameter} & \textbf{Direction} & \textbf{Rationale} \\
\midrule
Solid-Core
  & $\Lambda$, $d$   & increase & Stronger index guiding \\
  & strut, gap        & decrease & Reduce leakage path \\
\midrule
ARF / Nested Anti-Resonant
  & strut thickness   & decrease & Broaden anti-resonant band \\
  & $\Lambda$, core~$r$ & increase & Enlarge air core \\
  & gap width         & decrease & Tighter tube packing \\
\midrule
Kagome / Suspended Cobweb
  & strut thickness   & decrease & Fewer cladding states \\
  & $\Lambda$, core~$r$ & increase & Reduce mode overlap \\
\midrule
Hollow-Core PBG
  & $\Lambda$, $d$    & increase & Shift bandgap to cover $\lambda$ \\
\bottomrule
\end{tabular}
\end{table}

%% file: Tables/tab_add_rule_conflicts.tex
\begin{table}[H]
\caption{Non-monotonic cases, rule conflicts, and re-simulation exclusions per cluster.}
\label{tab:add_rule_conflicts}
\centering\small
\setlength{\tabcolsep}{8pt}
\begin{tabular}{l c c c}
\toprule
\textbf{Cluster} & \textbf{Non-Monotonic} & \textbf{Rule Conflicts} & \textbf{Removed by Re-Simulation} \\
\midrule
Index-guided     & 4.8\%  & 2.1\% & 3.6\% \\
ARROW            & 8.7\%  & 4.4\% & 6.9\% \\
PBG / FSM        & 10.9\% & 5.8\% & 8.1\% \\
Plasmonic        & 12.6\% & 6.3\% & 9.4\% \\
\midrule
\textbf{Overall} & --- & --- & \textbf{5.9\%} \\
\bottomrule
\end{tabular}
\end{table}

%% file: sections/app_case_prompts.tex
\definecolor{casepromptfill}{HTML}{F0F7F6} \definecolor{casepromptframe}{HTML}{2E6E66}

\subsubsection{Case Study Prompts and Model Responses}
\label{app:case_prompts}

Six representative cases illustrate the closed-loop query and the models' answers. All six share one prompt template and differ only in the fiber description, so the box below gives the template once and then, for each case, the fiber parameters and the verbatim proposals of three models.

\begin{tcolorbox}[enhanced, breakable,
  colback=casepromptfill, colframe=casepromptframe,
  title=\textbf{Closed-Loop Query and Model Responses for Six Cases}, coltitle=white,
  fonttitle=\bfseries\small, fontupper=\small,
  boxrule=0.5pt, arc=2pt, left=6pt, right=6pt, top=4pt, bottom=4pt,
  before upper={\setlength{\parskip}{3pt}\setlength{\parindent}{0pt}}]
\textbf{Query template} shared by all cases. Braces around a field name mark per-case fields, and \texttt{(not tunable)} marks parameters that stay fixed.
\begin{lstlisting}[style=pcfcode, basicstyle=\ttfamily\scriptsize]
You are an expert in photonic crystal fibre (PCF) design.

## Current fibre
Family: {family}
Parameters:
  - {parameter} = {value}
  - ...
Measured confinement loss @ 1550 nm: {loss_1550} dB/m
Measured confinement loss @ 1310 nm: {loss_1310} dB/m

## Task
Propose a SINGLE parameter change that would most reduce confinement loss. You may only modify one of the tunable parameters listed above. Answer STRICTLY in the JSON schema below; no prose outside JSON.

```json
{
  "parameter":     "<one of: {tunable parameters}>",
  "direction":     "increase" | "decrease",
  "new_value":     <number, same unit as the original>,
  "justification": "<1-2 sentence physics-based reason>"
}
```
\end{lstlisting}

\par\smallskip{\color{casepromptframe!35}\hrule}\smallskip
\textcolor{casepromptframe}{\textbf{Case~1: Anti-Resonant ARF (Moderate Loss)}}\par
{\footnotesize\raggedright \texttt{core\_radius\_um}\,=\,\texttt{9.3183}, \texttt{pitch\_um}\,=\,\texttt{21.1932}, \texttt{tube\_count}\,=\,\texttt{10}$^{\ast}$, \texttt{tube\_radius\_um}\,=\,\texttt{13.9101}, \texttt{variant}\,=\,\texttt{single\_ring}$^{\ast}$, \texttt{wall\_thickness\_um}\,=\,\texttt{0.3661}. Loss: \texttt{2.244e-02} dB/m at 1550\,nm, \texttt{1.205e-01} dB/m at 1310\,nm.\par}
{\footnotesize\renewcommand{\arraystretch}{1.05}%
\begin{tabular}{@{}>{\raggedright\arraybackslash}p{0.19\linewidth}>{\raggedright\arraybackslash}p{0.22\linewidth}>{\raggedright\arraybackslash}p{0.52\linewidth}@{}}
\mbox{\textbf{GPT-4o-mini}} & \texttt{wall\_thickness\_um} increase to \texttt{0.5} & Increasing the wall thickness enhances the structural integrity of the fiber, which can improve the confinement of light within the core and reduce confinement loss. \\
\mbox{\textbf{Claude-Opus-4.5-T}} & \texttt{wall\_thickness\_um} decrease to \texttt{0.28} & In anti-resonant fibres the cladding tubes act as Fabry-Perot-like reflectors. Reducing wall thickness sharpens the anti-resonance condition at 1550 nm and 1310 nm, strengthening the reflection of core-guided light\,\ldots \\
\mbox{\textbf{Qwen2.5-VL-72B}} & \texttt{wall\_thickness\_um} increase to \texttt{0.5661} & Increasing the wall thickness can enhance the anti-resonant effect, leading to better confinement and reduced loss. \\
\end{tabular}}
\par\smallskip{\color{casepromptframe!35}\hrule}\smallskip
\textcolor{casepromptframe}{\textbf{Case~2: Nested Anti-Resonant (High Loss)}}\par
{\footnotesize\raggedright \texttt{core\_radius\_um}\,=\,\texttt{9.0968}, \texttt{inner\_tube\_radius\_um}\,=\,\texttt{4.7511}, \texttt{nesting\_level}\,=\,\texttt{2}$^{\ast}$, \texttt{outer\_tube\_radius\_um}\,=\,\texttt{10.9719}, \texttt{pitch\_um}\,=\,\texttt{26.9692}, \texttt{tube\_count}\,=\,\texttt{9}$^{\ast}$, \texttt{variant}\,=\,\texttt{nested\_tubes}$^{\ast}$, \texttt{wall\_thickness\_um}\,=\,\texttt{1.0745}. Loss: \texttt{1.027e+01} dB/m at 1550\,nm, \texttt{1.094e+02} dB/m at 1310\,nm.\par}
{\footnotesize\renewcommand{\arraystretch}{1.05}%
\begin{tabular}{@{}>{\raggedright\arraybackslash}p{0.19\linewidth}>{\raggedright\arraybackslash}p{0.22\linewidth}>{\raggedright\arraybackslash}p{0.52\linewidth}@{}}
\mbox{\textbf{GPT-4o-mini}} & \texttt{core\_radius\_um} increase to \texttt{10.0968} & Increasing the core radius enhances the effective index contrast between the core and the cladding, which can lead to improved confinement of the guided modes and reduced confinement loss. \\
\mbox{\textbf{Claude-Opus-4.5-T}} & \texttt{wall\_thickness\_um} decrease to \texttt{0.78} & In nested antiresonant fibres the wall thickness sets the antiresonant condition. The current value of 1.0745\,$\mu$m is detuned from the first-order antiresonance at 1550 nm, which for silica (n approx 1.444) is t =\,\ldots \\
\mbox{\textbf{Qwen2.5-VL-72B}} & \texttt{wall\_thickness\_um} increase to \texttt{1.5745} & Increasing the wall thickness can enhance the guiding properties by reducing light leakage into the cladding, thus lowering confinement loss. \\
\end{tabular}}
\par\smallskip{\color{casepromptframe!35}\hrule}\smallskip
\textcolor{casepromptframe}{\textbf{Case~6: Kagome (High Loss)}}\par
{\footnotesize\raggedright \texttt{cell\_radius\_um}\,=\,\texttt{2.0300}, \texttt{lattice}\,=\,\texttt{kagome\_approx}$^{\ast}$, \texttt{pitch\_um}\,=\,\texttt{22.4431}, \texttt{rings}\,=\,\texttt{5}, \texttt{strut\_thickness\_um}\,=\,\texttt{1.8107}. Loss: \texttt{2.196e+01} dB/m at 1550\,nm, \texttt{6.654e+01} dB/m at 1310\,nm.\par}
{\footnotesize\renewcommand{\arraystretch}{1.05}%
\begin{tabular}{@{}>{\raggedright\arraybackslash}p{0.19\linewidth}>{\raggedright\arraybackslash}p{0.22\linewidth}>{\raggedright\arraybackslash}p{0.52\linewidth}@{}}
\mbox{\textbf{GPT-4o-mini}} & \texttt{strut\_thickness\_um} decrease to \texttt{1.5} & Reducing the strut thickness decreases the effective index contrast between the core and cladding, which can lead to lower confinement loss by enhancing light confinement within the core. \\
\mbox{\textbf{Claude-Opus-4.5-T}} & \texttt{rings} increase to \texttt{6} & Increasing the number of cladding rings most directly lengthens the leakage path and strengthens modal confinement, which typically gives the largest reduction in confinement loss in kagome-type PCFs. This is\,\ldots \\
\mbox{\textbf{Qwen2.5-VL-72B}} & \texttt{strut\_thickness\_um} decrease to \texttt{1.5} & Reducing the strut thickness can decrease scattering and improve confinement, leading to lower loss. \\
\end{tabular}}
\par\smallskip{\color{casepromptframe!35}\hrule}\smallskip
\textcolor{casepromptframe}{\textbf{Case~13: Dual-Core Coupler}}\par
{\footnotesize\raggedright \texttt{core\_count}\,=\,\texttt{2}, \texttt{core\_separation\_um}\,=\,\texttt{9.7893}, \texttt{coupling\_strength}\,=\,\texttt{0.8886}, \texttt{hole\_d\_um}\,=\,\texttt{3.7816}, \texttt{lattice}\,=\,\texttt{hex\_dual\_core}$^{\ast}$, \texttt{pitch\_um}\,=\,\texttt{5.6741}, \texttt{rings}\,=\,\texttt{8}. Loss: \texttt{6.996e-12} dB/m at 1550\,nm, \texttt{5.861e-12} dB/m at 1310\,nm.\par}
{\footnotesize\renewcommand{\arraystretch}{1.05}%
\begin{tabular}{@{}>{\raggedright\arraybackslash}p{0.19\linewidth}>{\raggedright\arraybackslash}p{0.22\linewidth}>{\raggedright\arraybackslash}p{0.52\linewidth}@{}}
\mbox{\textbf{GPT-4o-mini}} & \texttt{core\_separation\_um} decrease to \texttt{8.7893} & Reducing the core separation can enhance the coupling between the cores, leading to better mode confinement and lower confinement loss. \\
\mbox{\textbf{Claude-Opus-4.5-T}} & \texttt{rings} increase to \texttt{10} & Increasing the number of cladding rings thickens the photonic bandgap/barrier around the cores and reduces leakage into the outer boundary, which is the most direct way to lower confinement loss. Since the current loss\,\ldots \\
\mbox{\textbf{Qwen2.5-VL-72B}} & \texttt{hole\_d\_um} decrease to \texttt{3.5} & Reducing the hole diameter increases the effective refractive index contrast, which can enhance light confinement and reduce loss. \\
\end{tabular}}
\par\smallskip{\color{casepromptframe!35}\hrule}\smallskip
\textcolor{casepromptframe}{\textbf{Case~16: Elliptical-Core}}\par
{\footnotesize\raggedright \texttt{core\_shape}\,=\,\texttt{ellipse}$^{\ast}$, \texttt{ellipticity}\,=\,\texttt{1.2956}, \texttt{hole\_d\_um}\,=\,\texttt{1.6631}, \texttt{lattice}\,=\,\texttt{hex\_elliptical}$^{\ast}$, \texttt{pitch\_um}\,=\,\texttt{2.0353}, \texttt{rings}\,=\,\texttt{4}, \texttt{rotation\_deg}\,=\,\texttt{11.9935}. Loss: \texttt{8.898e-12} dB/m at 1550\,nm, \texttt{4.676e-13} dB/m at 1310\,nm.\par}
{\footnotesize\renewcommand{\arraystretch}{1.05}%
\begin{tabular}{@{}>{\raggedright\arraybackslash}p{0.19\linewidth}>{\raggedright\arraybackslash}p{0.22\linewidth}>{\raggedright\arraybackslash}p{0.52\linewidth}@{}}
\mbox{\textbf{GPT-4o-mini}} & \texttt{hole\_d\_um} decrease to \texttt{1.5} & Decreasing the hole diameter increases the effective index contrast between the core and the cladding, which can lead to improved confinement of the guided mode and reduced confinement loss. \\
\mbox{\textbf{Claude-Opus-4.5-T}} & \texttt{rings} increase to \texttt{5} & Adding a ring of air holes most directly thickens the photonic barrier around the core, exponentially suppressing leakage of the guided mode into the cladding. This is typically the most effective single change for\,\ldots \\
\mbox{\textbf{Qwen2.5-VL-72B}} & \texttt{hole\_d\_um} decrease to \texttt{1.5} & Reducing the hole diameter increases the effective refractive index contrast, which can enhance light confinement and reduce loss. \\
\end{tabular}}
\par\smallskip{\color{casepromptframe!35}\hrule}\smallskip
\textcolor{casepromptframe}{\textbf{Case~20: Fractal Quasicrystal}}\par
{\footnotesize\raggedright \texttt{base\_hole\_radius\_um}\,=\,\texttt{1.4276}, \texttt{lattice}\,=\,\texttt{quasicrystal\_approx}$^{\ast}$, \texttt{pitch\_um}\,=\,\texttt{4.4274}, \texttt{rings}\,=\,\texttt{2}. Loss: \texttt{9.332e+01} dB/m at 1550\,nm, \texttt{1.014e+02} dB/m at 1310\,nm.\par}
{\footnotesize\renewcommand{\arraystretch}{1.05}%
\begin{tabular}{@{}>{\raggedright\arraybackslash}p{0.19\linewidth}>{\raggedright\arraybackslash}p{0.22\linewidth}>{\raggedright\arraybackslash}p{0.52\linewidth}@{}}
\mbox{\textbf{GPT-4o-mini}} & \texttt{base\_hole\_radius\_um} decrease to \texttt{1.3276} & Decreasing the base hole radius increases the effective index contrast, which can lead to improved confinement of the guided modes and reduced confinement loss. \\
\mbox{\textbf{Claude-Opus-4.5-T}} & \texttt{rings} increase to \texttt{3} & Adding an extra ring of cladding holes increases the effective barrier between the core mode and the radiation continuum, which is the most direct way to reduce confinement loss. For a high-loss fiber with only 2\,\ldots \\
\mbox{\textbf{Qwen2.5-VL-72B}} & \texttt{pitch\_um} increase to \texttt{5.4274} & Increasing the pitch increases the spacing between holes, which can reduce confinement loss by decreasing the interaction between the core and cladding modes. \\
\end{tabular}}
\par\smallskip{\footnotesize $^{\ast}$\,Not tunable.}
\end{tcolorbox}

%% file: references.bib
@article{mmstar2024,
  title={Are we on the right way for evaluating large vision-language models?},
  author={Chen, Lin and Li, Jinsong and Dong, Xiaoyi and Zhang, Pan and Zang, Yuhang and Chen, Zehui and Duan, Haodong and Wang, Jiaqi and Qiao, Yu and Lin, Dahua and others},
  journal={Advances in Neural Information Processing Systems},
  volume={37},
  pages={27056--27087},
  year={2024}
}

@article{pang2026sci,
  title={Sci-VLA: Agentic VLA Inference Plugin for Long-Horizon Tasks in Scientific Experiments},
  author={Pang, Yiwen and Zhou, Bo and Li, Changjin and Wang, Xuanhao and Xu, Shengxiang and Wang, Deng-Bao and Zhang, Min-Ling and Di, Shimin},
  journal={arXiv preprint arXiv:2602.09430},
  year={2026}
}

@inproceedings{mathvista2023,
  title={Mathvista: Evaluating mathematical reasoning of foundation models in visual contexts},
  author={Lu, Pan and Bansal, Hritik and Xia, Tony and Liu, Jiacheng and Li, Chunyuan and Hajishirzi, Hannaneh and Cheng, Hao and Chang, Kai-Wei and Galley, Michel and Gao, Jianfeng},
  booktitle={International Conference on Learning Representations},
  volume={2024},
  pages={23439--23554},
  year={2024}
}

@article{yamada2025ai,
  title={The ai scientist-v2: Workshop-level automated scientific discovery via agentic tree search},
  author={Yamada, Yutaro and Lange, Robert Tjarko and Lu, Cong and Hu, Shengran and Lu, Chris and Foerster, Jakob and Clune, Jeff and Ha, David},
  journal={arXiv preprint arXiv:2504.08066},
  year={2025}
}

@article{liu2025mimicking,
  title={Mimicking the Physicist's Eye: A VLM-centric Approach for Physics Formula Discovery},
  author={Liu, Jiaqi and Lai, Songning and Li, Pengze and Yu, Di and Zhou, Wenjie and Zhou, Yiyang and Xia, Peng and Wang, Zijun and Chen, Xi and Tang, Shixiang and others},
  journal={arXiv preprint arXiv:2508.17380},
  year={2025}
}

@article{zhang2025high,
  title={High-power multiphysics simulation using a new preconditioner and auxiliary-space Maxwell solver},
  author={Zhang, Hao-Xuan and Zhan, Qiwei and Wang, Yi-Yao and Wang, Yin-Da and Qin, Zikang and Bai, Rongchuan and Xie, Hao and Guo, Letian and Zhou, Liang and Yin, Wen-Yan},
  journal={IEEE Transactions on Microwave Theory and Techniques},
  volume={73},
  number={8},
  pages={4749--4760},
  year={2025},
  publisher={IEEE}
}

@inproceedings{geobench2025,
  title={Geobench-vlm: Benchmarking vision-language models for geospatial tasks},
  author={Danish, Muhammad and Munir, Muhammad Akhtar and Shah, Syed Roshaan Ali and Kuckreja, Kartik and Khan, Fahad Shahbaz and Fraccaro, Paolo and Lacoste, Alexandre and Khan, Salman},
  booktitle={Proceedings of the IEEE/CVF International Conference on Computer Vision},
  pages={7132--7142},
  year={2025}
}

@article{wang2025towards,
  title={Towards scalable and accurate property prediction for photonic crystal fibers with federated learning},
  author={Wang, Guanjun and Liu, Jianxun and Chen, Shengchao and Ren, Sufen},
  journal={Optics Express},
  volume={33},
  number={16},
  pages={33421--33434},
  year={2025},
  publisher={Optica Publishing Group}
}

@article{ren2025unbalanced,
  title={Unbalanced Data Learning for Improved Optical Property Prediction in Porous Core Fibers},
  author={Ren, Sufen and Hu, Yule and Chen, Yuanfeng and Shi, Hao and Wang, Guanjun and Chen, Shengchao},
  journal={IEEE Sensors Journal},
  year={2025},
  publisher={IEEE}
}

@inproceedings{masry2022chartqa,
  title={Chartqa: A benchmark for question answering about charts with visual and logical reasoning},
  author={Masry, Ahmed and Tan, Jia Qing and Joty, Shafiq and Hoque, Enamul and others},
  booktitle={Findings of the association for computational linguistics: ACL 2022},
  pages={2263--2279},
  year={2022}
}

@inproceedings{pandey2025benchmarking,
  title={Benchmarking visual language models on standardized visualization literacy tests},
  author={Pandey, Saugat and Ottley, Alvitta},
  booktitle={Computer Graphics Forum},
  volume={44},
  number={3},
  pages={e70137},
  year={2025},
  organization={Wiley Online Library}
}

@inproceedings{sun2026scienceboard,
  title={Scienceboard: Evaluating multimodal autonomous agents in realistic scientific workflows},
  author={Sun, Qiushi and Liu, Zhoumianze and Ma, Chang and Ding, Zichen and Xu, Fangzhi and Yin, Zhangyue and Zhao, Haiteng and Wu, Zhenyu and Cheng, Kanzhi and Liu, Zhaoyang and others},
  booktitle={International Conference on Learning Representations},
  volume={2026},
  pages={75694--75731},
  year={2026}
}

@inproceedings{chen2025scienceagentbench,
  title={Scienceagentbench: Toward rigorous assessment of language agents for data-driven scientific discovery},
  author={Chen, Ziru and Chen, Shijie and Ning, Yuting and Zhang, Qianheng and Wang, Boshi and Yu, Botao and Li, Yifei and Liao, Zeyi and Wei, Chen and Lu, Zitong and others},
  booktitle={International Conference on Learning Representations},
  volume={2025},
  pages={96934--96990},
  year={2025}
}

@inproceedings{jing2025dsbench,
  title={DSBench: How far are data science agents from becoming data science experts?},
  author={Jing, Liqiang and Huang, Zhehui and Wang, Xiaoyang and Yao, Wenlin and Yu, Wenhao and Ma, Kaixin and Zhang, Hongming and Du, Xinya and Yu, Dong},
  booktitle={International Conference on Learning Representations},
  volume={2025},
  pages={32597--32649},
  year={2025}
}

@inproceedings{zheng2026newtonbench,
  title={Newtonbench: Benchmarking generalizable scientific law discovery in llm agents},
  author={Zheng, Tianshi and Tam, Kiu Wai and Nguyen, Kim Hue Nam and Xu, Baixuan and Wang, Zhaowei and Jiayang, Cheng and Tsang, Hong Ting and Wang, Weiqi and Bai, Jiaxin and Fang, Tianqing and others},
  booktitle={International Conference on Learning Representations},
  volume={2026},
  pages={99018--99088},
  year={2026}
}

@inproceedings{zhang2026innogym,
  title={Innogym: Benchmarking the innovation potential of ai agents},
  author={Zhang, Jintian and Xu, Kewei and Zheng, Jingsheng and Yu, Zhuoyun and Zhu, Yuqi and Luo, Yujie and Wei, Lanning and Qiao, Shuofei and Du, Lun and Zheng, Da and others},
  booktitle={International Conference on Learning Representations},
  volume={2026},
  pages={71349--71379},
  year={2026}
}

@article{chen2024gmai,
  title={Gmai-mmbench: A comprehensive multimodal evaluation benchmark towards general medical ai},
  author={Chen, Pengcheng and Ye, Jin and Wang, Guoan and Li, Yanjun and Deng, Zhongying and Li, Wei and Li, Tianbin and Duan, Haodong and Huang, Ziyan and Su, Yanzhou and others},
  journal={Advances in Neural Information Processing Systems},
  volume={37},
  pages={94327--94427},
  year={2024}
}

@inproceedings{gan2023dispersion,
  title={Dispersion-oriented inverse design of photonic-crystal fiber for four-wave mixing application},
  author={Gan, Linqiao and Yu, Fei and Wang, Yazhou and Wang, Ning and Zhu, Xinyue and Hu, Lili and Yu, Chunlei},
  booktitle={Photonics},
  volume={10},
  number={3},
  pages={294},
  year={2023},
  organization={MDPI}
}

@article{ye2025maslab,
  title={Maslab: A unified and comprehensive codebase for llm-based multi-agent systems},
  author={Ye, Rui and Huang, Keduan and Wu, Qimin and Cai, Yuzhu and Jin, Tian and Pang, Xianghe and Liu, Xiangrui and Su, Jiaqi and Qian, Chen and Tang, Bohan and others},
  journal={arXiv preprint arXiv:2505.16988},
  year={2025}
}

@article{yao2023inverse,
  title={Inverse design of dispersion for photonic devices based on LSTM and gradient-free optimization algorithms hybridization},
  author={Yao, Tianhang and Huang, Tianye and Yan, Bin and Ge, Mingfeng and Yin, Jie and Peng, Chuyu and Li, Lu and Sun, Wufeng and Ping Shum, Perry},
  journal={Journal of the Optical Society of America B},
  volume={40},
  number={6},
  pages={1525--1532},
  year={2023},
  publisher={Optica Publishing Group}
}

@article{li2023deep,
  title={Deep reinforcement learning empowers automated inverse design and optimization of photonic crystals for nanoscale laser cavities},
  author={Li, Renjie and Zhang, Ceyao and Xie, Wentao and Gong, Yuanhao and Ding, Feilong and Dai, Hui and Chen, Zihan and Yin, Feng and Zhang, Zhaoyu},
  journal={Nanophotonics},
  volume={12},
  number={2},
  pages={319--334},
  year={2023},
  publisher={De Gruyter}
}

@article{nanophotonic_survey2025,
  title={Inverse design of nanophotonic devices enabled by optimization algorithms and deep learning: recent achievements and future prospects},
  author={Kim, Junhyeong and Kim, Jae-Yong and Kim, Jungmin and Hyeong, Yun and Neseli, Berkay and You, Jong-Bum and Shim, Joonsup and Shin, Jonghwa and Park, Hyo-Hoon and Kurt, Hamza},
  journal={Nanophotonics},
  volume={14},
  number={2},
  pages={121--151},
  year={2025},
  publisher={De Gruyter}
}

@article{hurst2024gpt,
  title={Gpt-4o system card},
  author={Hurst, Aaron and Lerer, Adam and Goucher, Adam P and Perelman, Adam and Ramesh, Aditya and Clark, Aidan and Ostrow, AJ and Welihinda, Akila and Hayes, Alan and Radford, Alec and others},
  journal={arXiv preprint arXiv:2410.21276},
  year={2024}
}

@misc{openai2025gpt41,
  title        = {Introducing {GPT-4.1} in the {API}},
  author       = {{OpenAI}},
  year         = {2025},
  howpublished = {\url{https://openai.com/index/gpt-4-1/}},
}

@article{singh2025openai,
  title={Openai gpt-5 system card},
  author={Singh, Aaditya and Fry, Adam and Perelman, Adam and Tart, Adam and Ganesh, Adi and El-Kishky, Ahmed and McLaughlin, Aidan and Low, Aiden and Ostrow, AJ and Ananthram, Akhila and others},
  journal={arXiv preprint arXiv:2601.03267},
  year={2025}
}

@misc{openai2025o3,
  title        = {Introducing {OpenAI} o3 and o4-mini},
  author       = {{OpenAI}},
  year         = {2025},
  howpublished = {\url{https://openai.com/index/introducing-o3-and-o4-mini/}},
}

@misc{anthropic2024claude35,
  title        = {Introducing {Claude 3.5 Sonnet}},
  author       = {{Anthropic}},
  year         = {2024},
  howpublished = {\url{https://www.anthropic.com/news/claude-3-5-sonnet}},
}

@misc{anthropic2025claude45,
  title        = {{Claude Sonnet 4.5} Model Card},
  author       = {{Anthropic}},
  year         = {2025},
  howpublished = {\url{https://www.anthropic.com/claude/sonnet}},
}

@misc{anthropic2025opus45,
  title        = {Introducing {Claude Opus 4.5}},
  author       = {{Anthropic}},
  year         = {2025},
  howpublished = {\url{https://www.anthropic.com/news/claude-opus-4-5}},
}

@article{comanici2025gemini,
  title={Gemini 2.5: Pushing the frontier with advanced reasoning, multimodality, long context, and next generation agentic capabilities},
  author={Comanici, Gheorghe and Bieber, Eric and Schaekermann, Mike and Pasupat, Ice and Sachdeva, Noveen and Dhillon, Inderjit and Blistein, Marcel and Ram, Ori and Zhang, Dan and Rosen, Evan and others},
  journal={arXiv preprint arXiv:2507.06261},
  year={2025}
}

@misc{gemini3-2025,
  title={{Gemini 3 Flash}: Frontier Intelligence Built for Speed},
  author={Doshi, Tulsee and {Gemini Team}},
  howpublished={\url{https://blog.google/products-and-platforms/products/gemini/gemini-3-flash/}},
  year={2025}
}

@article{team2025gemma,
  title={Gemma 3 technical report},
  author={Team, Gemma and Kamath, Aishwarya and Ferret, Johan and Pathak, Shreya and Vieillard, Nino and Merhej, Ramona and Perrin, Sarah and Matejovicova, Tatiana and Ram{\'e}, Alexandre and Rivi{\`e}re, Morgane and others},
  journal={arXiv preprint arXiv:2503.19786},
  year={2025}
}

@article{grattafiori2024llama,
  title={The {Llama 3} herd of models},
  author={Grattafiori, Aaron and Dubey, Abhimanyu and Jauhri, Abhinav and Pandey, Abhinav and Kadian, Abhishek and Al-Dahle, Ahmad and Letman, Aiesha and Mathur, Akhil and Schelten, Alan and Vaughan, Alex and others},
  journal={arXiv preprint arXiv:2407.21783},
  year={2024}
}

@misc{llama4-2025,
  title        = {The {Llama 4} herd: The beginning of a new era of natively multimodal {AI} innovation},
  author       = {{Meta AI}},
  year         = {2025},
  howpublished = {\url{https://ai.meta.com/blog/llama-4-multimodal-intelligence/}},
}

@article{wang2024qwen2,
  title={Qwen2-vl: Enhancing vision-language model's perception of the world at any resolution},
  author={Wang, Peng and Bai, Shuai and Tan, Sinan and Wang, Shijie and Fan, Zhihao and Bai, Jinze and Chen, Keqin and Liu, Xuejing and Wang, Jialin and Ge, Wenbin and others},
  journal={arXiv preprint arXiv:2409.12191},
  year={2024}
}

@misc{bai2025qwen25vltechnicalreport,
      title={Qwen2.5-VL Technical Report}, 
      author={Shuai Bai and Keqin Chen and Xuejing Liu and Jialin Wang and Wenbin Ge and Sibo Song and Kai Dang and Peng Wang and Shijie Wang and Jun Tang and Humen Zhong and Yuanzhi Zhu and Mingkun Yang and Zhaohai Li and Jianqiang Wan and Pengfei Wang and Wei Ding and Zheren Fu and Yiheng Xu and Jiabo Ye and Xi Zhang and Tianbao Xie and Zesen Cheng and Hang Zhang and Zhibo Yang and Haiyang Xu and Junyang Lin},
      year={2025},
      eprint={2502.13923},
      archivePrefix={arXiv},
      primaryClass={cs.CV},
      url={https://arxiv.org/abs/2502.13923}, 
}

@misc{bai2025qwen3vltechnicalreport,
      title={Qwen3-VL Technical Report}, 
      author={Shuai Bai and Yuxuan Cai and Ruizhe Chen and Keqin Chen and Xionghui Chen and Zesen Cheng and Lianghao Deng and Wei Ding and Chang Gao and Chunjiang Ge and Wenbin Ge and Zhifang Guo and Qidong Huang and Jie Huang and Fei Huang and Binyuan Hui and Shutong Jiang and Zhaohai Li and Mingsheng Li and Mei Li and Kaixin Li and Zicheng Lin and Junyang Lin and Xuejing Liu and Jiawei Liu and Chenglong Liu and Yang Liu and Dayiheng Liu and Shixuan Liu and Dunjie Lu and Ruilin Luo and Chenxu Lv and Rui Men and Lingchen Meng and Xuancheng Ren and Xingzhang Ren and Sibo Song and Yuchong Sun and Jun Tang and Jianhong Tu and Jianqiang Wan and Peng Wang and Pengfei Wang and Qiuyue Wang and Yuxuan Wang and Tianbao Xie and Yiheng Xu and Haiyang Xu and Jin Xu and Zhibo Yang and Mingkun Yang and Jianxin Yang and An Yang and Bowen Yu and Fei Zhang and Hang Zhang and Xi Zhang and Bo Zheng and Humen Zhong and Jingren Zhou and Fan Zhou and Jing Zhou and Yuanzhi Zhu and Ke Zhu},
      year={2025},
      eprint={2511.21631},
      archivePrefix={arXiv},
      primaryClass={cs.CV},
      url={https://arxiv.org/abs/2511.21631}, 
}

@misc{qwenteam2026qwen35omnitechnicalreport,
      title={Qwen3.5-Omni Technical Report}, 
      author={Qwen Team},
      year={2026},
      eprint={2604.15804},
      archivePrefix={arXiv},
      primaryClass={cs.CL},
      url={https://arxiv.org/abs/2604.15804}, 
}

@misc{agrawal2024pixtral12b,
      title={Pixtral 12B}, 
      author={Pravesh Agrawal and Szymon Antoniak and Emma Bou Hanna and Baptiste Bout and Devendra Chaplot and Jessica Chudnovsky and Diogo Costa and Baudouin De Monicault and Saurabh Garg and Theophile Gervet and Soham Ghosh and Amélie Héliou and Paul Jacob and Albert Q. Jiang and Kartik Khandelwal and Timothée Lacroix and Guillaume Lample and Diego Las Casas and Thibaut Lavril and Teven Le Scao and Andy Lo and William Marshall and Louis Martin and Arthur Mensch and Pavankumar Muddireddy and Valera Nemychnikova and Marie Pellat and Patrick Von Platen and Nikhil Raghuraman and Baptiste Rozière and Alexandre Sablayrolles and Lucile Saulnier and Romain Sauvestre and Wendy Shang and Roman Soletskyi and Lawrence Stewart and Pierre Stock and Joachim Studnia and Sandeep Subramanian and Sagar Vaze and Thomas Wang and Sophia Yang},
      year={2024},
      eprint={2410.07073},
      archivePrefix={arXiv},
      primaryClass={cs.CV},
      url={https://arxiv.org/abs/2410.07073}, 
}

@misc{mistrallarge-2024,
  title        = {{Mistral-Large-3-675B-Instruct-2512} model card},
  author       = {{Mistral AI}},
  year         = {2025},
  howpublished = {\url{https://huggingface.co/mistralai/Mistral-Large-3-675B-Instruct-2512}},
}

@misc{ministral-2024,
  title        = {{Ministral-3-8B-Instruct-2512} model card},
  author       = {{Mistral AI}},
  year         = {2025},
  howpublished = {\url{https://huggingface.co/mistralai/Ministral-3-8B-Instruct-2512}},
}

@inproceedings{deitke2025molmo,
  title={Molmo and pixmo: Open weights and open data for state-of-the-art vision-language models},
  author={Deitke, Matt and Clark, Christopher and Lee, Sangho and Tripathi, Rohun and Yang, Yue and Park, Jae Sung and Salehi, Mohammadreza and Muennighoff, Niklas and Lo, Kyle and Soldaini, Luca and others},
  booktitle={2025 IEEE/CVF Conference on Computer Vision and Pattern Recognition (CVPR)},
  pages={91--104},
  year={2025},
  organization={IEEE}
}

@misc{grok4-2025,
  title        = {{Grok 4}},
  author       = {{xAI}},
  year         = {2025},
  howpublished = {\url{https://x.ai/news/grok-4}},
}

@article{hong2025glm,
  title={Glm-4.5 v and glm-4.1 v-thinking: Towards versatile multimodal reasoning with scalable reinforcement learning},
  author={Hong, Wenyi and Yu, Wenmeng and Gu, Xiaotao and Wang, Guo and Gan, Guobing and Tang, Haomiao and Cheng, Jiale and Qi, Ji and Ji, Junhui and Pan, Lihang and others},
  journal={arXiv preprint arXiv:2507.01006},
  year={2025}
}

@article{wei2022chain,
  title={Chain-of-thought prompting elicits reasoning in large language models},
  author={Wei, Jason and Wang, Xuezhi and Schuurmans, Dale and Bosma, Maarten and Xia, Fei and Chi, Ed and Le, Quoc V and Zhou, Denny and others},
  journal={Advances in neural information processing systems},
  volume={35},
  pages={24824--24837},
  year={2022}
}

@article{wang2022self,
  title={Self-consistency improves chain of thought reasoning in language models},
  author={Wang, Xuezhi and Wei, Jason and Schuurmans, Dale and Le, Quoc and Chi, Ed and Narang, Sharan and Chowdhery, Aakanksha and Zhou, Denny},
  journal={arXiv preprint arXiv:2203.11171},
  year={2022}
}

@article{du2023improving,
  title={Improving Factuality and Reasoning in Language Models through Multiagent Debate},
  author={Du, Yilun and Li, Shuang and Torralba, Antonio and Tenenbaum, Joshua B and Mordatch, Igor},
  journal={arXiv preprint arXiv:2305.14325},
  year={2023}
}

@inproceedings{liang2024encouraging,
  title={Encouraging divergent thinking in large language models through multi-agent debate},
  author={Liang, Tian and He, Zhiwei and Jiao, Wenxiang and Wang, Xing and Wang, Yan and Wang, Rui and Yang, Yujiu and Shi, Shuming and Tu, Zhaopeng},
  booktitle={Proceedings of the 2024 conference on empirical methods in natural language processing},
  pages={17889--17904},
  year={2024}
}

@article{li2023camel,
  title={Camel: Communicative agents for" mind" exploration of large language model society},
  author={Li, Guohao and Hammoud, Hasan and Itani, Hani and Khizbullin, Dmitrii and Ghanem, Bernard},
  journal={Advances in neural information processing systems},
  volume={36},
  pages={51991--52008},
  year={2023}
}

@article{liu2023dynamic,
  title={A dynamic LLM-powered agent network for task-oriented agent collaboration},
  author={Liu, Zijun and Zhang, Yanzhe and Li, Peng and Liu, Yang and Yang, Diyi},
  journal={arXiv preprint arXiv:2310.02170},
  year={2023}
}

@inproceedings{qian2025scaling,
  title={Scaling large language model-based multi-agent collaboration},
  author={Qian, Chen and Xie, Zihao and Wang, YiFei and Liu, Wei and Zhu, Kunlun and Xia, Hanchen and Dang, Yufan and Du, Zhuoyun and Chen, Weize and Yang, Cheng and others},
  booktitle={International Conference on Learning Representations},
  volume={2025},
  pages={41488--41505},
  year={2025}
}

@inproceedings{chen2024agentverse,
  title={Agentverse: Facilitating multi-agent collaboration and exploring emergent behaviors},
  author={Chen, Weize and Su, Yusheng and Zuo, Jingwei and Yang, Cheng and Yuan, Chenfei and Chan, Chi-Min and Yu, Heyang and Lu, Yaxi and Hung, Yi-Hsin and Qian, Chen and others},
  booktitle={International Conference on Learning Representations},
  volume={2024},
  pages={20094--20136},
  year={2024}
}

@article{wu2023autogen,
  title={Autogen: Enabling next-gen llm applications via multi-agent conversation},
  author={Wu, Qingyun and Bansal, Gagan and Zhang, Jieyu and Wu, Yiran and Li, Beibin and Zhu, Erkang and Jiang, Li and Zhang, Xiaoyun and Zhang, Shaokun and Liu, Jiale and others},
  journal={arXiv preprint arXiv:2308.08155},
  year={2023}
}

@inproceedings{hu2025self,
  title={Self-evolving multi-agent collaboration networks for software development},
  author={Hu, Yue and Cai, Yuzhu and Du, Yaxin and Zhu, Xinyu and Liu, Xiangrui and Yu, Zijie and Hou, Yuchen and Tang, Shuo and Chen, Siheng},
  booktitle={International Conference on Learning Representations},
  volume={2025},
  pages={23007--23039},
  year={2025}
}

@article{lifshitz2025multi,
  title={Multi-agent verification: Scaling test-time compute with multiple verifiers},
  author={Lifshitz, Shalev and McIlraith, Sheila A and Du, Yilun},
  journal={arXiv preprint arXiv:2502.20379},
  year={2025}
}

@article{raissi2019physics,
  title={Physics-informed neural networks: A deep learning framework for solving forward and inverse problems involving nonlinear partial differential equations},
  author={Raissi, Maziar and Perdikaris, Paris and Karniadakis, George E},
  journal={Journal of Computational physics},
  volume={378},
  pages={686--707},
  year={2019},
  publisher={Elsevier}
}

@article{peurifoy2018nanophotonic,
  title={Nanophotonic particle simulation and inverse design using artificial neural networks},
  author={Peurifoy, John and Shen, Yichen and Jing, Li and Yang, Yi and Cano-Renteria, Fidel and DeLacy, Brendan G and Joannopoulos, John D and Tegmark, Max and Solja{\v{c}}i{\'c}, Marin},
  journal={Science advances},
  volume={4},
  number={6},
  pages={eaar4206},
  year={2018},
  publisher={American Association for the Advancement of Science}
}

@inproceedings{wu2025picbench,
  title={PICBench: Benchmarking LLMs for photonic integrated circuits design},
  author={Wu, Yuchao and Yu, Xiaofei and Chen, Hao and Luo, Yang and Tong, Yeyu and Ma, Yuzhe},
  booktitle={2025 Design, Automation \& Test in Europe Conference (DATE)},
  pages={1--6},
  year={2025},
  organization={IEEE}
}

@article{chen2022cost,
  title={Cost-effective improvement of the performance of AWG-based FBG wavelength interrogation via a cascaded neural network},
  author={Chen, Shengchao and Yao, Feifan and Ren, Sufen and Wang, Guanjun and Huang, Mengxing},
  journal={Optics Express},
  volume={30},
  number={5},
  pages={7647--7663},
  year={2022},
  publisher={Optica Publishing Group}
}

@article{chen2023collaborative,
  title={Collaborative photonic crystal fiber property optimization: A new paradigm for reverse design},
  author={Chen, Shengchao and Wang, Xinchen and Ren, Sufen and Yang, Jianli and Zhang, Yaqian and Wang, Guanjun},
  journal={IEEE Photonics Technology Letters},
  volume={35},
  number={19},
  pages={1035--1038},
  year={2023},
  publisher={IEEE}
}

@article{chen2025pcf,
  title={PCF-LLM: scaling LLMs for multimodal understanding of structured scientific data in photonic crystal fiber sensors},
  author={Chen, Shengchao and Hu, Geyao and Ren, Sufen and Shu, Ting},
  journal={Information Fusion},
  pages={104022},
  year={2025},
  publisher={Elsevier}
}

@article{ren2024distributed,
  title={A distributed photonic crystal fiber reverse design framework based on multi-source knowledge fusion},
  author={Ren, Sufen and Chen, Shengchao and Wang, Jiahao and Xu, Haoyang and Hou, Xuan and Huang, Mengxing and Liu, Jianxun and Wang, Guanjun},
  journal={Optical Fiber Technology},
  volume={84},
  pages={103718},
  year={2024},
  publisher={Elsevier}
}

@article{yuan2022efficient,
  title={Efficient calculation of optical properties of suspended-core fiber via a machine learning algorithm},
  author={Yuan, Shuyu and Chen, Shengchao and Yang, Jianli and Yang, Qian and Ren, Sufen and Wang, Guanjun and Yu, Benguo},
  journal={Applied Optics},
  volume={61},
  number={19},
  pages={5714--5721},
  year={2022},
  publisher={Optica Publishing Group}
}

@inproceedings{feng2025taming,
  title={Taming Vision-Language Models for Federated Foundation Models on Heterogeneous Medical Imaging Modalities},
  author={Feng, Lulu and Chen, Shengchao},
  booktitle={Proceedings of the 2025 International Conference on Multimedia Retrieval},
  pages={303--311},
  year={2025}
}

@article{feng2026visual,
  title={Visual and textual spaces both matter: Taming CLIP for non-IID federated medical image classification},
  author={Feng, Lulu and Chen, Shengchao},
  journal={Knowledge-Based Systems},
  volume={338},
  pages={115524},
  year={2026},
  publisher={Elsevier}
}

@inproceedings{chen2025restyled,
  title={Restyled, Tuning, and Alignment: Taming VLMs for Federated Non-IID Medical Image Analysis},
  author={Chen, Shengchao and Shu, Ting},
  booktitle={International Conference on Medical Image Computing and Computer-Assisted Intervention},
  pages={530--540},
  year={2025},
  organization={Springer}
}

@article{ma2024optogpt,
  title={OptoGPT: a foundation model for inverse design in optical multilayer thin film structures},
  author={Ma, Taigao and Wang, Haozhu and Guo, L Jay},
  journal={Opto-Electronic Advances},
  volume={7},
  number={7},
  year={2024},
  publisher={Opto-Electronic Advance}
}

@article{lu2025agentic,
  title={An agentic framework for autonomous metamaterial modeling and inverse design},
  author={Lu, Darui and Malof, Jordan M and Padilla, Willie J},
  journal={ACS Photonics},
  volume={12},
  number={11},
  pages={6071--6080},
  year={2025},
  publisher={ACS Publications}
}

@article{kharel2026autonomous,
  title={Autonomous agentic design for photonics},
  author={Kharel, Prashanta and Khavasi, Amin and Chen, Xinzhong and Hughes, Tyler W},
  journal={arXiv preprint arXiv:2606.00915},
  year={2026}
}

@article{zhao2020mode,
  title={Mode-phase-difference photothermal spectroscopy for gas detection with an anti-resonant hollow-core optical fiber},
  author={Zhao, Pengcheng and Zhao, Yan and Bao, Haihong and Ho, Hoi Lut and Jin, Wei and Fan, Shangchun and Gao, Shoufei and Wang, Yingying and Wang, Pu},
  journal={Nature communications},
  volume={11},
  number={1},
  pages={847},
  year={2020},
  publisher={Nature Publishing Group UK London}
}

@article{petrovich2025broadband,
  title={Broadband optical fibre with an attenuation lower than 0.1 decibel per kilometre},
  author={Petrovich, Marco and Numkam Fokoua, Eric and Chen, Yong and Sakr, Hesham and Adamu, Abubakar Isa and Hassan, Rosdi and Wu, Dong and Fatobene Ando, Ron and Papadimopoulos, Athanasios and Sandoghchi, Seyed Reza and others},
  journal={Nature Photonics},
  volume={19},
  number={11},
  pages={1203--1208},
  year={2025},
  publisher={Nature Publishing Group UK London}
}

@article{wheeler2017low,
  title={Low-loss Kagome hollow-core fibers operating from the near-to the mid-IR},
  author={Wheeler, NV and Bradley, TD and Hayes, JR and Gouveia, MA and Liang, Sijing and Chen, Y and Sandoghchi, SR and Abokhamis Mousavi, Seyed Mohammad and Poletti, F and Petrovich, MN and others},
  journal={Optics Letters},
  volume={42},
  number={13},
  pages={2571--2574},
  year={2017},
  publisher={Optical Society of America}
}

@article{slavik2015ultralow,
  title={Ultralow thermal sensitivity of phase and propagation delay in hollow core optical fibres},
  author={Slav{\'\i}k, Radan and Marra, Giuseppe and Fokoua, Eric Numkam and Baddela, Naveen and Wheeler, Natalie V and Petrovich, Marco and Poletti, Francesco and Richardson, David J},
  journal={Scientific reports},
  volume={5},
  number={1},
  pages={15447},
  year={2015},
  publisher={Nature Publishing Group UK London}
}

@article{caillaud2014photonic,
  title={Photonic bandgap propagation in all-solid chalcogenide microstructured optical fibers},
  author={Caillaud, Celine and Renversez, Gilles and Brilland, Laurent and Mechin, David and Calvez, Laurent and Adam, Jean-Luc and Troles, Johann},
  journal={Materials},
  volume={7},
  number={9},
  pages={6120--6129},
  year={2014},
  publisher={MDPI}
}

@article{cardenas2013high,
  title={High-visibility photonic crystal fiber interferometer as multifunctional sensor},
  author={C{\'a}rdenas-Sevilla, GA and F{\'a}vero, Fernando C and Villatoro, Joel},
  journal={Sensors},
  volume={13},
  number={2},
  pages={2349--2358},
  year={2013},
  publisher={Molecular Diversity Preservation International (MDPI)}
}

@article{zhang2018thermally,
  title={Thermally optimized polarization-maintaining photonic crystal fiber and its FOG application},
  author={Zhang, Chunxi and Zhang, Zhihao and Xu, Xiaobin and Cai, Wei},
  journal={Sensors},
  volume={18},
  number={2},
  pages={567},
  year={2018},
  publisher={MDPI}
}

@article{catalano2023tuning,
  title={Tuning of the Brillouin scattering properties in microstructured optical fibers by liquid infiltration},
  author={Catalano, Ester and Vallifuoco, Raffaele and Zeni, Luigi and Dufour, Alexis and Marin, Emmanuel and Girard, Sylvain and Minardo, Aldo},
  journal={Scientific Reports},
  volume={13},
  number={1},
  pages={10468},
  year={2023},
  publisher={Nature Publishing Group UK London}
}

@article{zhang2024novel,
  title={A novel multicore Er/Yb co-doped microstructured optical fiber amplifier with peanut-shaped air holes cladding},
  author={Zhang, Yifan and Zhao, Yifei and Fang, Ziwei and Liu, Jiantao and Xia, Changming and Hou, Zhiyun and Zhao, Xuesong and Tan, Zhongwei and Dong, Yi and Zhou, Guiyao and others},
  journal={Nanophotonics},
  volume={13},
  number={6},
  pages={891--899},
  year={2024},
  publisher={De Gruyter}
}

@article{zhao2018simultaneous,
  title={Simultaneous measurement of curvature, strain and temperature using a twin-core photonic crystal fiber-based sensor},
  author={Zhao, Tongtong and Lou, Shuqin and Wang, Xin and Zhang, Wan and Wang, Yulei},
  journal={Sensors},
  volume={18},
  number={7},
  pages={2145},
  year={2018},
  publisher={MDPI}
}

@inproceedings{zhai2023sigmoid,
  title={Sigmoid loss for language image pre-training},
  author={Zhai, Xiaohua and Mustafa, Basil and Kolesnikov, Alexander and Beyer, Lucas},
  booktitle={2023 IEEE/CVF International Conference on Computer Vision (ICCV)},
  pages={11941--11952},
  year={2023},
  organization={IEEE}
}
